\documentclass[10pt,twocolumn,letterpaper]{article}

\usepackage[pagenumbers]{cvpr} % To force page numbers, e.g. for an arXiv version
\usepackage[dvipsnames]{xcolor}

\definecolor{cvprblue}{rgb}{0.21,0.49,0.74}
\usepackage[pagebackref,breaklinks,colorlinks,citecolor=cvprblue]{hyperref}
\usepackage[accsupp]{axessibility} % Improves PDF readability for those with disabilities.

\newcommand{\Tref}[1]{Table~\ref{#1}}

\newcommand{\fref}[1]{Fig.~\ref{#1}}

\newcommand{\sref}[1]{Sec.~\ref{#1}}

\def\paperID{12223} % *** Enter the Paper ID here
\def\confName{CVPR}
\def\confYear{2024}

\title{3D Geometry-aware Deformable Gaussian Splatting for Dynamic View Synthesis}

\author{%
	Zhicheng Lu$^{1}$\footnotemark[1], \quad Xiang Guo$^{1}$\footnotemark[1], \quad Le Hui$^{1}$\footnotemark[2], \quad Tianrui Chen$^{1, 2}$,  \\ \quad Min Yang$^{2}$, \quad 
	Xiao Tang$^{2}$, \quad Feng Zhu$^{2}$, \quad Yuchao Dai$^{1}$\footnotemark[2] \\
	$^1$Northwestern Polytechnical University
	$^2$Samsung R\&D Institute\\
	\texttt{\small\{zhichenglu, guoxiang, cherryxchen\}@mail.nwpu.edu.cn}\\
	\texttt{\small\{daiyuchao, huile\}@nwpu.edu.cn}  \;\;
	\texttt{\small\{min16.yang, xiao1.tang, f15.zhu\}@samsung.com} 
}

\begin{document}
\maketitle

{
\renewcommand{\thefootnote}{\fnsymbol{footnote}}
\footnotetext[1]{Equal contributions.}
\footnotetext[2]{Corresponding authors.}
}
\begin{abstract}
In this paper, we propose a 3D geometry-aware deformable Gaussian Splatting method for dynamic view synthesis. 
Existing neural radiance fields (NeRF) based solutions learn the deformation in an implicit manner, which cannot incorporate 3D scene geometry. Therefore, the learned deformation is not necessarily geometrically coherent, which results in unsatisfactory dynamic view synthesis and 3D dynamic reconstruction.
Recently, 3D Gaussian Splatting provides a new representation of the 3D scene, building upon which the 3D geometry could be exploited in learning the complex 3D deformation. 
Specifically, the scenes are represented as a collection of 3D Gaussian, where each 3D Gaussian is optimized to move and rotate over time to model the deformation. To enforce the 3D scene geometry constraint during deformation, we explicitly extract 3D geometry features and integrate them in learning the 3D deformation. 
In this way, our solution achieves 3D geometry-aware deformation modeling, which enables improved dynamic view synthesis and 3D dynamic reconstruction. Extensive experimental results on both synthetic and real datasets prove the superiority of our solution, which achieves new state-of-the-art performance. The project is available at \href{https://npucvr.github.io/GaGS/}{https://npucvr.github.io/GaGS/}. 
\end{abstract}    
\section{Introduction}
\label{sec:intro}

Dynamic View Synthesis (DVS) aims at rendering novel photorealistic views at arbitrary viewpoints and any input time step given a monocular video of a \emph{dynamic} scene, which has broad applications in virtual reality and augmented reality. Recently, empowered with effective representations such as neural radiance fields (NeRF)~\cite{mildenhall2020_nerf_eccv20} and Gaussian Splatting~\cite{3Dgaussian}, novel view synthesis for static scenes has been greatly advanced. However, this success cannot be extended to its dynamic counterpart directly. This is mainly due to the difficulty in modeling and representing the scene deformation. Due to the inherent motion/shape ambiguity in monocular dynamic 3D representation, dynamic scene modeling and synthesis are more challenging, especially for monocular video with limited observations.

\begin{figure}[!t]
\centering
\includegraphics[width=\columnwidth]{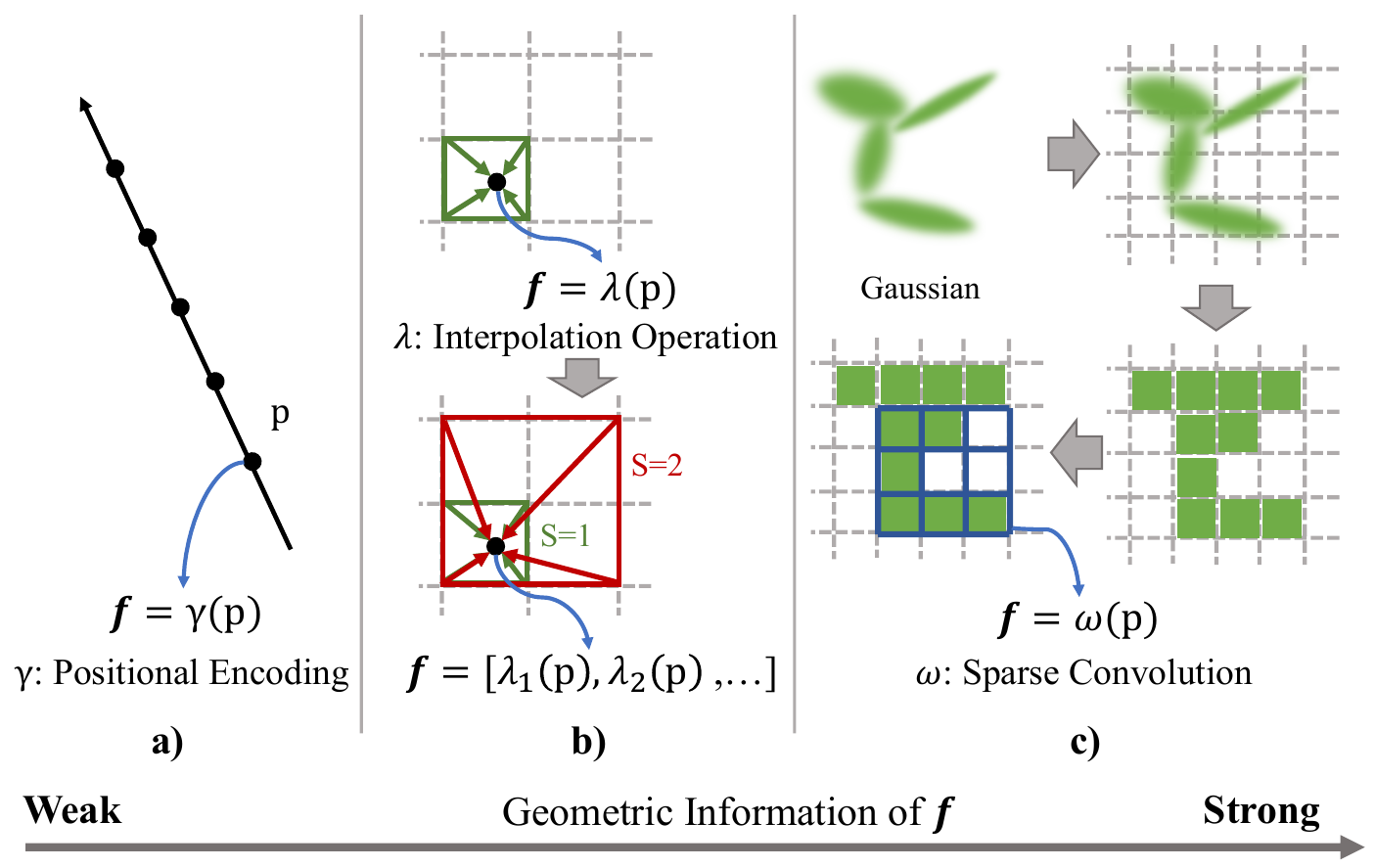}
\vspace{-0.7cm}
\caption{\textbf{Geometric information exploited by different methods}. a) Early dynamic NeRF methods such as DNeRF\cite{pumarola2021_dnerf_cvpr21} directly encode the coordinate $\mathbf{p}$ of the sample point as input feature for deformation network. b) Interpolation is used to fuse features from neighbouring grids and mulitscale interpolation enhances the local geometry information~\cite{guo2022_NDVG_arxiv, liu2023_robust_CVPR, fang2022_TANV_arxiv, turki2023_SUDS_CVPR}. c) We propose to voxelize a set of Gaussian distributions and use a sparse convolution network to extract geometry-aware features for deformation learning.}
\label{fig:teaser}
\vspace{-0.6cm}
\end{figure}

In addressing the above challenges, one common strategy is to represent the dynamic scenes as a combination of a static canonical field and a deformation model~\cite{pumarola2021_dnerf_cvpr21, park2021_nerfies_iccv21, park2021hypernerf, tretschk2020_nonrigid_iccv21, fang2022_TANV_arxiv, guo2022_NDVG_arxiv, Guo2023_Forward_ICCV, turki2023_SUDS_CVPR, liu2023_robust_CVPR}, whereas the bottleneck lies in representing the diverse and complex real-world 3D deformation. To represent geometrically consistent 3D deformation, the local geometric/structural information is critical, since the deformations of the objects in the real world are highly correlated to their 3D structures. 
Furthermore, the motions of the object points are deeply coupled with the motions of their neighboring points.
Thus, how to incorporate the local geometric information to learn locally smooth and consistent 3D deformations becomes the research focus in DVS.

Recently, different deformation models have utilized the local geometric information, but they all have their limitations. As shown in \fref{fig:teaser} \textbf{a)}, originally in D-NeRF \cite{pumarola2021_dnerf_cvpr21}, the feature (positional encoding) of each sampled point is extracted \emph{independently} with each other. Following works notice that this method could not handle the complex dynamic scene since the extracted features contain little information from neighboring points. In \fref{fig:teaser} \textbf{b)}, interpolation is introduced to fuse features of neighboring grids. NDVG \cite{guo2022_NDVG_arxiv} and RoDynRF \cite{liu2023_robust_CVPR} gradually increase the voxel resolution so that the large voxel size could cover a larger area, introducing the local smoothness at the early stage of the training. However, this strategy has a limited cover range of local areas and cannot work at a later training stage. TiNeuVox \cite{fang2022_TANV_arxiv} and SUDS \cite{turki2023_SUDS_CVPR} interpolate with multi-scales. Nevertheless, the interpolation operation is rather simple in extracting local geometric information and introduces un-smoothness and artifacts \cite{hu2023_TriMipRF_ICCV, barron2023_zipnerf_ICCV}. 

In modling the nonrigid deformation, it is crucial to account for the consistency in the motion of local neighborhood. Since point-level MLP has a limited receptive field, which cannot capture the local geometric features of point clouds. To utilize the local geometric information effectively, we propose to use 3D sparse convolution. As shown in \fref{fig:teaser} \textbf{c)}, building upon the recent explicit point cloud based Gaussian Splatting representation, we introduce a sparse convolution network to extract 3D geometry-aware features. Compared with simple feature interpolation, the convolutional neural network is superior in extracting local information and has a much larger reception field. Also, we treat the 3D Gaussian distributions as point clouds, which enable sparse 3D convolution for time and memory efficiency. Note that FDNeRF \cite{Guo2023_Forward_ICCV} uses a 3D U-Net to inpaint the missing area in the voxel grid. But this inpaint network is not used for deformation modeling, while the rendering speed and voxel resolution are also limited.

Originally in Guassian Splatting~\cite{3Dgaussian}, the rotation parameter of each Gaussian is represented by quaternion. However, quaternion representation for rotation is discontinuous in parameter space for neural network learning~\cite{zhou2019continuity}. We introduce the continuous 6D rotation~\cite{zhou2019continuity} to ensure that the network learns a continuous function in the parameter space, which accurately represents the rotational states of each Gaussian at different time. 

Overall, our method mainly has two components: a Gaussian canoncial field and a deformation field. The Gaussian canonical field consists of 3D Gaussian distributions and a geometry-aware feature learning network. The explicit 3D Gaussian distribution represents the geometry of the canonical scene, and the sparse 3D CNN network extracts local structural/geometric information for each Gaussian. The deformation field estimates a transformation for each Gaussian in the canonical field, which transfers the Gaussian from the canonical field to the given timestamp. Finally, we use 3D Gaussian splatting to render images for the given timestamp.

% \todo{[Contributions]}
Our main contributions are summarized as:
\begin{itemize}
    \item We propose a geometry-aware feature extraction network based on 3D Gaussian distribution to better utilize local geometric information.
    \item We propose to use continuous 6D rotation representation and modified density control strategy to adapt Gaussian splatting to dynamic scenes.
    \item Extensive experiments on both synthetic and real datasets show that our method surpasses competing methods by a wide margin.
\end{itemize}

%=======================================================================
\section{Related Work}

\begin{figure*}[th]
\centering
\includegraphics[width=0.95\textwidth]{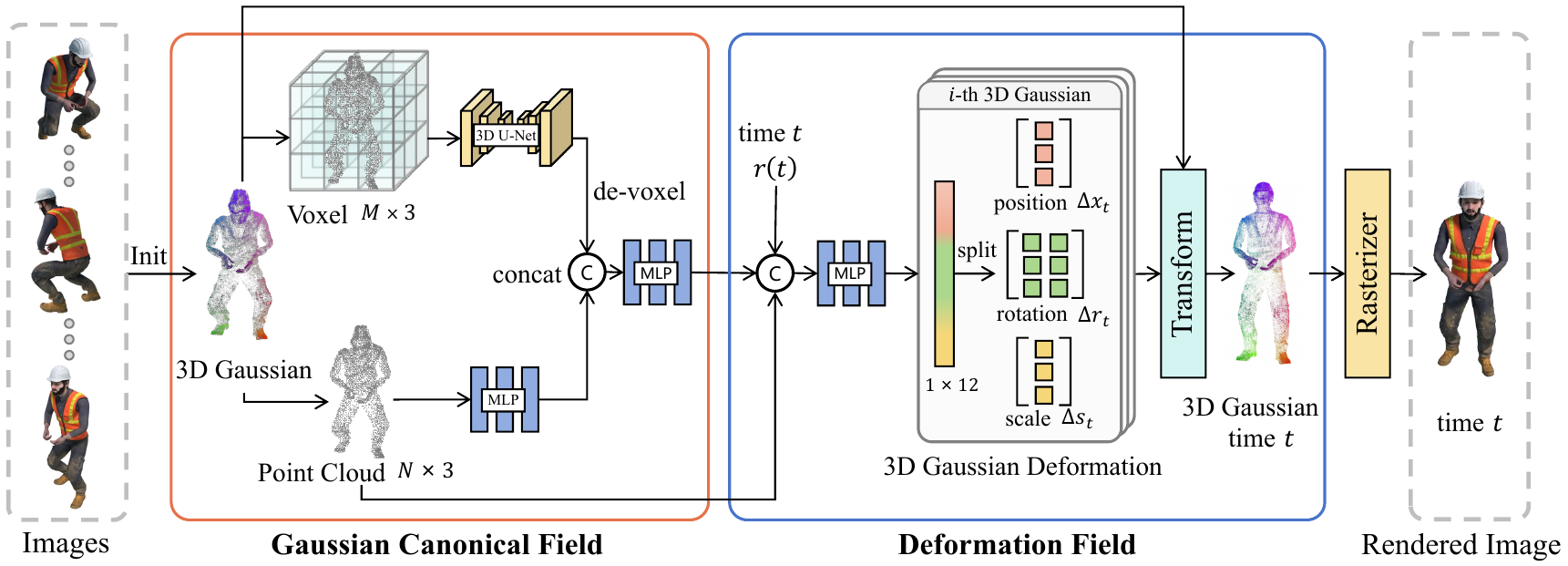}
\vspace{-8pt}
\caption{The pipeline of our proposed 3D geometry-aware deformable Gaussian splitting. In the Gaussian canonical field, we reconstruct a static scene in canonical space using 3D Gaussian distributions. We extract positional features using an MLP, as well as local geometric features using a 3D U-Net, fused by another MLP to form the geometry-aware features. In the deformation field, taking the geometry-aware features and timestamp $t$, an MLP estimates the 3D Gaussian deformation, which transfers the canonical 3D Gaussian distributions to timestamp $t$. Finally, a rasterizer renders the transformed 3D Gaussian to images.}
\label{fig:framework}
\vspace{-18pt}
\end{figure*}

\subsection{Novel View Synthesis}

Novel View Synthesis (NVS) is a well-known task in both computer vision and graphics~\cite{buehler2001_unstructured,chen1993view,levoy1996light,greene1986environment}. Surveys such as~\cite{shum2000review,tewari2020state,tewari2021advances} provide comprehensive discussions. 
Explicit NVS methods generally reconstruct an explicit 3D model of a scene in the form of point clouds~\cite{aliev2020_neural_eccv20}, voxels, or meshes~\cite{riegler2020_freenvs_eccv20,riegler2021_stablenvs_cvpr21,thies2019_deferred_tog19,hedman2018_deepibr_tog18}. 
%    These 3D representations contain numerous information of a scene. 
Once the geometry of the scene is represented, novel view images can be rendered from arbitrary viewpoints via manipulating the camera pose parameters. 
Other methods ~\cite{kalantari2016learning_tog16,penner2017_soft_tog17,choi2019_extreme_nvs_iccv19,riegler2020_freenvs_eccv20,riegler2021_stablenvs_cvpr21,flynn2016_deepstereo_cvpr16,xu2019_deep_tog19} tackle NVS by estimating depth maps using multi-view geometry, whereas the features are aggregated from co-visible frames.

%   Recently, neural implicit representations have shown great promise in the field of NVS and 3D modeling.

Neural Radiance Fields (NeRF)~\cite{mildenhall2020_nerf_eccv20} is a groundbreaking approach that utilizes Multi-Layer Perceptrons (MLPs) to represent scenes implicitly. This methodology enables the modeling of a 5D radiance field, resulting in the impressive synthesis of views for static scenes. 
Numerous subsequent works expand the capabilities of NeRF by adapting it to various scenarios, such as handling larger and unbounded scenes ~\cite{Zhang20arxiv_nerf++,tancik2022_blocknerf_cvpr,xiangli2022bungeenerf_eccv,Rematas2022_urbanNerf_CVPR,martin2021_nerfw_cvpr21},
% handling in-the-wild scenes~\cite{martin2021_nerfw_cvpr21}, 
scene editing and relighting, ~\cite{boss2021nerd,srinivasan2021nerv,zhang2021nerfactor,yang2022s}, \cite{barron2021mipnerf, hu2023_TriMipRF_ICCV, barron2023_zipnerf_ICCV}, and improving the generalization ability~\cite{chen2021mvsnerf,trevithick2021grf,yu2021_pixelnerf_cvpr21,wang2021_ibrnet_cvpr21}.
Meanwhile, researchers focus on achieving more efficient rendering and optimization in a NeRF-like framework. \cite{neff2021_donerf_egsr21,lindell2021_autoint_cvpr21, piala2021terminerf, liu2020_nsvf_nips20,yu2021_plenoctrees_iccv21,lombardi2021mixture, kplanes, hexplane} investigate efficient sampling methods along each ray for color accumulation, while~\cite{rebain2021_derf_cvpr21, Reiser2021_kiloNeRF_iccv21} partition the scene into multiple sub-regions as an efficient pre-processing, and~\cite{yu2021plenoxels,sun2021direct,muller2022instant,kplanes, hexplane} exploit voxel-grid representation to speed up the optimization. 
Very recently, \cite{3Dgaussian} proposes to use 3D Guassian distribution to represent the scene, obtaining promising results. However, these methods are mainly applicable to static scenes, and fail in scenes with dynamic objects.

\subsection{Dynamic View Synthesis}
A recent trend in NVS is to extend the success in static NVS to dynamic NVS. 
%          Numerous studies have undertaken the endeavor of expanding the capabilities of NeRF, transitioning from its traditional focus on static scenes to encompass dynamic scenes involving non-rigid and deformable objects.
% ~\cite{gao2021dynamic_iccv21,li2021_nsff_cvpr21,xian2021_space_cvpr21,tretschk2020_nonrigid_iccv21,park2021_nerfies_iccv21,pumarola2021_dnerf_cvpr21,wang2021_dctnerf_arxiv,du2021_nerflow_iccv21}.
One viable strategy is to construct a 4D spatial-temporal representation. Yoon \etal.~\cite{yoon2020_nvidiadataset_cvpr20} combine single-view and multi-view depth to achieve NVS by 3D warping. Gao \etal.~\cite{gao2021dynamic_iccv21} use a time-invariant model and a time-varying model to represent the static part and dynamic part of a scene, respectively, and use scene flow for motion modeling. NeRFlow~\cite{du2021_nerflow_iccv21} proposes a 4D spatial-temporal representation of a dynamic scene. Xian \etal.~\cite{xian2021_space_cvpr21} map a spatial-temporal location to the color and volume density by a 4D spatial-temporal radiance field. NSFF~\cite{li2021_nsff_cvpr21} represents a dynamic scene as a continuously changing function, encompassing various aspects of the scene, including appearance, geometry, and 3D scene motion. DCT-NeRF~\cite{wang2021_dctnerf_arxiv} uses the Discrete Cosine Transform (DCT) to replace the scene flow in NSFF~\cite{li2021_nsff_cvpr21} to enable smoother motion trajectories. HexPlane~\cite{hexplane} and K-Plane~\cite{kplanes} project 4D spatial-temporal space to multiple 2D planes.

On the other hand, works such as ~\cite{pumarola2021_dnerf_cvpr21, park2021_nerfies_iccv21, park2021hypernerf, nerfplayer, tretschk2020_nonrigid_iccv21, guo2022_NDVG_arxiv, Guo2023_Forward_ICCV, fang2022_TANV_arxiv, liu2023_robust_CVPR, turki2023_SUDS_CVPR} decode the dynamic scene with a canonical field and a deformation field. 
Along this pipeline, D-NeRF~\cite{pumarola2021_dnerf_cvpr21} first proposes the canonical-based framework. However, the deformation network utilizes positional features with little geometry information, which cannot handle complex dynamic scenarios well. Nerfies~\cite{park2021_nerfies_iccv21} proposes a coarse-to-fine optimization method for coordinate-based models that allows for more robust optimization. HyperNeRF~\cite{park2021hypernerf} lifts the canonical field into a higher dimensional space to handle topological changes. NDVG~\cite{guo2022_NDVG_arxiv} and RoDynRF~\cite{liu2023_robust_CVPR} gradually increase the voxel resolution, which has two benefits. 
TiNeuVox~\cite{fang2022_TANV_arxiv} and SUDS~\cite{turki2023_SUDS_CVPR} interpolate the features with multi-scales for deformation learning. The multi-scales interpolation covers a larger reception field, which benefits modeling varying motions.

Very recently, with the stunning debut of 3D Gaussian~\cite{3Dgaussian}, some works introduce this point-based representation into their pipelines to synthesize high-fidelity images of a dynamic scene. Deformable 3DGS \cite{yang2023deformable3dgs} proposes a deformable 3D Gaussian framework with a novel annealing smoothing training mechanism, which achieves real-time rendering in dynamic scenes.
Wu \etal \cite{4DGaussian} introduce a 4D Gaussian Splatting representation and utilize a deformation field to model both Gaussian motions and shape changes. 
%   To predict motions and shape deformations of objects, their network connects different adjacent Gaussians with a HexPlane. 
However, the multi-scale HexPlane interpolation has limited ability in extracting the geometry information, which is still insufficient for modeling complex motions.The projection-based representation compresses the 3D space to 2D space, losing 3D geometric information for deformation learning. In contrast, our canonical-based method can fully leverage the 3D information in 3D space.

\section{Method}
\label{sec:method}
In this section, we present our 3D geometry-aware deformable Gaussian Splating solution for dynamic view synthesis, where an overview of our pipeline is illustrated in Fig.~\ref{fig:framework}. 
Given a set of images or monocular video of a dynamic scene with frames with corresponding time labels and known camera intrinsic and extrinsic parameters, our goal is to synthesize a novel view at any desired view at any desired time.  
Our method mainly consists of two core components: the Gaussian canonical field is used to learn the reconstruction of static scenes, while the deformation field is used to learn object deformation. First, we review the static 3D Gaussian splatting in \sref{Preliminary}. Then, we introduce the proposed Gaussian canonical field in \sref{Gaussian Canonical Field}, which consists of 3D Gaussian distributions and a geometry feature learning network. Next, in \sref{Deformation Field}, we propose a 3D geometry-aware deformation field to learn transformations for given time steps, which transform our canonical 3D Gaussian distributions to corresponding times. In \sref{Rasterization}, we explain the process of rendering images from transformed 3D Gaussian distributions. Finally, we present our losses and density control modifications in \sref{Optimization}.

\subsection{Preliminary}
\label{Preliminary}
3D-GS~\cite{3Dgaussian} represents the scene with sparse 3D Gaussians distributions. Each Gaussian has an anisotropic covariance $\mathbf{\Sigma}\in\mathbb{R}^{3\times3}$ and a mean value $\mu\in\mathbb{R}^3$:
\begin{equation}
    \mathbf{G}(\mathbf{x})=e^{-\frac{1}{2}(\mathbf{x}-\mu)^{\top}\mathbf{\Sigma}^{-1}(\mathbf{x}-\mu)}.
\end{equation}
The covariance matrix $\mathbf{\Sigma}$ can be decomposed into a scaling matrix $\mathbf{S}\in\mathbb{R}^{3\times3}$ and a rotation matrix $\mathbf{R} \in \mathrm{SO}(3)$. This ensures that the covariance matrix is positive semi-definite, while reducing the learning difficulty of 3D Gaussians:
\begin{equation}
\mathbf{\Sigma}=\mathbf{R}\mathbf{S}\mathbf{S}^{\top}\mathbf{R}^{\top}.
\end{equation}
To render an image from a designated viewpoint, the covariance matrix $\mathbf{\Sigma}^{\prime}$ in camera coordinates can be calculated by giving a viewing transformation $\mathbf{W}$, followed by~\cite{Zwicker_Pfister_Baar_Gross_2001}:
\begin{equation}
    \mathbf{\Sigma}^{\prime}=\mathbf{J} \mathbf{W}\mathbf{\Sigma} \mathbf{W}^{\top} \mathbf{J}^{\top},
\end{equation}
where $\mathbf{J}$ is the Jacobian of the affine approximation of the projective transformation, and $\mathbf{W}$ is the world to camera transformation matrix.

Each Gaussian is parameterized into the following attributes: position $\mathbf{x}\in\mathbb{R}^3$, color defined by spherical harmonics coefficients $\mathbf{c}\in\mathbb{R}^k$, rotation $\mathbf{r}\in\mathbb{R}^4$, scale $\mathbf{s}\in\mathbb{R}^3$, and opacity $o\in\mathbb{R}$. Point-based $\alpha$-blending and volumetric rendering like NeRF~\cite{mildenhall2020_nerf_eccv20} essentially share the same image formation model for the splatting process. Specifically, the color $\mathbf{C}$ of each pixel is influenced by the related Gaussians:
\begin{equation}
\mathbf{C}=\sum_{i=1}^N \mathbf{T}_i\alpha_i\mathbf{c}_i,
\end{equation}
where $\alpha_i$ represents the density of the Gaussian point computed by a Gaussian with covariance $\mathbf{\Sigma}$ multiplied by its opacity.

\subsection{Gaussian Canonical Field}
\label{Gaussian Canonical Field}
% In this section, we introduce how to define Gaussian parameters in canonical space and how do we utilize the structural properties naturally possessed by point clouds.
In this section, we first reconstruct a static scene in canonical space. Then, we propose a geometric branch, which enables geometry feature learning of the 3D Gaussian distributions for the subsequent deformation field.

\noindent\textbf{Gaussian parameters.}
Similar to 3D-GS~\cite{3Dgaussian}, each Gaussian in the canonical space is characterized by position, color, scale, and opacity. Note that for rotation, we are inspired by ~\cite{zhou2019continuity} to use a continuous 6D rotation representation. Compared with the quaternion representation used in 3D-GS, the 6D rotation representation can benefit our method in estimating the deformation of each Gaussian from canonical space to time-space, especially in helping the neural networks to learn smooth rotation variation from time to time. Specifically, we set learnable parameter $\left[a_1,a_2\right]$ for each Gaussian to denote its rotation in canonical space, where $a_1$ and $a_2$ are the column vectors of three rows, respectively. They are initialized to $[1~0~0]^\top$ and $[0~1~0]^\top$, corresponding precisely to the identity rotation matrix. The mapping from this 6D representation vector to $\mathrm{SO}(3)$ matrix is defined as ~\cite{zhou2019continuity}:
\begin{equation}
f_\text{V2M}\left( {\left[ {\begin{array}{*{20}{c}}
|&|\\
{{a_1}}&{{a_2}}\\
|&|
\end{array}} \right]} \right) = \left[ {\begin{array}{*{20}{c}}
|&|&|\\
{{b_1}}&{{b_2}}&{{b_3}}\\
|&|&|
\end{array}} \right],
\end{equation}

\begin{equation}
{b_i} = {\left[ {\left\{ {\begin{array}{*{20}{c}}
{\mathcal{N}({a_1})}&{{\rm{if }}~i = 1}\\
{\mathcal{N}({a_2} - ({b_1} \cdot {a_2}){b_1})}&{{\rm{if }}~i = 2}\\
{{b_1} \times {b_2}}&{{\rm{if }}~i = 3}
\end{array}} \right.} \right]^{\top}},
\end{equation}
where $\mathcal{N}(\cdot)$ denotes a normalization function. ``$\cdot$'' represents the inner product of a vector and ``$\times$'' represents vector cross product. $\text{V2M}$ in $f_\text{{V2M}}$ means the transform from 6D vector to rotation matrix.

\noindent\textbf{Geometry feature learning.}
To capture the local geometric structure of the canonical scene, we regard the 3D Gaussian as the 3D point cloud, $i.e.$, we only use the 3D coordinates of the 3D Gaussian. In order to handle a large number of point clouds, we leverage a simple two-branch structure: the geometric branch learns local features of point clouds across different receptive fields, while the identity branch preserves the independent point-level features at high resolution. By integrating the geometric branch and identity branch, we can efficiently obtain point-level features at high resolution while embedding the local geometric information of the point cloud. 

The geometric branch leverages the sparse convolution~\cite{liu2015sparse} on the sparse voxels to extract local geometric features at different receptive fields. Given the point cloud $\textbf{P}\in\mathbb{R}^{N \times3}$, we first transform the high-resolution point clouds into low-resolution voxels by dividing the space through fixed grid size $s$:
\begin{equation}
    \textbf{V} = \operatorname{floor}(\textbf{P} / s),
\end{equation}
where the size of $\textbf{V}$ is $M\times3$ and $M$ is the number of voxels. Then, we construct a sparse 3D U-Net by stacking a set of sparse convolutions with a skip connection. Taking $\textbf{V}$ as input, we perform sparse 3D U-Net to aggregate local features (dubbed as $\textbf{F}_{v}\in\mathbb{R}^{M\times C}$) of the point clouds.

The identity branch uses a multi-layer perception (MLP) to map the 3D coordinate of the point cloud into the embedding space (dubbed as $\textbf{F}_{p}\in\mathbb{R}^{N\times C}$) to maintain the independence of point features. To accurately characterize the local geometric structure of the canonical scene, we fuse the voxel features with local information onto point features. Specifically, we transform the voxel feature $\textbf{F}_{v}$ back to the corresponding points to obtain point-level features $\textbf{F}_{p}^{'}\in\mathbb{R}^{N\times C}$ by assigning the voxel features to the corresponding points within it. Finally, we concatenate $\textbf{F}_{p}^{'}$ and $\textbf{F}_{p}$ to obtain the fused point-level feature followed by an MLP layer as:
\begin{equation}
\textbf{F}_{\text{fuse}}=\operatorname{MLP}(\operatorname{Concat}(\textbf{F}_{p}^{'}, \textbf{F}_{p})).
\end{equation}

%For subsequent To encode the  features, we simple devoxelize the voxel-level features to point-level

\subsection{Deformation Field}
\label{Deformation Field}
In this section, we propose a deformation field that estimates the deformation of each 3D Gaussian in the canonical space based on a given time $t$. 

\noindent\textbf{Deformation estimation.}
We adopt an MLP as the decoder $\mathcal{G}_\Phi$, which takes the geometry feature learned from the geometry branch in the Gaussian canonical field, the position of each Gaussian, and timestamp as input, outputs the deformation of each Gaussian from canonical space to time $t$, including position deformation $\Delta\mathbf{x_t}\in\mathbb{R}^3$, rotation deformation  $\Delta\mathbf{r_t}\in\mathbb{R}^6$ and scale deformation $\Delta\mathbf{s_t}\in\mathbb{R}^3$:
\begin{equation}
\Delta\mathbf{x_t},\Delta\mathbf{r_t},\Delta\mathbf{s_t}=\mathcal{G}_\Phi(\textbf{F}_{\text{fuse}},\gamma(\textbf{x}),\gamma(t)),
\end{equation}
where $\gamma(\cdot)$ denotes the positional encoding in NeRF~\cite{mildenhall2020_nerf_eccv20}, which maps a one dimension signal from $\mathbb{R}$ into a higher dimensional space $\mathbb{R}^{2L}$:
\begin{gather}
\begin{split}
\gamma(p)=~ & (\sin{(2^0\pi p)}, \cos{(2^0\pi p)}, \\
& ...,\\
& \sin{(2^{L-1}\pi p)}, \cos{(2^{L-1}\pi p)}).
\end{split}
\end{gather}

Note that we set the color parameters $\mathbf{c}$ and opacity $o$ of canonical 3D Gaussian distributions constant over time. These two factors are highly related to the physical properties of the Gaussian distributions, and we want each distribution to represent the same object area over the timeline.

\noindent\textbf{Transformation.}
Using the estimated deformation for time $t$ above, we could transform the 3D Gaussian distributions to current time by
\begin{gather}
\begin{split}
 \mathbf{x}_t & =~\mathbf{x}+\Delta\mathbf{x_t}, \\
 \mathbf{s}_t & =~\mathbf{s}+\Delta\mathbf{s_t}, \\
 \mathbf{r}_t & =~f_\text{V2M}(\Delta\mathbf{r_t})\times f_\text{V2M}(\mathbf{r}). \\
\end{split}
\end{gather}

\subsection{Rasterization}
\label{Rasterization}

Once we have completed preparing the attributes of each Gaussian$(\mathbf{x}_t, \mathbf{c}, \mathbf{r}_t, \mathbf{s}_t, o)$, we use the differentiable tile rasterizer~\cite{3Dgaussian} to render the image at any desired viewpoint at this timestamp:
\begin{equation}
\hat{\mathbf{C}_t}=Rasterizer(\mathbf{x}_t, \mathbf{c}, \mathbf{r}_t, \mathbf{s}_t, o, \mathbf{K}, [\mathbf{R}|\mathbf{T}]),
\end{equation}
where $\mathbf{K}$ and $[\mathbf{R}|\mathbf{T}]$ represent the camera's intrinsic and extrinsic parameters, respectively.

\subsection{Optimization}
\label{Optimization}
To optimize the model, we use the photometric loss, and a motion loss, and also adapt the density control from 3D-GS~\cite{3Dgaussian} with our modifications.

\noindent\textbf{Photometric loss.}
The photometric loss consists of the $L_1$ loss and structural similarity loss $L_{D-SSIM}$ between the rendered image $\hat{\mathbf{C}}_t$ and ground truth image $\mathbf{C}_{t}$.
\begin{equation}
L_{photo}=(1-\lambda)L_1+\lambda L_{D-SSIM}.
\end{equation}

\begin{table*}[t!]
\setlength{\tabcolsep}{3pt} %
\centering
\caption{\small{
Quantitative comparison between our method and competing methods on the D-NeRF dataset. The best results are highlighted in bold.
% Comparison of our method with others on PSNR and MS-SSIM (higher is better) on real scene datasets. 
% \tabfirst{Red} text indicates the best and \tabsecond{blue} text indicates the second best result.
}}
\label{table:dnerf}
%\vspace{-0.5\baselineskip}
\vspace{-9pt}
\resizebox{0.95\textwidth}{!}{%
\footnotesize{
\begin{tabular}{ccccccccccccccccc}
\toprule
& \multicolumn{3}{c}{ Hell Warrior } & \multicolumn{3}{c}{ Mutant } & \multicolumn{3}{c}{ Hook } & \multicolumn{3}{c}{ Bouncing Balls } \\
Method & PSNR$\uparrow$ & SSIM$\uparrow$ & LPIPS$\downarrow$ & PSNR$\uparrow$ & SSIM$\uparrow$ & LPIPS$\downarrow$ & PSNR$\uparrow$ & SSIM$\uparrow$ & LPIPS$\downarrow$ & PSNR$\uparrow$ & SSIM $\uparrow$ & LPIPS$\downarrow$ \\
% \cmidrule(lr){1-1} \cmidrule(lr){2-3} \cmidrule(lr){4-5} \cmidrule(lr){6-7} \cmidrule(lr){8-9}  \cmidrule(lr){10-11}
\hline 3D-GS~\cite{3Dgaussian} & 15.3924 & 0.8776 & 0.1300 & 21.7554 & 0.9359 & 0.0575 & 18.6933 &	0.8733 & 0.1144 &  22.5575 & 0.9485 & 0.0647 \\
% T-NeRF & 23.19 & 0.93 & 0.08 & 30.56 & 0.96 & 0.04 & 27.21 & 0.94 & 0.06 & 32.01 & 0.97 & 0.04 \\

D-NeRF \cite{pumarola2021_dnerf_cvpr21} & 25.0293 & 0.9506 & 0.0691 & 31.2900 & 0.9739 & 0.0268 & 29.2567 & 0.9650 & 0.1174 & 38.9300 & 0.9900 & 0.1031 \\

TiNeuVox-B\cite{fang2022_TANV_arxiv} & 28.2058 & 0.9661 & 0.0631 & 33.9029 & 0.9771 &	0.0301 & 31.7929 & 0.9718 &	0.0436 & 40.8536 &	0.9913 & 0.0401 \\

NDVG \cite{guo2022_NDVG_arxiv} & 26.4933 & 0.9600 & 0.0670 & 34.4131 & 0.9801 & 0.0270 & 30.0009 & 0.9626 & 0.0463 & 37.5157 & 0.9874 & 0.0751 \\

FDNeRF \cite{Guo2023_Forward_ICCV} & 27.7120 & 0.9665 & 0.0508 & 34.9727 & 0.9810 & 0.0312 & 32.2867 & 0.9756 & 0.0388 & 40.0191 & 0.9912 & 0.0395 \\

4D-GS \cite{4DGaussian} & 28.1196 &	0.9730 & 0.0276 & 38.3411 &	0.9936 & 0.0062 & 33.1560 &	0.9810 & 0.0168 & 40.7418 &	0.9941 &	0.0105\\

Ours & \textbf{32.2712} & \textbf{0.9835} & \textbf{0.0164} & \textbf{41.4284} & \textbf{0.9969} & \textbf{0.0029} & \textbf{36.9647} & \textbf{0.9916} &	\textbf{0.0076} & \textbf{43.5929} & \textbf{0.9960} & \textbf{0.0061} \\

\hline & \multicolumn{3}{c}{ Lego } & \multicolumn{3}{c}{ T-Rex } & \multicolumn{3}{c}{ Stand Up } & \multicolumn{3}{c}{ Jumping Jacks } \\

Method & PSNR$\uparrow$ & SSIM$\uparrow$ & LPIPS$\downarrow$ & PSNR$\uparrow$ & SSIM$\uparrow$ & LPIPS$\downarrow$ & PSNR$\uparrow$ & SSIM$\uparrow$ & LPIPS$\downarrow$ & PSNR$\uparrow$ & SSIM $\uparrow$ & LPIPS$\downarrow$ \\

\hline 3D-GS~\cite{3Dgaussian} & 23.0991 & 0.9329 &	0.0567 & 25.7496 & 0.9567 &	0.0474 & 19.3779 &	0.9200 & 0.0909 & 20.7163 &	0.9227 & 0.0980 \\
% T-NeRF & 23.82 & 0.90 & 0.15 & 30.19 & 0.96 & 0.13 & 31.24 & 0.97 & 0.02 & 32.01 & 0.97 & 0.03 \\

D-NeRF \cite{pumarola2021_dnerf_cvpr21} & 21.6427 & 0.8394 & 0.1654 & 31.7568 & 0.9767 & 0.0396 & 32.7992 & 0.9818 & 0.0215 & 32.8031 & 0.9810 & 0.0373 \\

TiNeuVox-B\cite{fang2022_TANV_arxiv} & 25.1748 & 0.9217 & 0.0689 & 32.7750 & 0.9783 &	0.0307 & 36.2031 &	0.9859 & 0.0199 & 34.7390 &	0.9823 & 0.0328 \\

NDVG \cite{guo2022_NDVG_arxiv} & 25.0416 & 0.9395 & 0.0534 & 32.6229 & 0.9781 & 0.0330 & 33.2158 & 0.9793 & 0.0302 & 31.2530 & 0.9737 & 0.0398 \\

FDNeRF \cite{Guo2023_Forward_ICCV} & 25.2700 & 0.9390 & 0.0460 & 30.7068 & 0.9731 & 0.0368 & 36.9107 & 0.9878 & 0.0188 & 33.5521 & 0.9812 & 0.0329 \\

4D-GS \cite{4DGaussian} & 25.4024 & 0.9434 &	0.0377 & 33.3912 &	0.9869 & 0.0130 & 38.2610 &	0.9923 & 0.0071 & 35.6656 &	0.9882 &	0.0159\\

Ours & \textbf{25.4411} & \textbf{0.9474} & \textbf{0.0329} & \textbf{39.0285} & \textbf{0.9952} & \textbf{0.0052} & \textbf{42.2101}	& \textbf{0.9966} &	\textbf{0.0028} & \textbf{37.9604} & \textbf{0.9928} & \textbf{0.0088} \\

\bottomrule %

\end{tabular}
}}
\vspace{-0.8\baselineskip}
\end{table*}

\begin{table*}[t!]
\setlength{\tabcolsep}{3pt} %
\centering
\caption{\small{
Quantitative comparison between our method and competing methods on the HyperNeRF dataset.The best results are highlighted in bold.
% Comparison of our method with others on PSNR and MS-SSIM (higher is better) on real scene datasets. 
% \tabfirst{Red} text indicates the best and \tabsecond{blue} text indicates the second best result.
}}
\label{table:hyper_detailed}
%\vspace{-0.5\baselineskip}
\vspace{-8pt}
\resizebox{0.7\textwidth}{!}{%
\footnotesize{
\begin{tabular}{cccccccccccc}
\toprule
& \multicolumn{2}{c}{ Chicken } & \multicolumn{2}{c}{ 3D Printer } & \multicolumn{2}{c}{ Broom } & \multicolumn{2}{c}{ Peel Banana } \\
Method & PSNR$\uparrow$ & MS-SSIM$\uparrow$ & PSNR$\uparrow$ & MS-SSIM$\uparrow$ &PSNR$\uparrow$ & MS-SSIM$\uparrow$&PSNR$\uparrow$ & MS-SSIM$\uparrow$\\
% \cmidrule(lr){1-1} \cmidrule(lr){2-3} \cmidrule(lr){4-5} \cmidrule(lr){6-7} \cmidrule(lr){8-9}  \cmidrule(lr){10-11}
\hline TiNeuVox\cite{fang2022_TANV_arxiv} & 28.2861 & \textbf{0.9474} & 22.7514 &	0.8392 &		21.2682&	0.6832	&	24.5136&	0.8743\\

NDVG \cite{guo2022_NDVG_arxiv} & 27.0536 & 0.9390 & 22.4196	&0.8389	&	21.4658&	0.7028	&	22.8204&	0.8279\\

FDNeRF \cite{Guo2023_Forward_ICCV} & 27.9627 & 0.9438&\textbf{22.8027}	&\textbf{0.8453}	&	\textbf{21.9091}&	\textbf{0.7154}	&	24.2515	&0.8645\\

3D-GS~\cite{3Dgaussian} &20.8915 & 0.7426 & 18.3991 &	0.6114 &	20.3953 &	0.6598 & 20.5654 &	0.8094\\

Ours &  \textbf{28.5342} & 0.9331 &22.0403&	0.8098&		20.8994&0.5241&		\textbf{25.5785}&	\textbf{0.9067}\\

\bottomrule %

\end{tabular}
}}
%\vspace{-\baselineskip}
\vspace{-15pt}
\end{table*}

\begin{table}[t!]
\setlength{\tabcolsep}{3pt} %
\centering
\caption{\small{
Quantitative comparison on HyperNeRF dataset: Average on Cut Lemon, Chicken, 3D Printer, and Split Cookie. The best results are highlighted in bold.
% Comparison of our method with others on PSNR and MS-SSIM (higher is better) on real scene datasets. 
% \tabfirst{Red} text indicates the best and \tabsecond{blue} text indicates the second best result.
}}
\label{table:hnerf}
%\vspace{-0.5\baselineskip}
\vspace{-5pt}
% \resizebox{\textwidth}{!}{%
\footnotesize{
\begin{tabular}{ccccc}
\toprule
Method  & PSNR$\uparrow$ & SSIM$\uparrow$ & LPIPS$\downarrow$ \\
% \cmidrule(lr){1-1} \cmidrule(lr){2-3} \cmidrule(lr){4-5} \cmidrule(lr){6-7} \cmidrule(lr){8-9}  \cmidrule(lr){10-11}
\hline TiNeuVox-B \cite{fang2022_TANV_arxiv} & 27.16 & 0.76 & 0.40 \\

3D-GS \cite{3Dgaussian} & 21.26 & 0.69 &0.40 \\

4D-GS \cite{4DGaussian} & 26.98 & 0.78 & 0.31\\

Ours & \textbf{27.52} & \textbf{0.80} & \textbf{0.25} \\

\bottomrule %

\end{tabular}
}
%\vspace{-\baselineskip}
\vspace{-12pt}
\end{table}

\noindent\textbf{Regularization.}
We accept the fact that in a scene, the proportion of dynamic points is much smaller than that of static points, and the motion amplitude at dynamic points is not too large. In other words, the point in a scene should be as static as possible,
\begin{equation}
% L^{motion}=\frac{1}{N}\left\|\mathbf{x_d}\right\|_1,
L_{motion}=\left\|\Delta\mathbf{x_t}\right\|_1.
\end{equation}

\noindent\textbf{Total loss.}
The total loss we used is defined as follows, 
\begin{equation}
L=L_{photo} + \omega L_{motion},
\end{equation}
where $\omega$ is a trade-off parameter to balance the components.

\noindent\textbf{Density control.}
\begin{figure}[!t]
	\centering
	\includegraphics[width=\columnwidth]{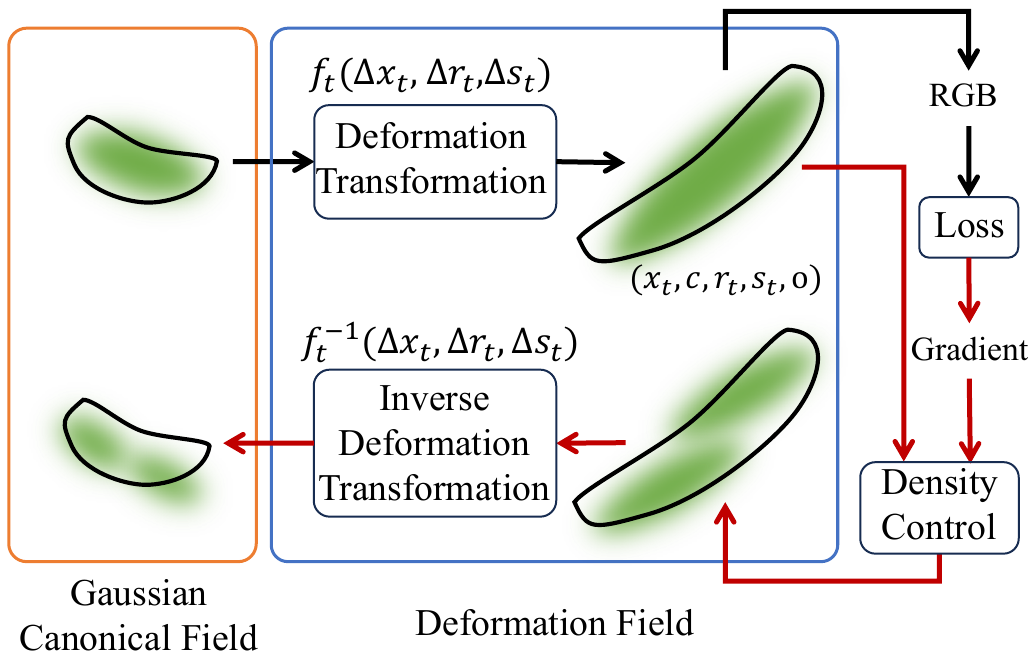}
	\vspace{-0.7cm}
	\caption{Our density control is designed for dynamic scenes. We control the densification of Gaussian distributions according to their transformed parameters at timestamp $t$ rather than parameters at canonical space.}
	\label{fig:density}
	\vspace{-18pt}
\end{figure}
3D-GS has shown that adaptive density control is essential in achieving high rendering performance. On the one hand, the Gaussians need to populate empty areas without geometric features. Thus, it simply creates a copy of the Gaussian for under-reconstructed regions. On the other hand, large Gaussians in regions with high variance need to be split into smaller Gaussians. We implement our method like 3D-GS but replace such Gaussians with two new ones, divide their scale by a factor of $\phi=1.6$, and initialize their position by using the original 3D Gaussian as a PDF for sampling.

Our method differs from 3D-GS in the following aspects. For 3D-GS, there only exists sets of Gaussians. However, in our case, we initialize the Gaussians in the canonical space, then estimate the deformations of these Gaussians, and transform their attributes into a timestamp space. As shown in \fref{fig:density}, we use the Gaussians at the current moment to render the image. Therefore, we determine whether the Gaussians need to conduct density control by the current attributes (like scale) at the current timestamp rather than the canonical attributes. Afterward, we inverse the transformation of the split/cloned Gaussian back to the canonical space.
\section{Experiments}

\subsection{Dataset}
In the paper, we use both synthetic and real datasets for evaluating our method. The synthetic dataset D-NeRF~\cite{pumarola2021_dnerf_cvpr21} contains 8 dynamic scenes, including Hell Warrior, Mutant, Hook, Bouncing Balls, Lego, T-Rex, Stand Up, and Jumping Jacks. The real dataset proposed by HyperNeRF \cite{park2021hypernerf}, including interp-cut-lemon, interp-cut-lemon1, vrig-chicken, vrig-3dprinter, misc-split-cookie, and misc-split-cookie. Following previous works~\cite{3Dgaussian}, we report three evaluation metrics, including Peak Signal-to-Noise Ratio (PSNR), Structural Similarity (SSIM), and Learned Perceptual Image Patch Similarity (LPIPS)~\cite{zhang2018_lpips_CVPR}.

\begin{figure*}[!t]
\centering
\includegraphics[width=0.9\textwidth]{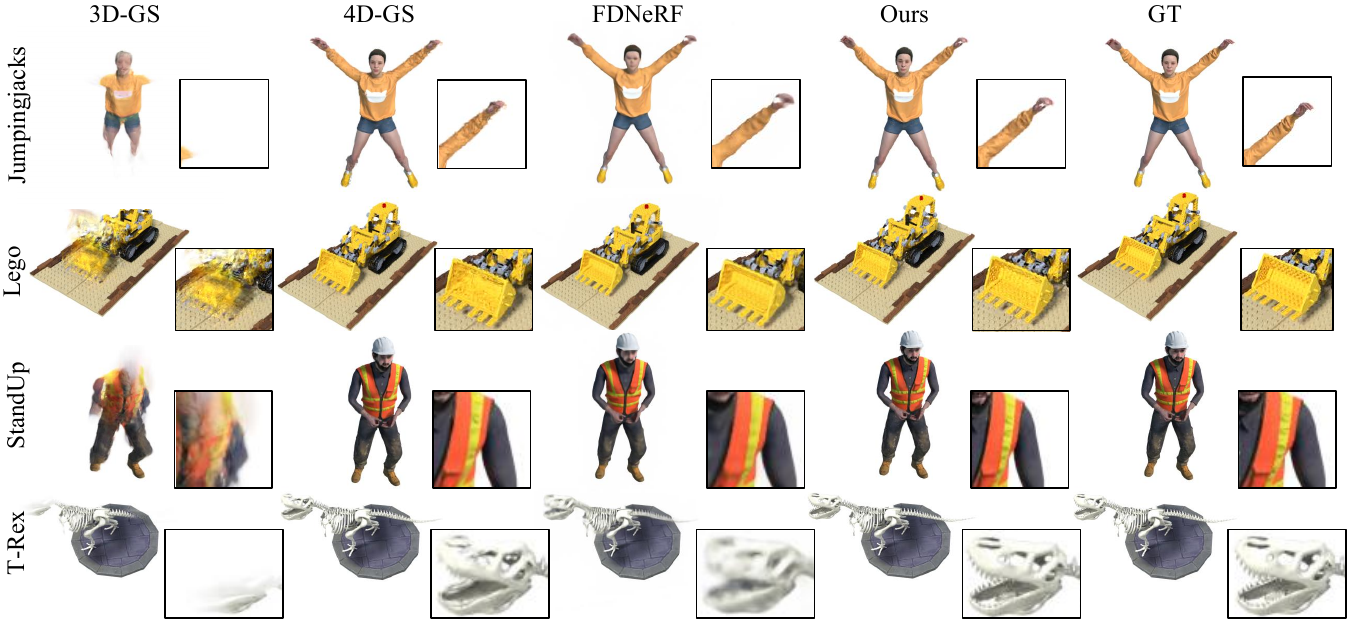}
\vspace{-2pt}
\caption{Qualitative comparisons between baselines and our method on the synthetic dataset.}
%\label{fig:synthetic}
\label{fig:syn-quality}
%\vspace{-\baselineskip}
\vspace{-15pt}
\end{figure*}

\begin{figure}[!t]
\centering
\includegraphics[width=0.48\textwidth]{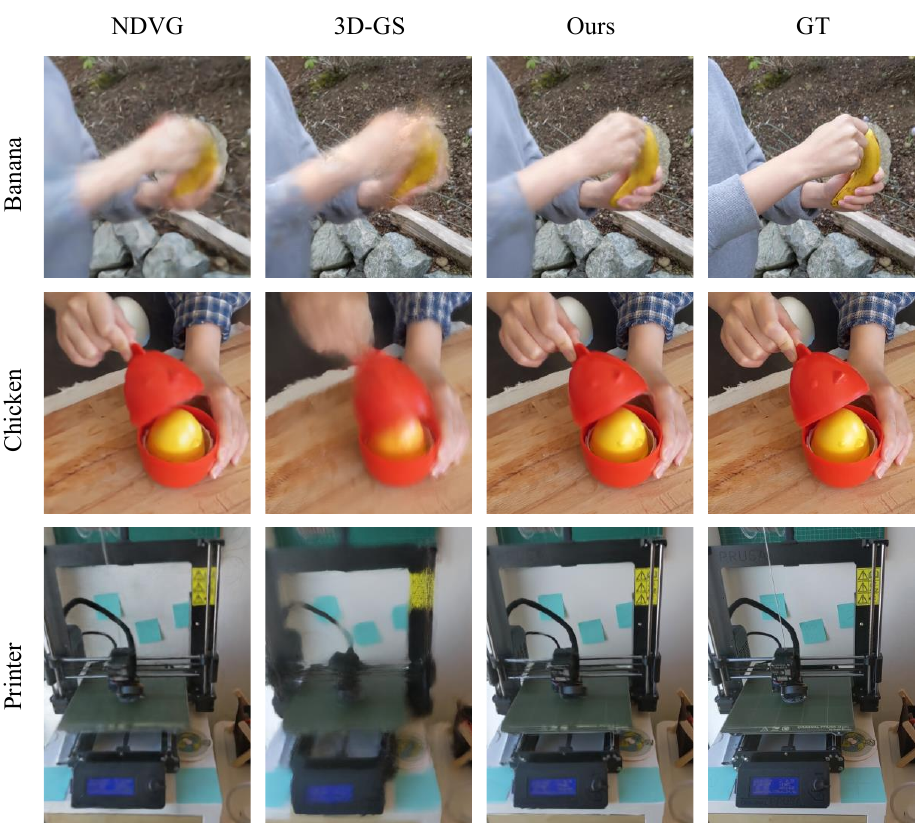}
\vspace{-2pt}
\caption{Qualitative comparisons between baselines and our method on the HyperNeRF real dataset\cite{park2021hypernerf}.}
\label{fig:real-quality}
\vspace{-15pt}

\end{figure}

\subsection{Implementation Details}
Our implementation is based on 3D-GS~\cite{3Dgaussian}. We trained a total of 40000 iterations, with the first 3000 iterations only optimizing static scenes, and then adding deformation fields to optimize dynamic scenes. The learning rate of our network takes an exponential decay from 8e-4 to 1.6e-6 with the Adam optimizer. Moreover, we use a 2-layer MLP with a width of 64 for the front point feature extraction, and a 3-layer MLP with a width of 64 for the back point feature fusion. Then 5 layers MLP with width 256 and skip connection is used for a decoder. For the positional encoding process, we use $L=10$ for position $\mathbf{x}$ and $L=6$ for timestamp $t$. For the D-NeRF dataset, which does not provide point clouds, we randomly initialize 150000 points. Meanwhile, for the HyperNeRF dataset, we use the point cloud provided in its dataset as the initial point cloud. All the experiments are tested on a single RTX 4090 GPU.

\subsection{Quantitative Results}

\noindent\textbf{Synthetic scenes.} We compare our method with recent state-of-the-art methods in the field, including 3D-GS, D-NeRF, TiNeuVox, NDVG, FDNeRF, and 4D-GS on the D-NeRF Dataset. As shown in~\Tref{table:dnerf}, we list the results of each scene. It can be observed that our method is significantly better than other methods in terms of all three metrics for physical canonical-based methods. On average, our method significantly improves PSNR compared with static Gaussian, 3D-GS. The computational costs are: training time around 2h (avg. on D-NeRF dataset), render FPS 12 (fixed viewpoint), model size (34MB points cloud + 14MB network). Since it inherently cannot model the deformation of the dynamic scene, 3D-GS performs poorly in dynamic view synthesis. 

% For the recent advanced dynamic method K-Planes, our method also shows better qualitative rendering quality with obvious advantages.

% \input{tables/hypernerf_detailed}
\noindent\textbf{Real scenes.} We further compare our method with some highly related works on the real scene dataset proposed by~\cite{park2021hypernerf}. We have shown the detailed results on chicken, 3D printer, broom, and peel banana in~\Tref{table:hyper_detailed}, and the average result on cut lemon, chicken, 3d printer, split cookie in~\Tref{table:hnerf}. It can be observed that our method achieves good performance compared with other state-of-the-art methods. Compared with synthetic datasets, real datasets are more challenging due to the narrow camera viewing range and pose ambiguity. The quantitative results can demonstrate the effectiveness of the proposed method in real scenes.

\subsection{Visualization Results}

\textbf{Visual comparison.} In addition to quantitative results, we also provide visualization results of different methods to demonstrate the superiority of our method. For better comparison, we show the rendered images of each synthetic scene from the same viewpoint in~\fref{fig:syn-quality}. By comparing the visualization results of different methods, it is shown that the rendered images by our method are closer to the ground truth images, indicating that our method can recover accurate and detailed images. In addition, we provide visualization results of the real scenes in~\fref{fig:real-quality}. Compared with TiNueVox \cite{fang2022_TANV_arxiv}, our method can recover the detailed structure of dynamic objects, like chicken and banana. 

%\noindent\textbf{Varying time.} To show the effect of our method to model dynamic scenes, we visualize the images from different timestamps. As shown in~\fref{fig:varytime}, the changes of the rendered objects are continuous under different timestamps. The visualization results show that our method can effectively capture the deformation of dynamic scenes.

\begin{figure}[!t]
	\centering
	\includegraphics[width=\columnwidth]{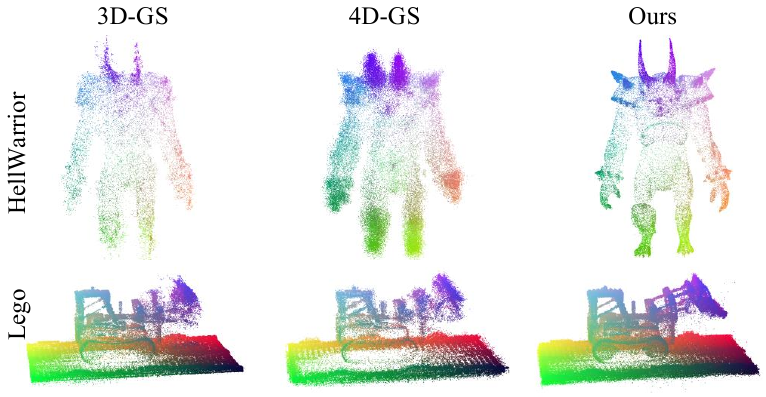}
	%\vspace{-1\baselineskip}
	\vspace{-15pt}
	\caption{Visulization of learned Gaussian. Colored with position coordinates}
	\label{fig:point}
	%\vspace{-0.4\baselineskip}
	\vspace{-13pt}
\end{figure}

\noindent\textbf{Gaussian visualization.} To verify the effectiveness of our method, we show the 3D point cloud of the 3D Gaussian. Specifically, we only use the 3D coordinates of the 3D Gaussian. As shown in~\fref{fig:point}, we provide the point clouds of different methods on the synthetic dataset, including 3D-GS~\cite{3Dgaussian}, 4D-GS~\cite{4DGaussian}, and ours. Note that the color of the point cloud is generated by 3D coordinates. Since 3D-DS cannot model dynamic scenes, the quality of the point cloud is poor. Comparing 4D-GS with ours, it can be observed that the point cloud of our method has a clear local geometric structure.

\begin{figure}[!t]
	\centering
	\includegraphics[width=\columnwidth]{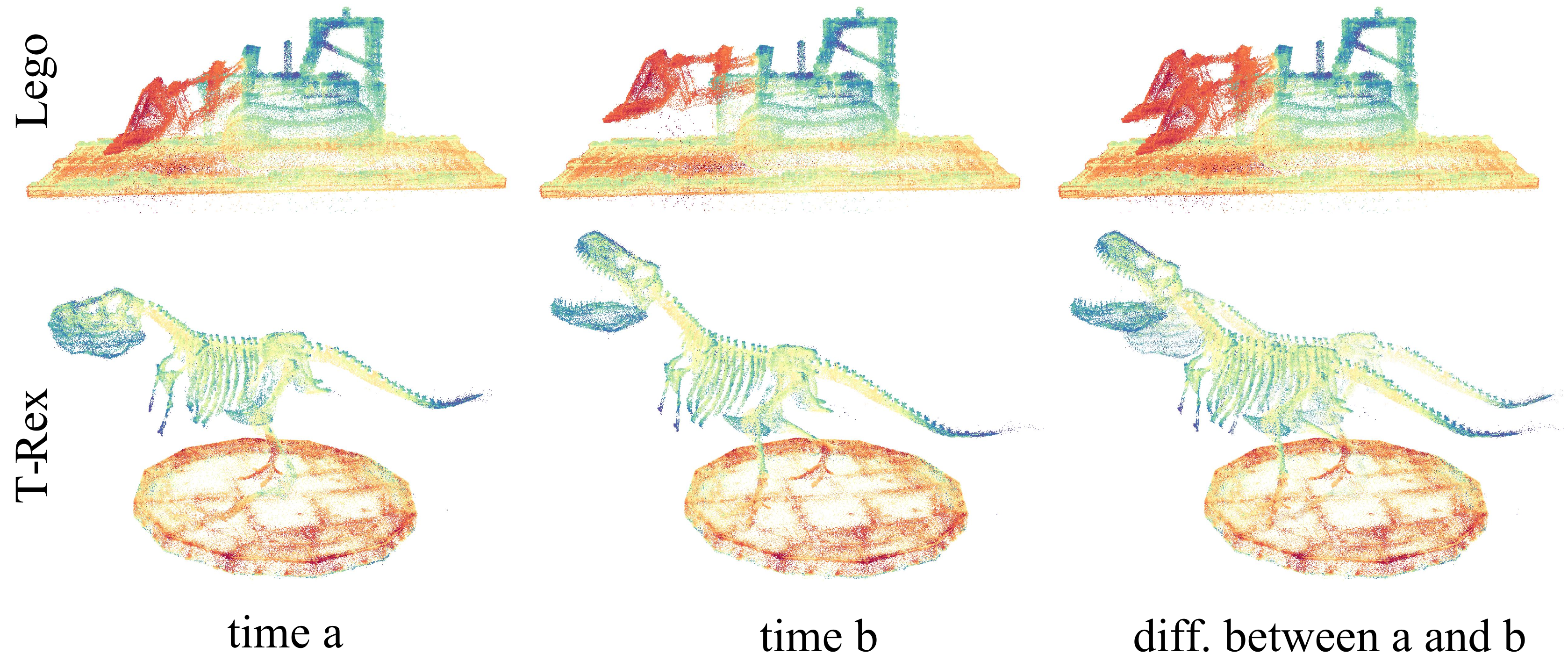}
	%\vspace{-0.8\baselineskip}
	\vspace{-15pt}
	\caption{Visulization of learned geometry-aware features.}
	\label{fig:feature}
	%\vspace{-0.9\baselineskip}
	\vspace{-5pt}
\end{figure}

\subsection{Ablation Study}

We conduct ablation studies on the synthetic dataset $(800 \times 800)$ to verify the effectiveness of our proposed components. In \Tref{table:ablation}, vanilla model is a simple MLP model without our components.

\noindent\textbf{Effect of geometric-aware features.} To learn the geometric information of the object in our Gaussian canonical field, we voxelize the 3D Gaussian distributions and extract geometric aware features using our 3D U-Net. To demonstrate the effectiveness of this design, we test our method with geometric branch blocks and leave others unchanged. In \Tref{table:ablation}, ours full has a clear advantage over w/o geo. feat., and our geometry branch plays the most important role among the components studied in the ablations. 

In \fref{fig:feature}, we visualize the learned geometric-aware features. We color the point clouds with the learned features, and it shows meaningful geometric information. Interestingly, we can see an obvious difference in the learned features between the moving objects (bucket of the lego and the t-rex body) and the static objects (body of the lego and the ground in t-rex). Also, our geometric-aware features reflect the local geometric structure. For example, the spines of the bones on the t-rex tail have similar features, and the smooth part of the tail bones have other patterns.

\noindent\textbf{Different geometric features.}
We use the PointNet-like architecture and plane projection (2D CNN) to conduct experiments. Compared with the results (dubbed as ``PointNet feat.'' and ``Plane feat.'') in \Tref{table:ablation}, it can be observed that our method achieves significant performance gains.
\begin{table}[t!]
\setlength{\tabcolsep}{3pt} %
\centering
\caption{\small{
\textbf{Ablation Study.} 
Ablation studys in terms of average PSNR, SSIM, and LPIPS. The best results are highlighted in bold.
% Comparison of our method with others on PSNR and MS-SSIM (higher is better) on real scene datasets. 
% \tabfirst{Red} text indicates the best and \tabsecond{blue} text indicates the second best result.
}}
\label{table:ablation}
\vspace{-0.5\baselineskip}
\resizebox{0.8\columnwidth}{!}{%
\footnotesize{
\begin{tabular}{ccccc}
\toprule
Method  & PSNR$\uparrow$ & SSIM$\uparrow$ & LPIPS$\downarrow$ \\
% \cmidrule(lr){1-1} \cmidrule(lr){2-3} \cmidrule(lr){4-5} \cmidrule(lr){6-7} \cmidrule(lr){8-9}  \cmidrule(lr){10-11}
\hline w/o geo. feat. & 37.5757 &	0.9841 & 0.0173\\

w/o 6D rotation &37.8750 &	0.9851 &	0.0154\\

canonical DC & 37.8026 & 0.9847 & 0.0166 \\

vanilla & 35.2307 & 0.9793 & 0.0242 \\

PointNet feat. & 36.7353 & 0.9826 & 0.0184 \\

Plane feat. & 35.9054 & 0.9811 & 0.0212 \\

ours full& \textbf{38.0134} &	\textbf{0.9853} & \textbf{0.0153}\\

\bottomrule %

\end{tabular}
}}
\vspace{-20pt}
\end{table}

\noindent\textbf{6D representation.}
To study the effect of 6D representation of the rotation parameters of the 3D Gaussian, we conduct an experiment that replaces the 6D vector with quaternion $\mathbf{q}$ which is used in the original 3D-GS. To deform the 3D Gaussian in canonical space, our deformation field estimates a $\Delta \mathbf{q_t}$ and gets $\mathbf{q_t}=\mathbf{q}+\Delta \mathbf{q_t}$, using the quaternion add operation. In \Tref{table:ablation}, quaternion demonstrates an obvious performance drop, which proves the effectiveness of the 6D representation.

\noindent\textbf{Density control.}
In terms of density control, we test the setting that only uses the 3D Gaussian in canonical space without considering the transform 3D Gaussian at other timestamps. In \Tref{table:ablation}, canonical DC shows a performance drop, as the canonical 3D Gaussian alone cannot reflect the over/under reconstruction information at all timestamps for dynamic scenes.

\section{Conclusion}
\label{sec:conclusion}
In this paper, we have proposed a 3D geometry aware Gaussian Splatting solution for dynamic view synthesis. We addressed the limitations of existing approaches from two perspectives: 1) we introduced 3D sparse convolution to extract local structural information effectively and efficiently for deformation learning, and 2) we represented the dynamic scenes as a collection of deforming 3D Gaussian distributions, which are optimized to deform (move, rotate, scaling) over time. Experimental results across synthetic and real datasets demonstrate the superiority of our solution in dynamic view synthesis and 3D reconstruction. We plan to further investigate explicit motion modeling by exploiting the foreground and background motion segmentation cues.

% \noindent\textbf{Limitation: }

\section*{Acknowledgments}
We thank the area chairs and the reviewers for their insightful and positive feedback. We also appreciate the reference provided by Ziyi Yang's work. This work was supported in part by the National Science Fund of China (Grant Nos. 62271410, 62306238) and the Fundamental Research Funds for the Central Universities.

\clearpage
\maketitlesupplementary

%\section{Overview}
%\label{sec:Overview}

%\begin{abstract}
This supplementary material provides additional implementation details and experimental results. First, we provide the implementation details of our proposed method. Then, we provide additional experimental results in the form of visualization and discuss the limitations and impacts of our method. We conclude with discussions on future work. The source code, network model, and results will be released.
%\end{abstract}

% % 
% Having the supplementary compiled together with the main paper means that:
% % 
% \begin{itemize}
% \item The supplementary can back-reference sections of the main paper, for example, we can refer to \cref{sec:intro};
% \item The main paper can forward reference sub-sections within the supplementary explicitly (e.g. referring to a particular experiment); 
% \item When submitted to arXiv, the supplementary will already included at the end of the paper.
% \end{itemize}
% % 
% To split the supplementary pages from the main paper, you can use \href{https://support.apple.com/en-ca/guide/preview/prvw11793/mac#:~:text=Delete%20a%20page%20from%20a,or%20choose%20Edit%20%3E%20Delete).}{Preview (on macOS)}, \href{https://www.adobe.com/acrobat/how-to/delete-pages-from-pdf.html#:~:text=Choose%20%E2%80%9CTools%E2%80%9D%20%3E%20%E2%80%9COrganize,or%20pages%20from%20the%20file.}{Adobe Acrobat} (on all OSs), as well as \href{https://superuser.com/questions/517986/is-it-possible-to-delete-some-pages-of-a-pdf-document}{command line tools}.

\section{Implementation Details}
\label{sec:B}
\subsection{Loss Function}
We apply the photometric loss and regularization for our optimization:
\begin{equation}
L_{total}=L_{photo}+\omega L_{motion},
\end{equation}
\begin{equation}
L_{photo}=(1-\lambda)L_{rgb}+\lambda L_{D-SSIM},
\end{equation}
where $L_{rgb}$ is the $L_1$ loss and $L_{SSIM}$ is the structural similarity loss between the rendered image $\hat{\textbf{C}}_t$ and ground truth image $\textbf{C}_t$. 
Generally, within a dynamic scene, the proportion of dynamic points is much smaller than that of the static points. Thus the motion amplitude at dynamic points is not too large. We proposed to exploit this fact by introducing the motion regularization term  $L_{motion}=\left\|\Delta\mathbf{x_t}\right\|_1$. In our experiments, we set $\lambda=0.2$ and $\omega=0.01$.

\subsection{Network Architecture}
\begin{figure}[h]
\centering
\includegraphics[width=\columnwidth]{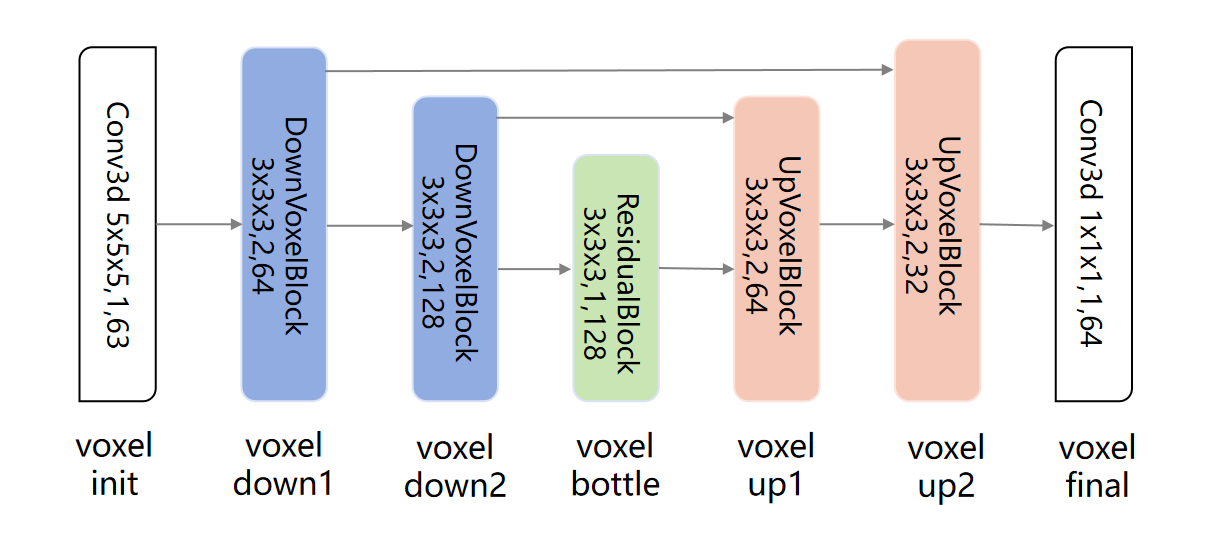}
%\vspace{-6pt}
\caption{Overall architecture of the geometric branch, which captures local geometric features using a 3D U-Net.}
\label{fig:unet}
\vspace{-6pt}
\end{figure}

\begin{figure}[h]
\centering
\includegraphics[width=\columnwidth]{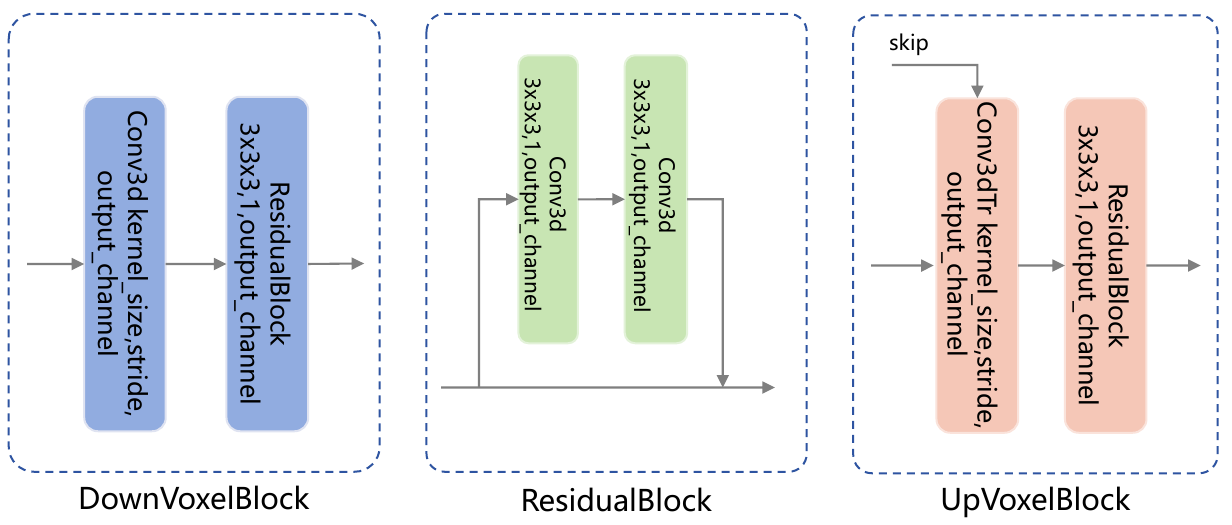}
%\vspace{-6pt}
\caption{Detailed structure of DownVoxelBlock, ResidualBlock, and UpVoxelBlock.}
\label{fig:unet_detail}
\vspace{-6pt}
\end{figure}

\begin{figure}[h]
\centering
\includegraphics[width=\columnwidth]{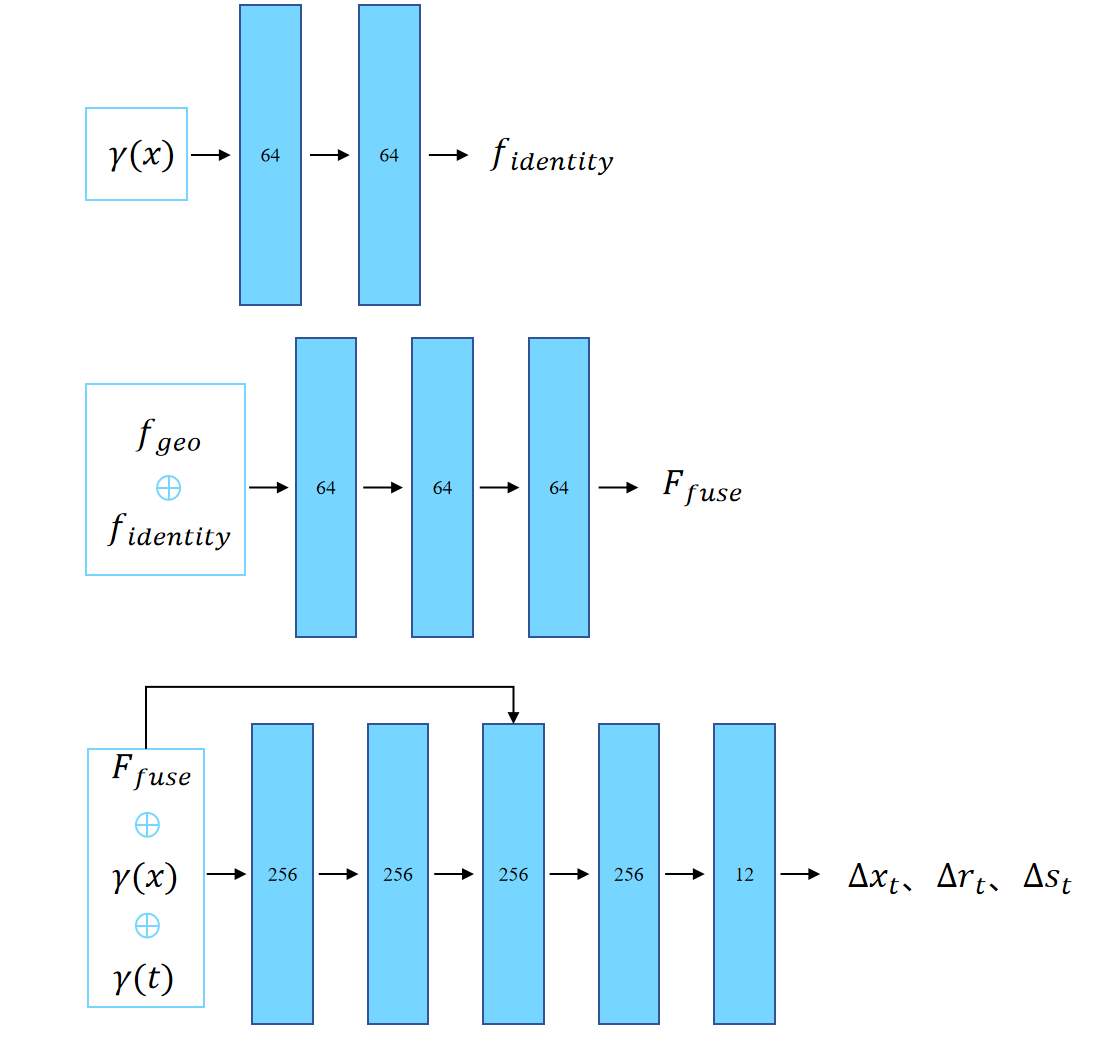}
%\vspace{-6pt}
\caption{Detailed structure of MLPs we have used in our method.}
\label{fig:net}
\vspace{-6pt}
\end{figure}

Here, we introduce the network architecture adopted in our method. The Gaussian Canonical Field consists of two branches: the geometric branch and the identity branch. As shown in~\fref{fig:unet}, the geometric branch takes the position of voxel points as input and outputs the geometrical features $f_{geo}$. It is roughly composed of three parts, namely DownVoxelBlock, ResidualBlock, and UpVoxelBlock. The specific structures of these three parts are shown in~\fref{fig:unet_detail}. 
For the identity branch, we use a simple MLP to get the embedding features $f_{identity}$, which maintains the independence of point features. Then we concatenate the features from the geometric branch and the identity branch, and pass them into another MLP to get fused features $\textbf{F}_{\text{fuse}}$. Finally, we take the fused features $\textbf{F}_{\text{fuse}}$, position of Gaussians $\gamma(x)$ and time $\gamma(t)$ into a decoder to get the deformations of position, rotation, and scale from the canonical space to time space. In~\fref{fig:net}, we demonstrate the specific structure of MLPs. Additionally, the intermediate hidden layers are shown in blue, the number inside each block signifies the vector's dimension. All layers are standard fully-connected layers, black arrows between layers indicate the ReLU activations. $\gamma(\cdot)$ is a positional encoding function, we use $L=10$ for position, and $L=6$ for timestamp. Similar to NeRF~\cite{mildenhall2020_nerf_eccv20}, we use a skip connection that concatenates the input to the third layer.

\begin{figure*}[th]
    \centering
    \addtolength{\tabcolsep}{-6.5pt}
    \footnotesize{
        \setlength{\tabcolsep}{1pt} % Default value: 6pt
        \begin{tabular}{p{8.2pt}ccccccc}
            & $iteration=0$ & $iteration=3000$ & $iteration=12000$ & $iteration=30000$ & $iteration=40000$   \\
        \raisebox{25pt}{\rotatebox[origin=c]{90}{Lego}}&
             % \raisebox{20pt}{\rotatebox[origin=c]{90}{Ours}}&
             \includegraphics[width=0.17\textwidth]{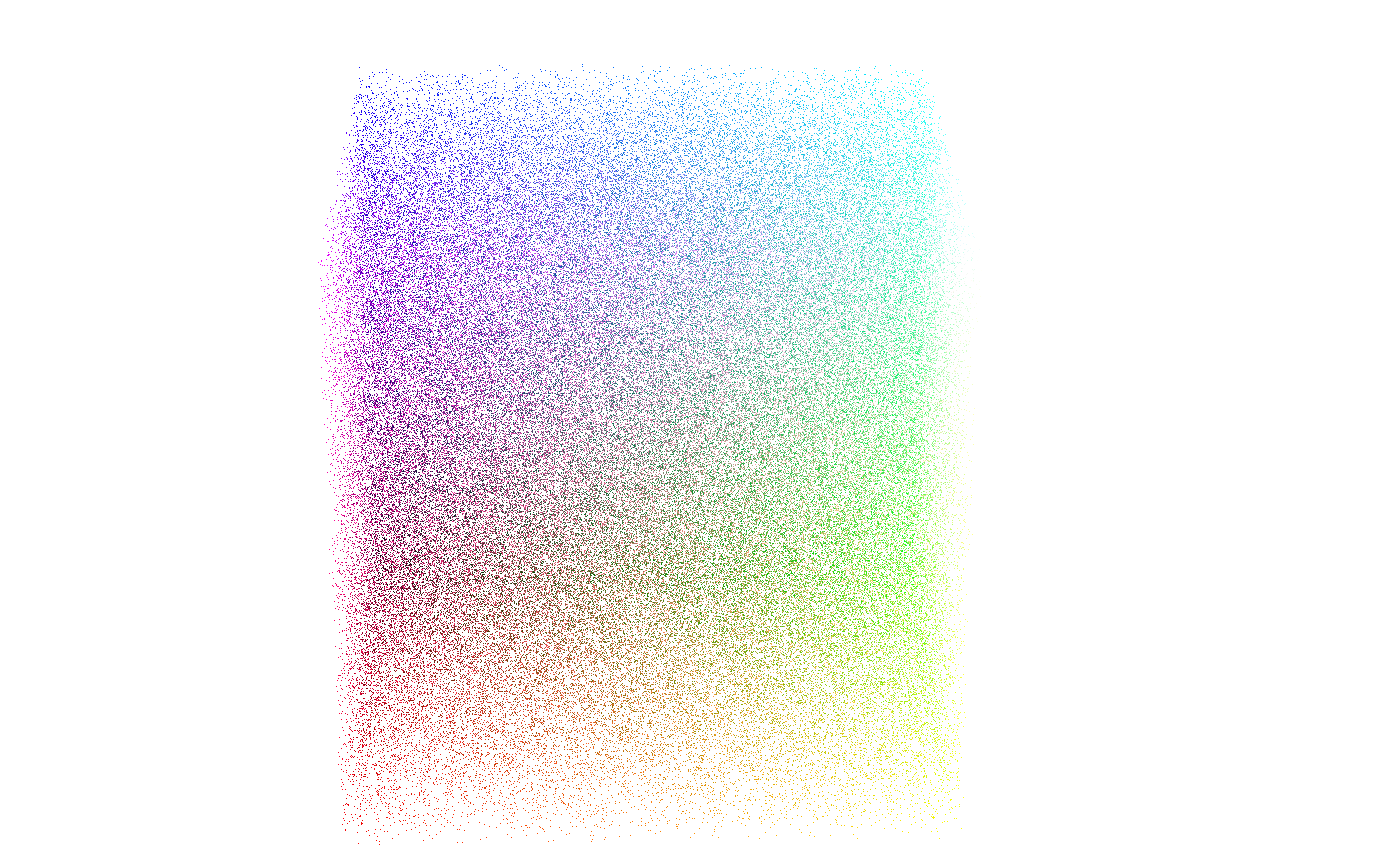} &
            \includegraphics[width=0.17\textwidth]{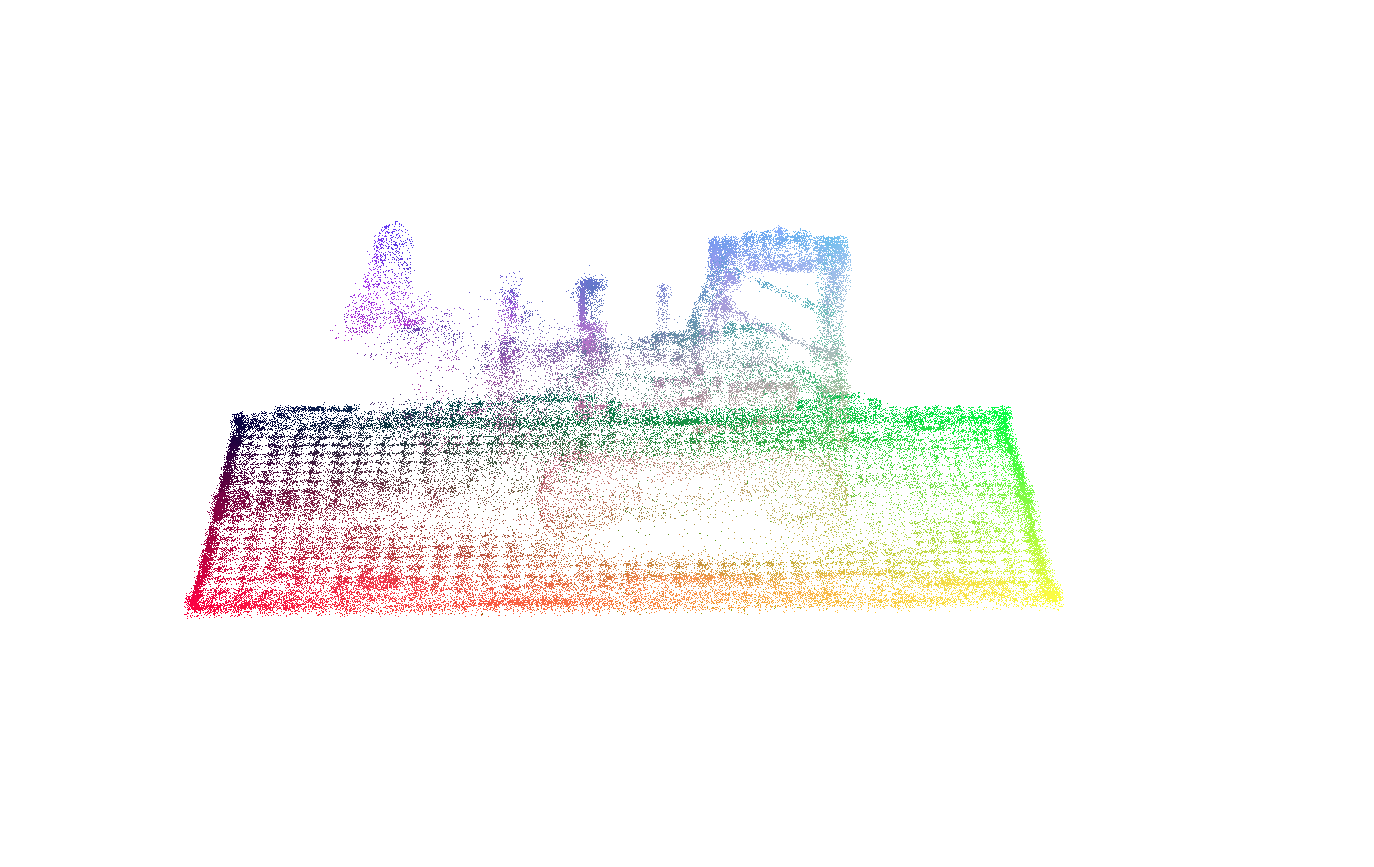} &
            \includegraphics[width=0.17\textwidth]{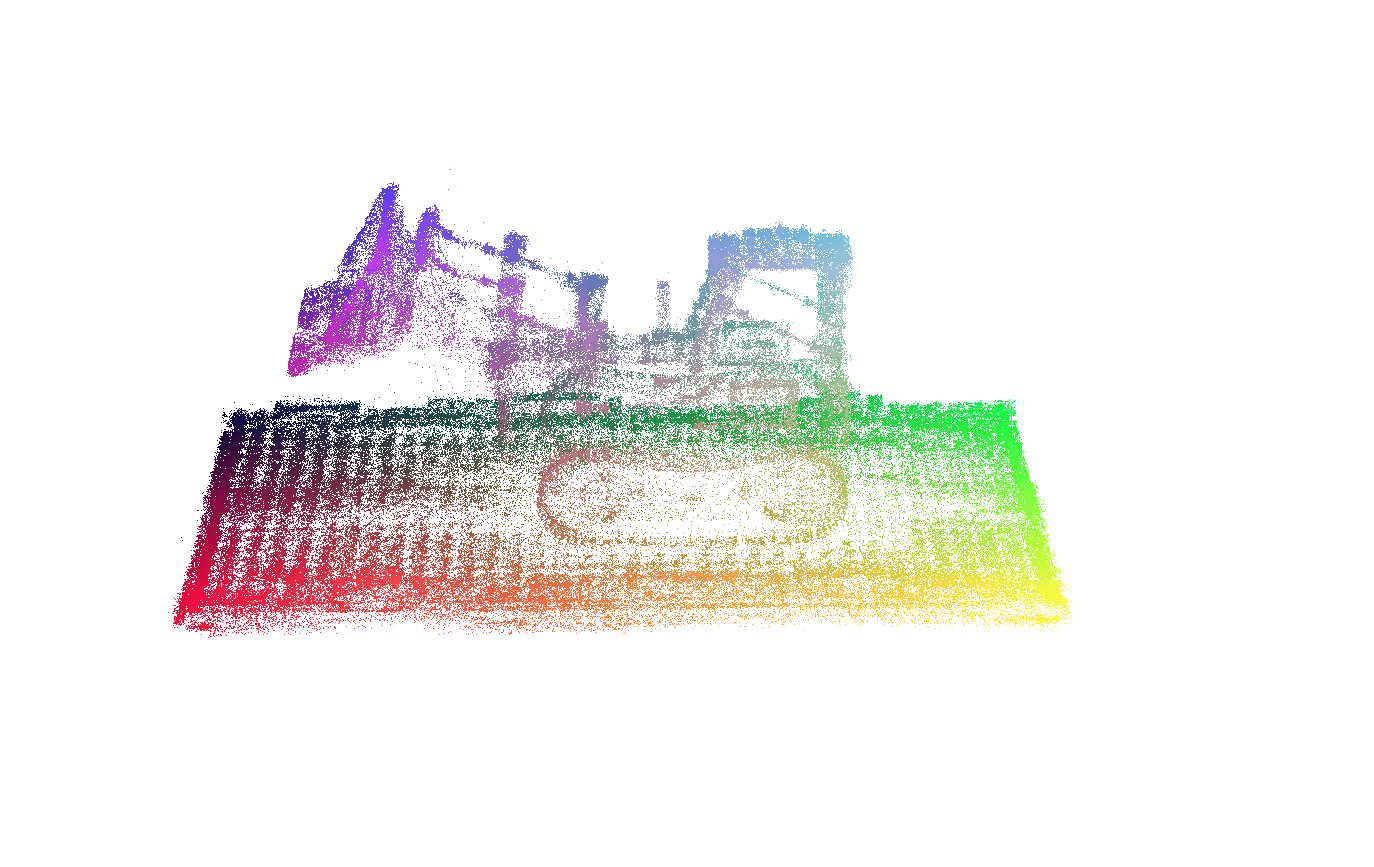} &
            \includegraphics[width=0.17\textwidth]{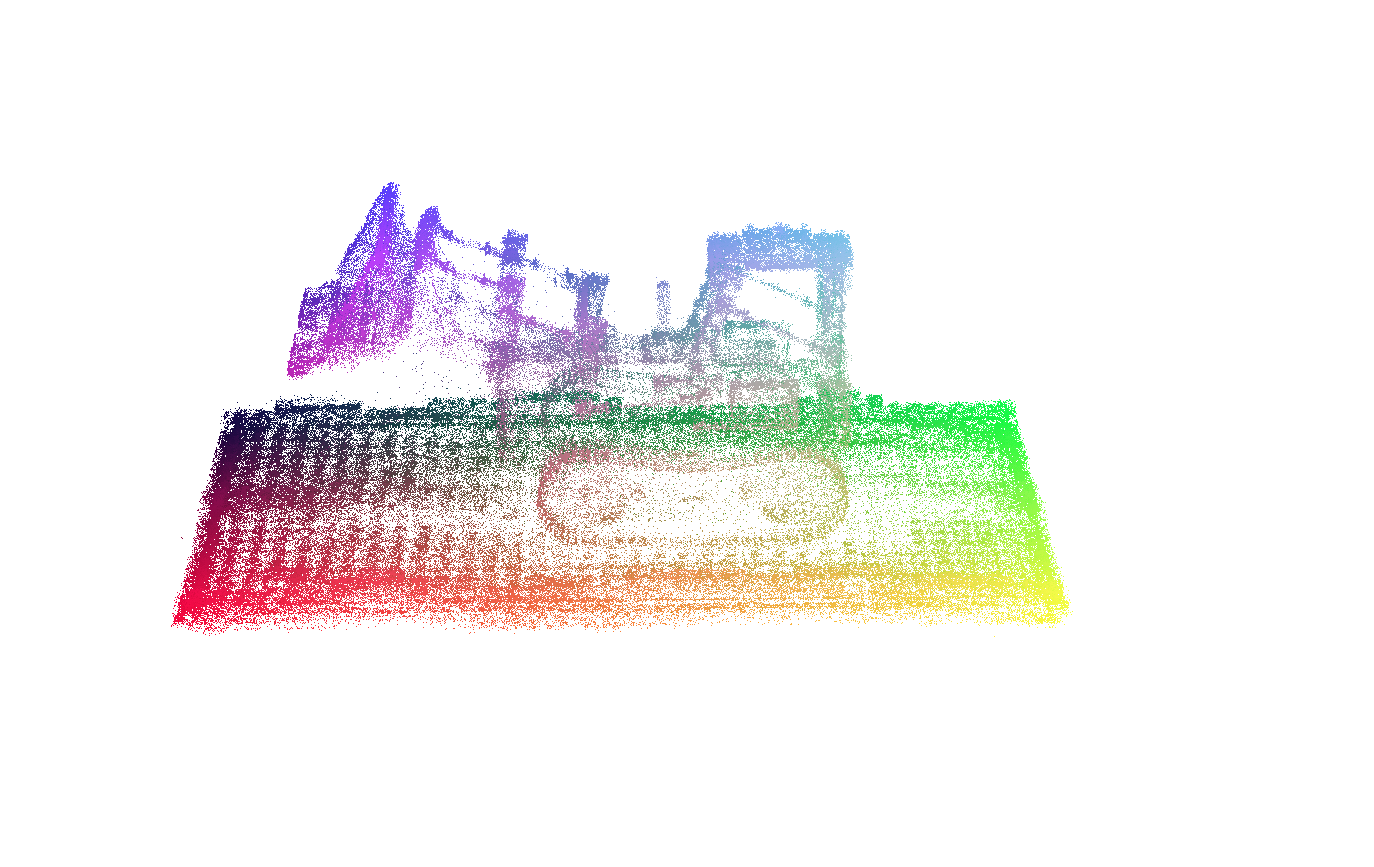} &
            \includegraphics[width=0.17\textwidth]{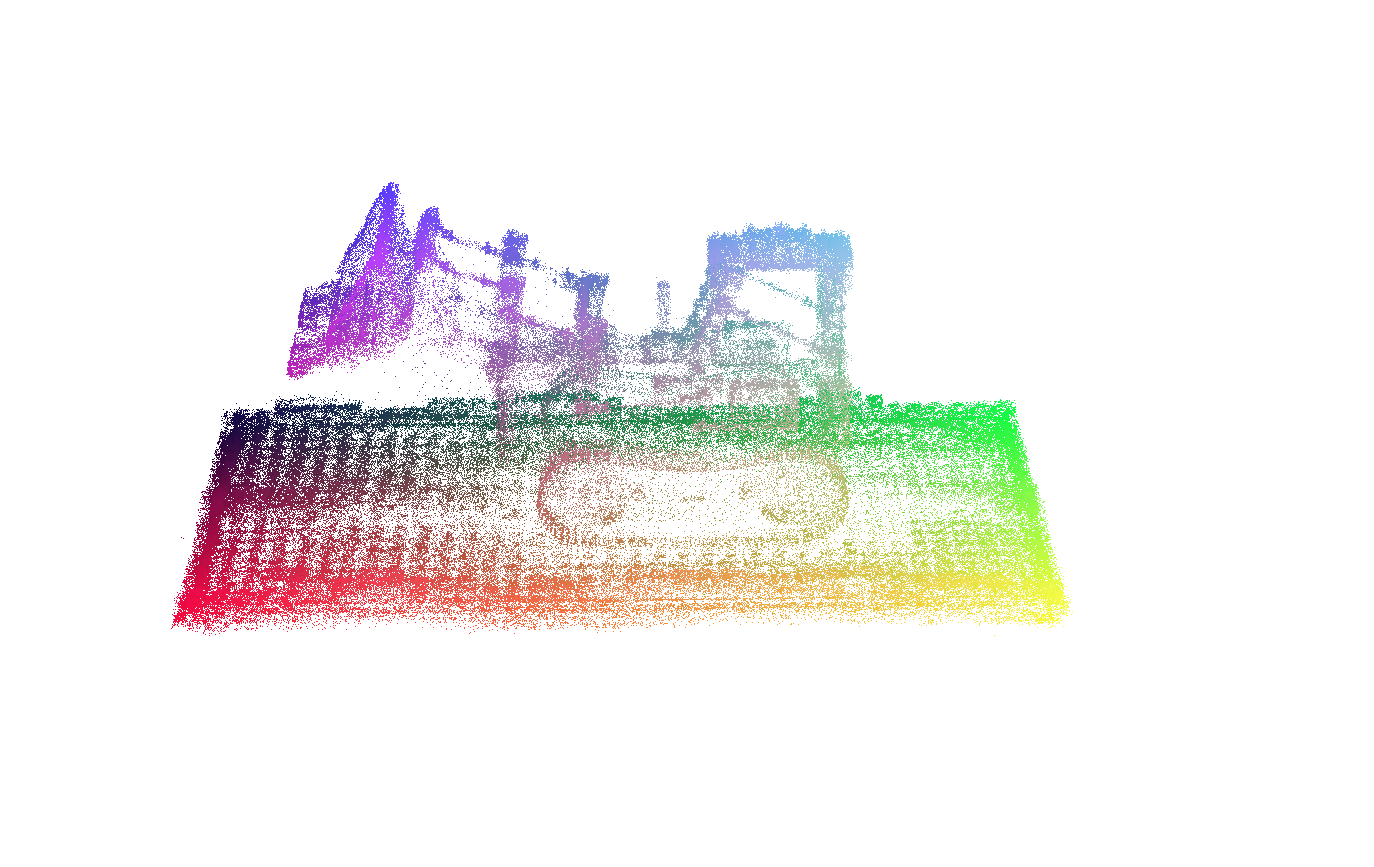} &
           
             \\
        \raisebox{20pt}{\rotatebox[origin=c]{90}{Peel Banana}}&
             \includegraphics[width=0.17\textwidth]{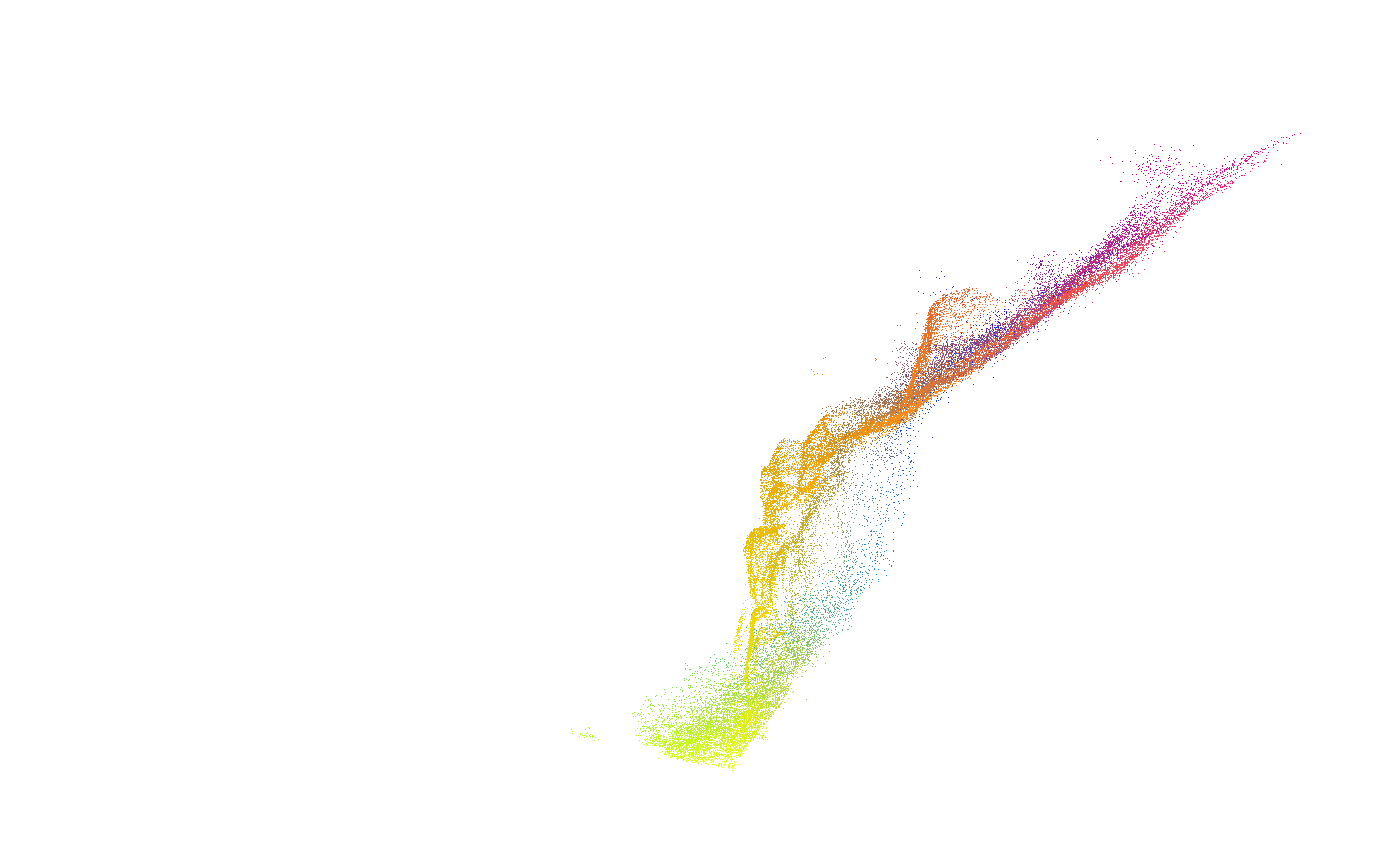} &
            \includegraphics[width=0.17\textwidth]{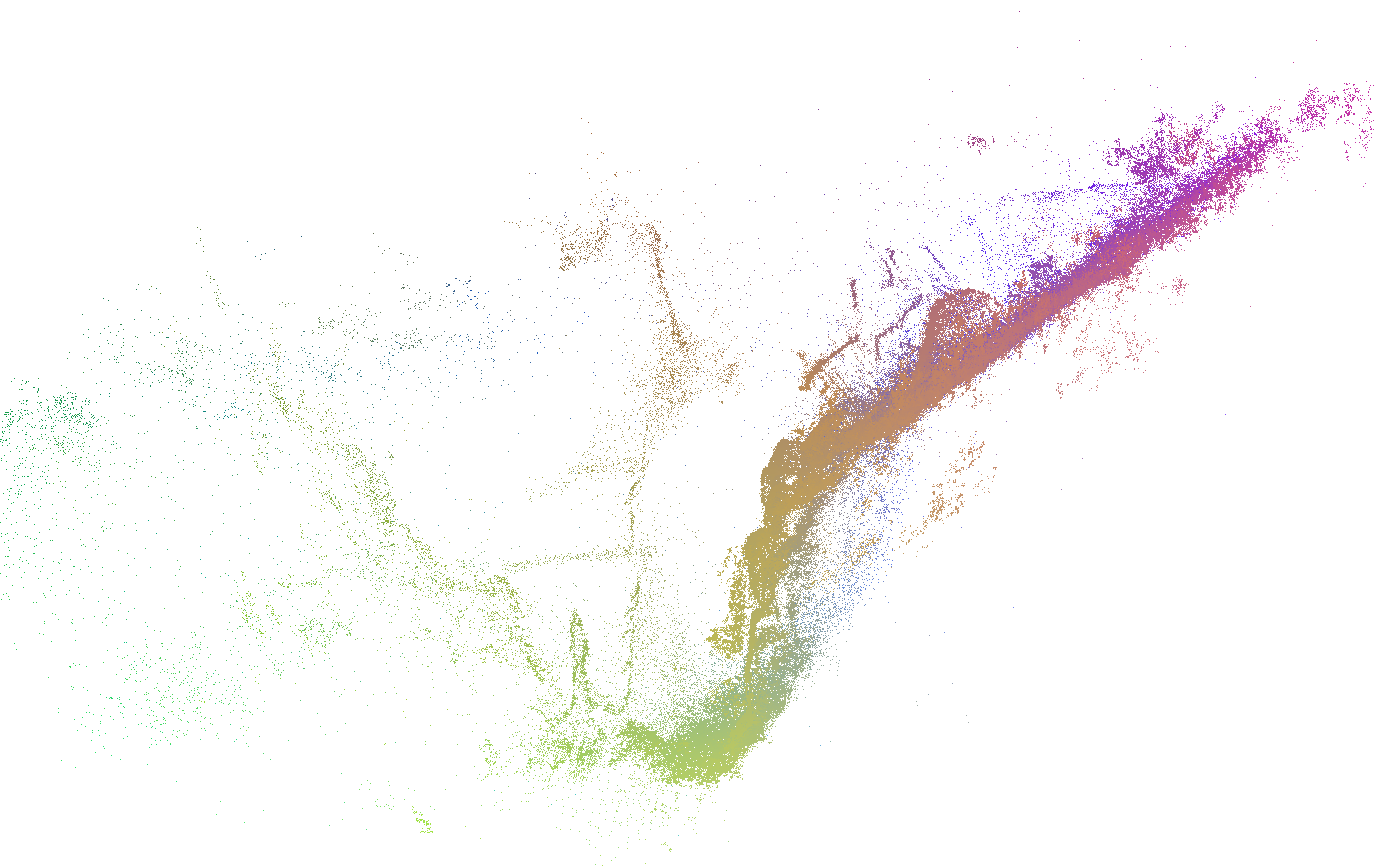} &
            \includegraphics[width=0.17\textwidth]{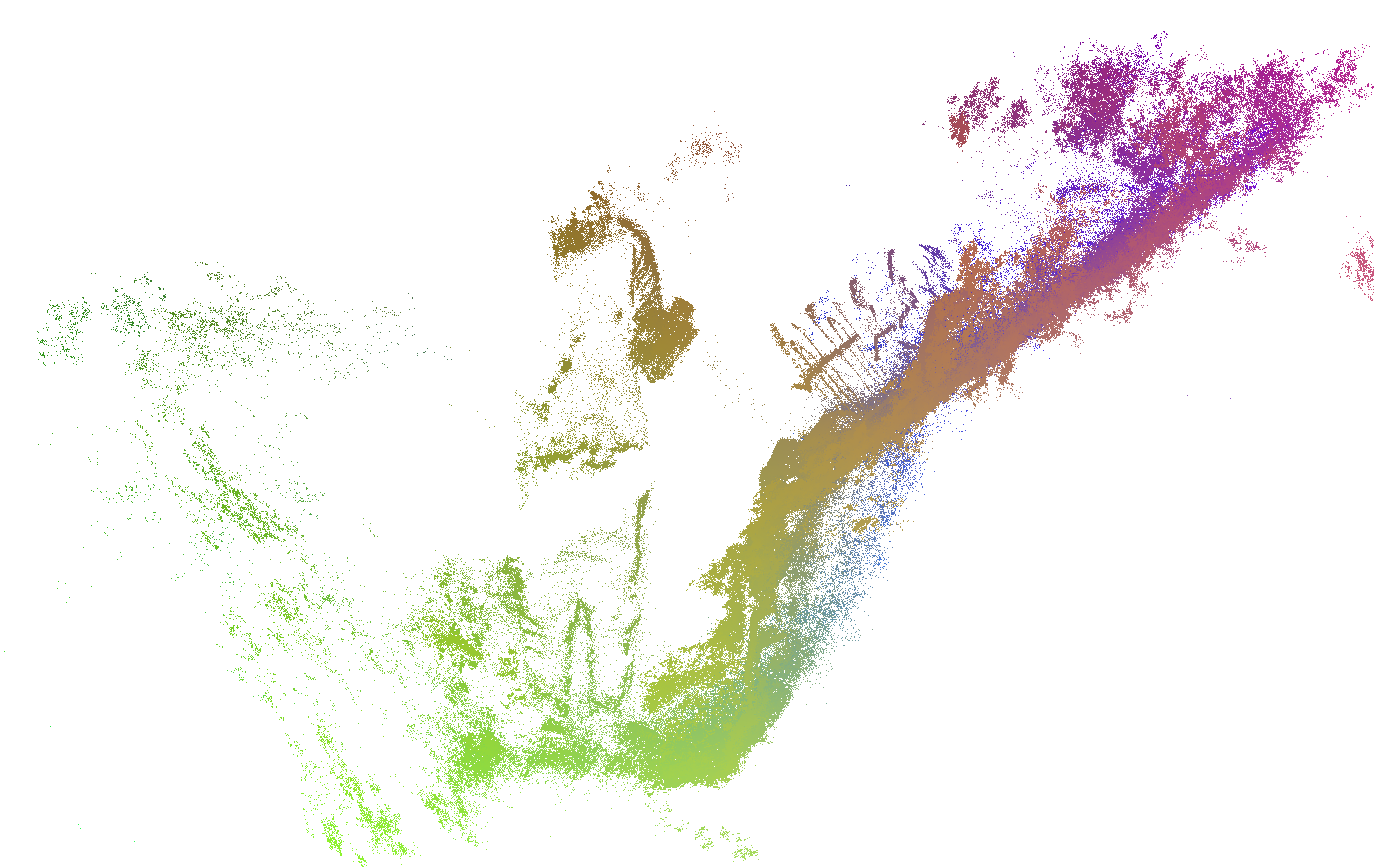} &
            \includegraphics[width=0.17\textwidth]{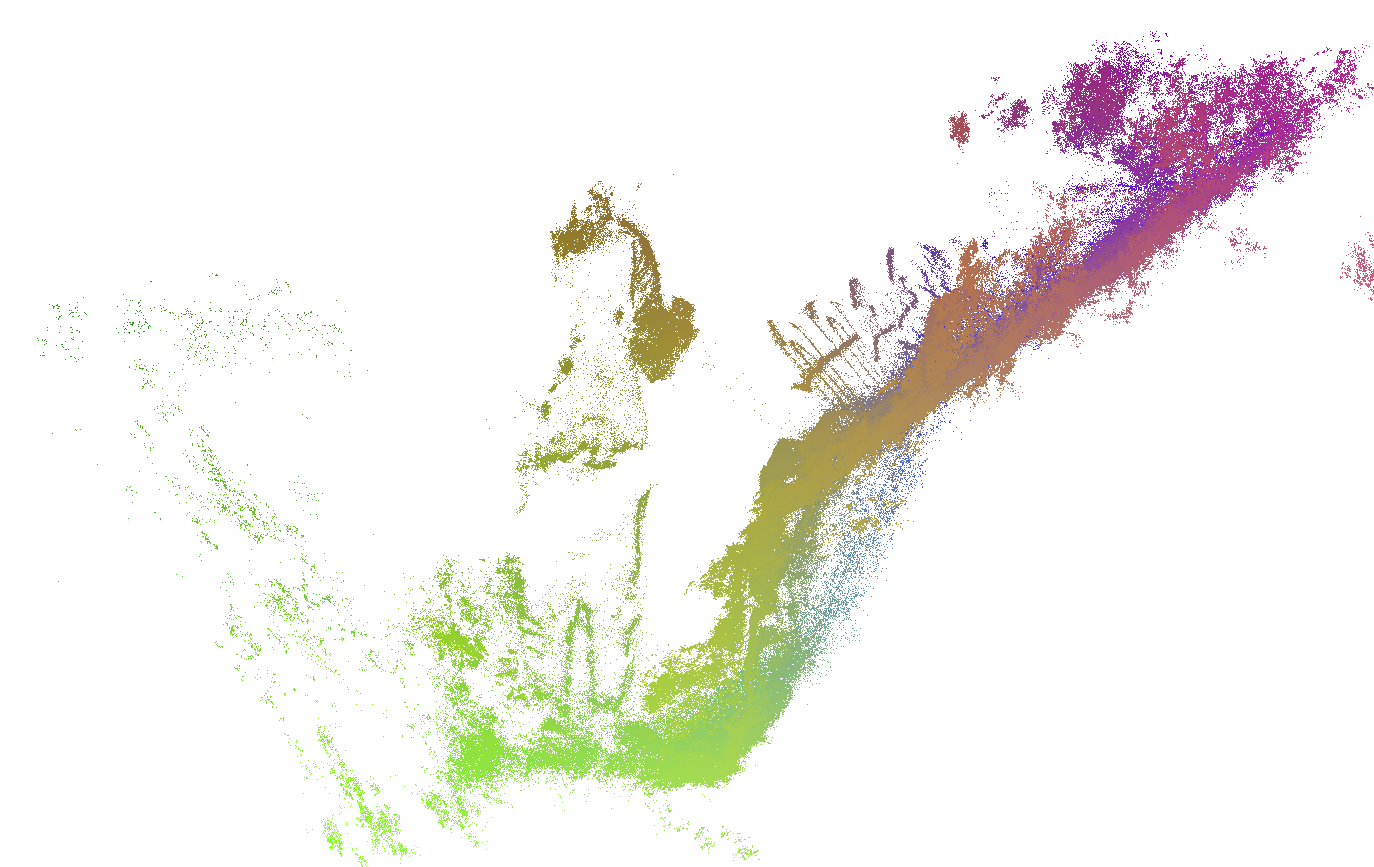} &
            \includegraphics[width=0.17\textwidth]{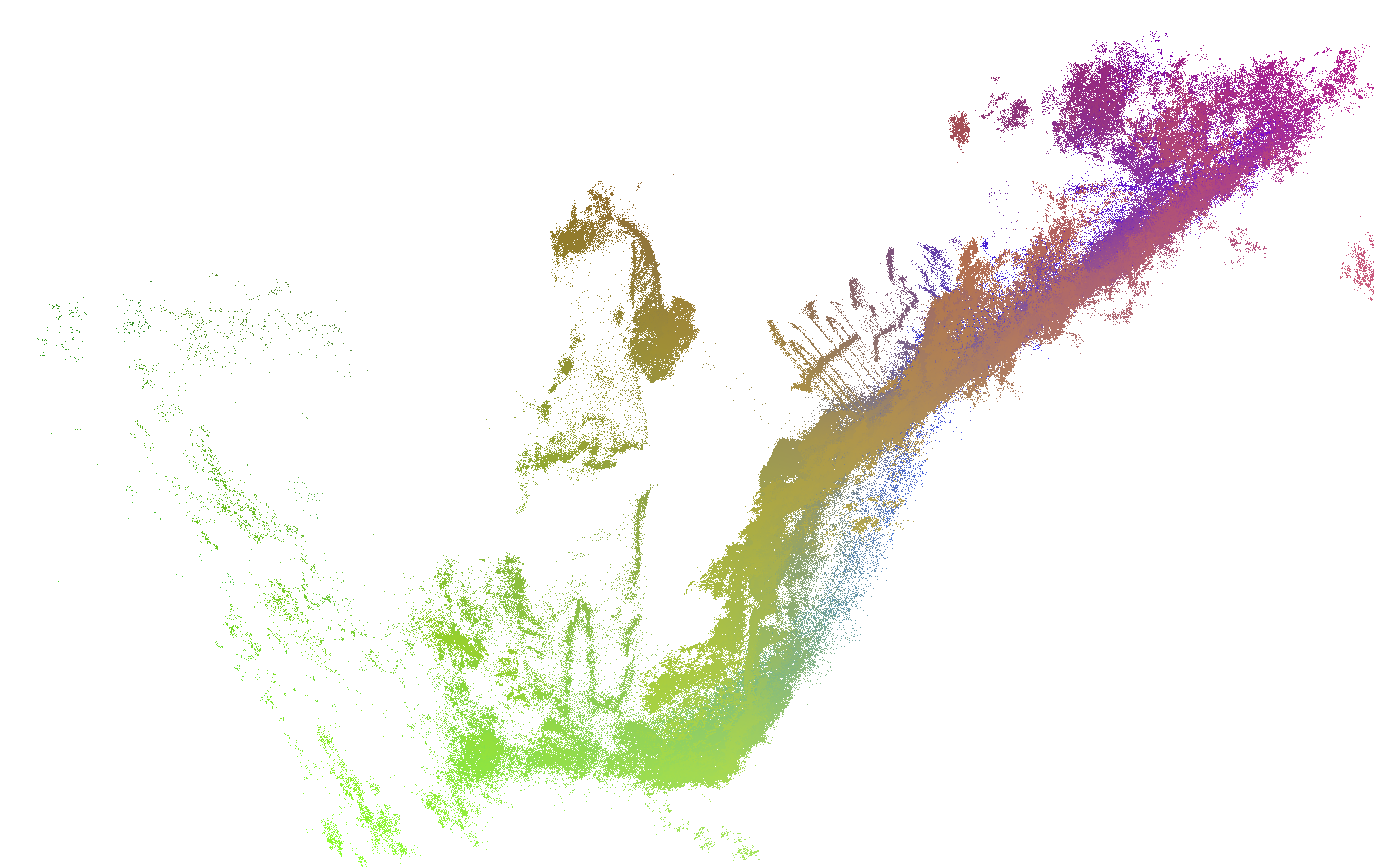} &

        \end{tabular}
    }
\vspace{-10pt}
	\caption{\textbf{Visualization of Canonical Point Cloud.} We show the evolution of point clouds in the canonical space with respect to the number of iterations.
}\label{fig:pcd}
\end{figure*}

\section{Results and Discussions}
\label{sec:C}

\subsection{Results on Neural 3D Video dataset}
We further evaluated our method on Neural 3D Video dataset~\cite{li2022neural}, which includes several videos captured with synchronized fixed GoPro camera system. We have evaluated our method in the following four scenarios:  Cook Spinach, Cut Roast Beef, Flame Steak and Sear Steak, each scene includes from 17 to 20 cameras for training and one central camera for evaluation. Following previous works, we downsample the images to 1352 $\times$ 1014 and report the per-scene PSNR, SSIM and LPIPS for each method, as shown in \Tref{table:supp}. We find our method is struggling in these long-time series. Although our method maintains high fidelity restoration in static regions, its capability is severely limited in dynamic regions.

\begin{table*}[hbt!]
	\centering
	\caption{Quantitative results on scenes from the Neural 3D Video Synthesis }
	\vspace{-5pt}
	\label{table:supp}
	\renewcommand{\arraystretch}{1.0}
	\begin{tabular}{l @{\hspace{1.\tabcolsep}}
			c @{\hspace{1.\tabcolsep}} c @{\hspace{1.\tabcolsep}} c @{\hspace{1.\tabcolsep}}
			c @{\hspace{1.\tabcolsep}} c @{\hspace{1.\tabcolsep}} c 
			@{\hspace{1.\tabcolsep}}
			c @{\hspace{1.\tabcolsep}} c @{\hspace{1.\tabcolsep}} c 
			@{\hspace{1.\tabcolsep}}
			c @{\hspace{1.\tabcolsep}} c @{\hspace{1.\tabcolsep}} c 
			@{\hspace{1.\tabcolsep}}
			%c @{\hspace{0.5\tabcolsep}} c @{\hspace{0.5\tabcolsep}} c
		}
		\toprule
		\multicolumn{1}{l}{Scene} & \multicolumn{3}{c}{Cook Spinach} & \multicolumn{3}{c}{Cut Roast Beef} & \multicolumn{3}{c}{Flame Steak} & \multicolumn{3}{c}{Sear Steak} \\ %& \multicolumn{3}{c}{Coffee Martini} \\
		\cmidrule(lr){2-4}\cmidrule(lr){5-7}\cmidrule(lr){8-10}\cmidrule(lr){11-13}%\cmidrule(lr){14-16}
		Method & PSNR$\uparrow$ & SSIM$\uparrow$ & LPIPS$\downarrow$ & PSNR$\uparrow$ & SSIM$\uparrow$ & LPIPS$\downarrow$ & PSNR$\uparrow$ & SSIM$\uparrow$ & LPIPS$\downarrow$ & PSNR$\uparrow$ & SSIM$\uparrow$ & LPIPS$\downarrow$ \\ %& PSNR & SSIM & LPIPS\\ 
		\midrule
		MixVoxels~\cite{wang2023mixed} & $31.39$ & $0.931$ & $0.113$ & $31.38$ & $0.928$ & $0.111$ & $30.15$ & $0.938$ & $0.108$ & $30.85$ & $0.940$ & $0.103$ \\%& $29.25$ & $0.901$ & $0.147$ \\
		K-Planes~\cite{kplanes} & $31.23$ & $0.926$ & $0.114$ & $31.87$ & $0.928$ & $0.114$ & $31.49$ & $0.940$ & $0.102$ & $30.28$ & $0.937$ & $0.104$ \\%& $29.30$ & $0.900$ & $0.134$ \\ 
		%DyNeRF & & & &  & \\
		Hexplanes\textsuperscript{\textdaggerdbl}~\cite{hexplane} & $31.05$ & $0.928$ & $0.114$ & $30.83$ & $0.927$ & $0.115$ & $30.42$ & $0.939$ & $0.104$ & $30.00$ & $0.939$ & $0.105$ \\ %& $28.45$ & $0.891$ & $0.149$ \\ %& $31.92$ & $0.936$ & $0.078$ & $31.65$ & $0.934$ & $0.077$ & $31.36$ & $0.949$ & $0.065$ & $30.87$ & $0.949$ & $0.066$ & $28.34$ & $0.897$ & $0.097$ \\
		% Hyperreel\textsuperscript{\textdagger} & $32.30$ & $0.941$ & $0.089$ & $32.92$ & $0.945$ & $0.084$ & $32.20$ & $0.949$ & $0.078$ & $32.57$ & $0.952$ & $0.077$ & $28.37$ & $0.892$ & $0.127$ \\
		Hyperreel~\cite{attal2023hyperreel} & $31.77$ & $0.932$ & \textbf{0.090} & \textbf{32.25} & $0.936$ & \textbf{0.086} & $31.48$ & $0.939$ & \textbf{0.083} & $31.88$ & $0.942$ & \textbf{0.080} \\%& $28.65$ & $0.897$ & $0.129$ \\
		NeRFPlayer\textsuperscript{\textdagger}~\cite{nerfplayer} & $30.58$ & $0.929$ & $0.113$ & $29.35$ & $0.908$ & $0.144$ & $31.93$ & $0.950$ & $0.088$ & $29.13$ & $0.908$ & $0.138$ \\ %& $31.53$ & $0.951$ & $0.085$ \\
		StreamRF~\cite{li2022streaming} & $30.89$ & $0.914$ & $0.162$ & $30.75$ & $0.917$ & $0.154$ & $31.37$ & $0.923$ & $0.152$ & $31.60$ & $0.925$ & $0.147$ \\ %& $28.13$ & $0.873$ & $0.219$ \\
		%Tensor4D & & & & &  \\
		%D-NeRF & & & & & \\
		SWAGS~\cite{shaw2023swags}  & \textbf{31.96} & 0.946 & 0.094 & 31.84 & \textbf{0.945} & 0.099 & \textbf{32.18} & 0.953 & 0.087 & 32.21 & 0.950 & 0.092 \\ %& 24.60 & 0.877 & 0.154 \\ 
		\textbf{Ours} & 31.39 & \textbf{0.947} & 0.144 & 29.87 & 0.944 & 0.156 & 31.35 & \textbf{0.954} & 0.129 & \textbf{32.62} & \textbf{0.955} & 0.130          \\
		\bottomrule
	\end{tabular}
	
\end{table*}

\subsection{More Visualization Results}
\noindent\textbf{Point Cloud}
For the D-NeRF synthetic scenes~\cite{pumarola2021_dnerf_cvpr21}, we randomly initialize 150000 points as the initial point cloud. We visualize the point cloud of the scene in the canonical space with different iterations. In~\fref{fig:pcd}, it can be observed that we can reconstruct the scene even from a random point cloud. Moreover, in complex scenes such as Peel Banana in the HyperNeRF dataset~\cite{park2021hypernerf}, we can also reconstruct the scene even if there are no dynamic parts in the input point clouds, as shown in~\fref{fig:pcd}. Our supplementary video also presents the trajectory of the scene's point cloud as it evolves over time. Our supplementary video is available at our homepage: \href{https://npucvr.github.io/GaGS/}{https://npucvr.github.io/GaGS/}.

\begin{figure*}
	\centering
	\addtolength{\tabcolsep}{-6.5pt}
	\footnotesize{
		\setlength{\tabcolsep}{1pt} % Default value: 6pt
		\begin{tabular}{p{8.2pt}ccccccc}
			& $p_0$ & $p_1$ & $p_2$ & $p_3$ & $p_4$   \\
			\raisebox{25pt}{\rotatebox[origin=c]{90}{Sear Steak}}&
			% \raisebox{20pt}{\rotatebox[origin=c]{90}{Ours}}&
			\includegraphics[width=0.17\textwidth]{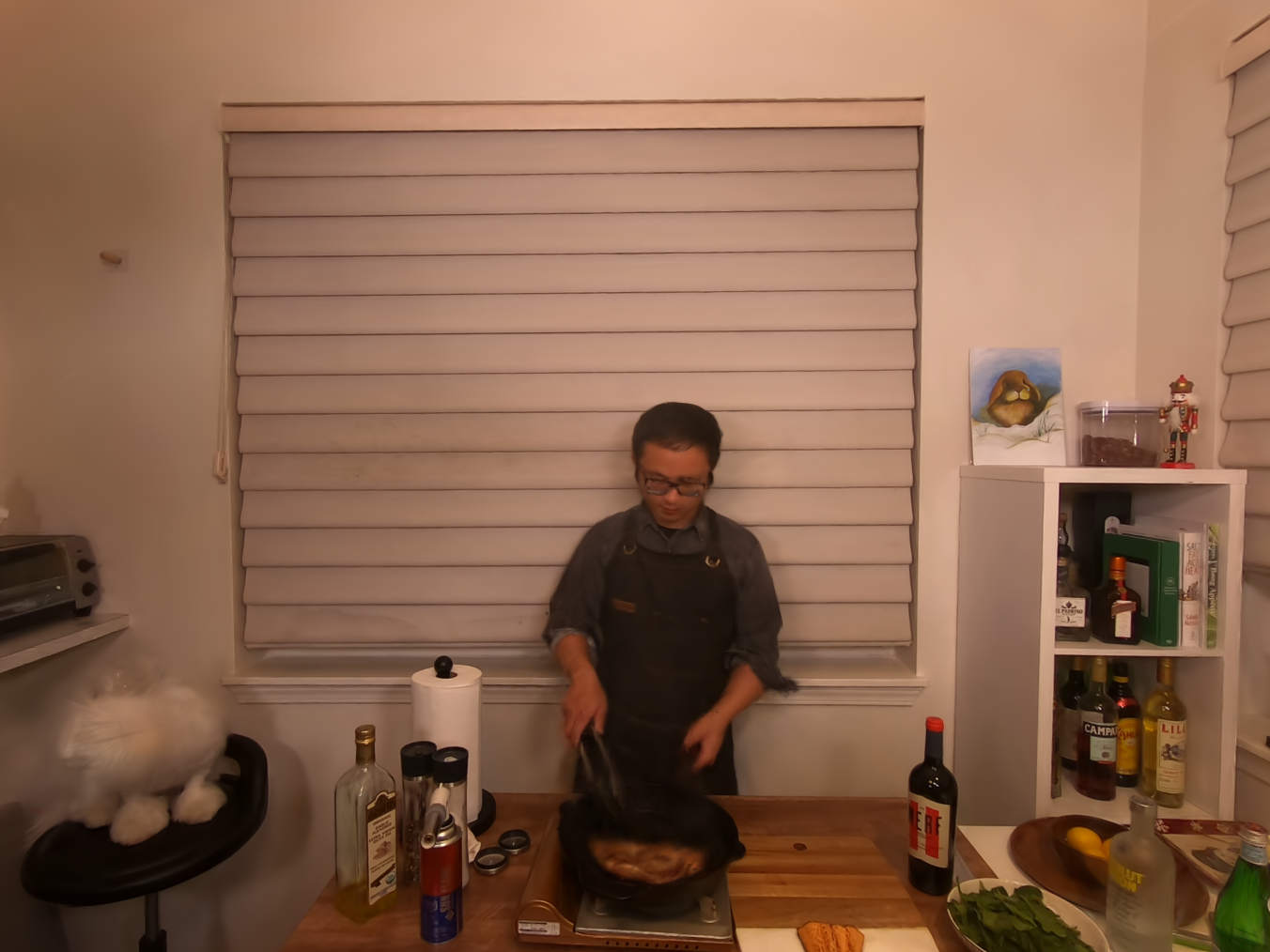} &
			\includegraphics[width=0.17\textwidth]{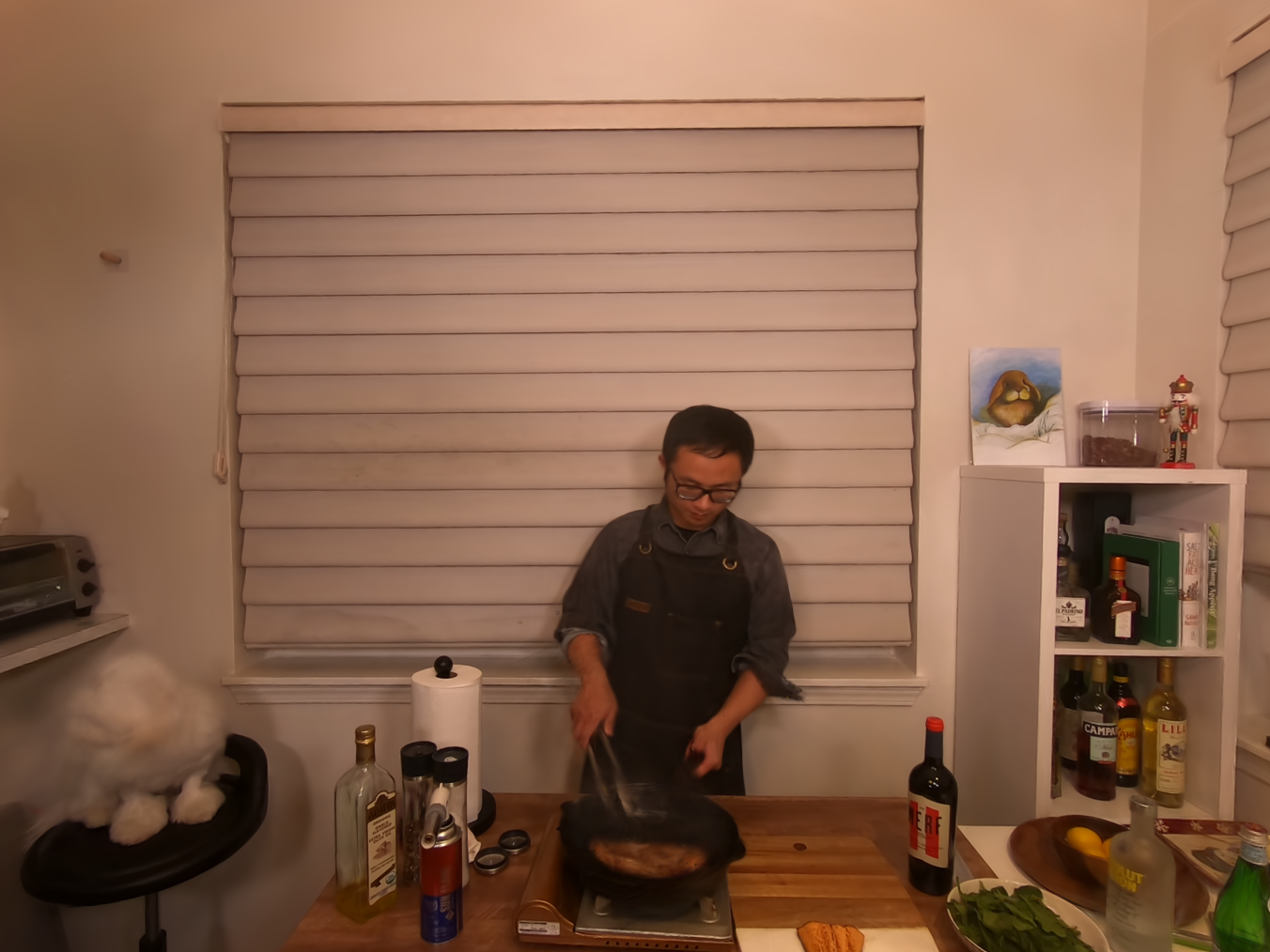} &
			\includegraphics[width=0.17\textwidth]{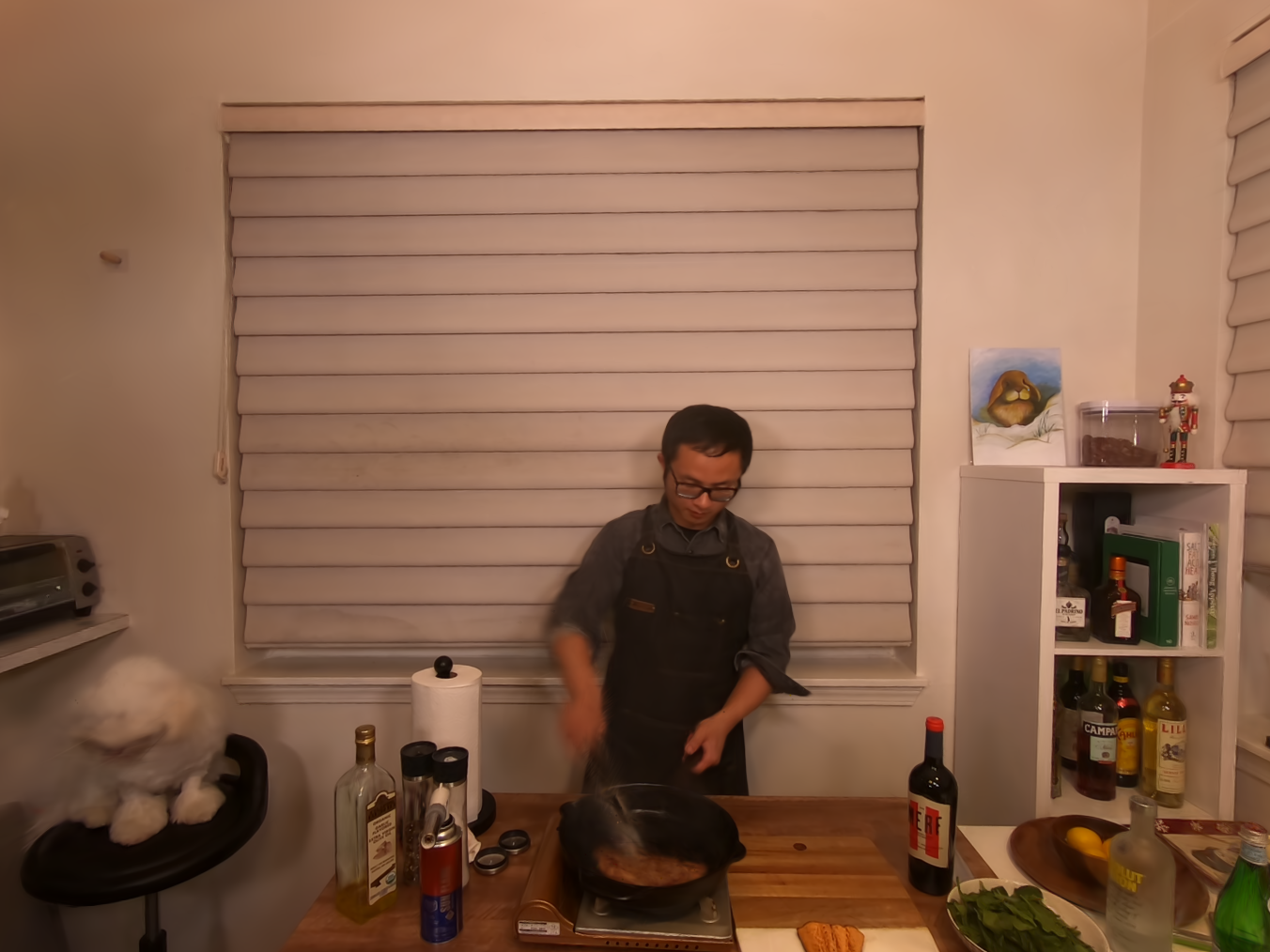} &
			\includegraphics[width=0.17\textwidth]{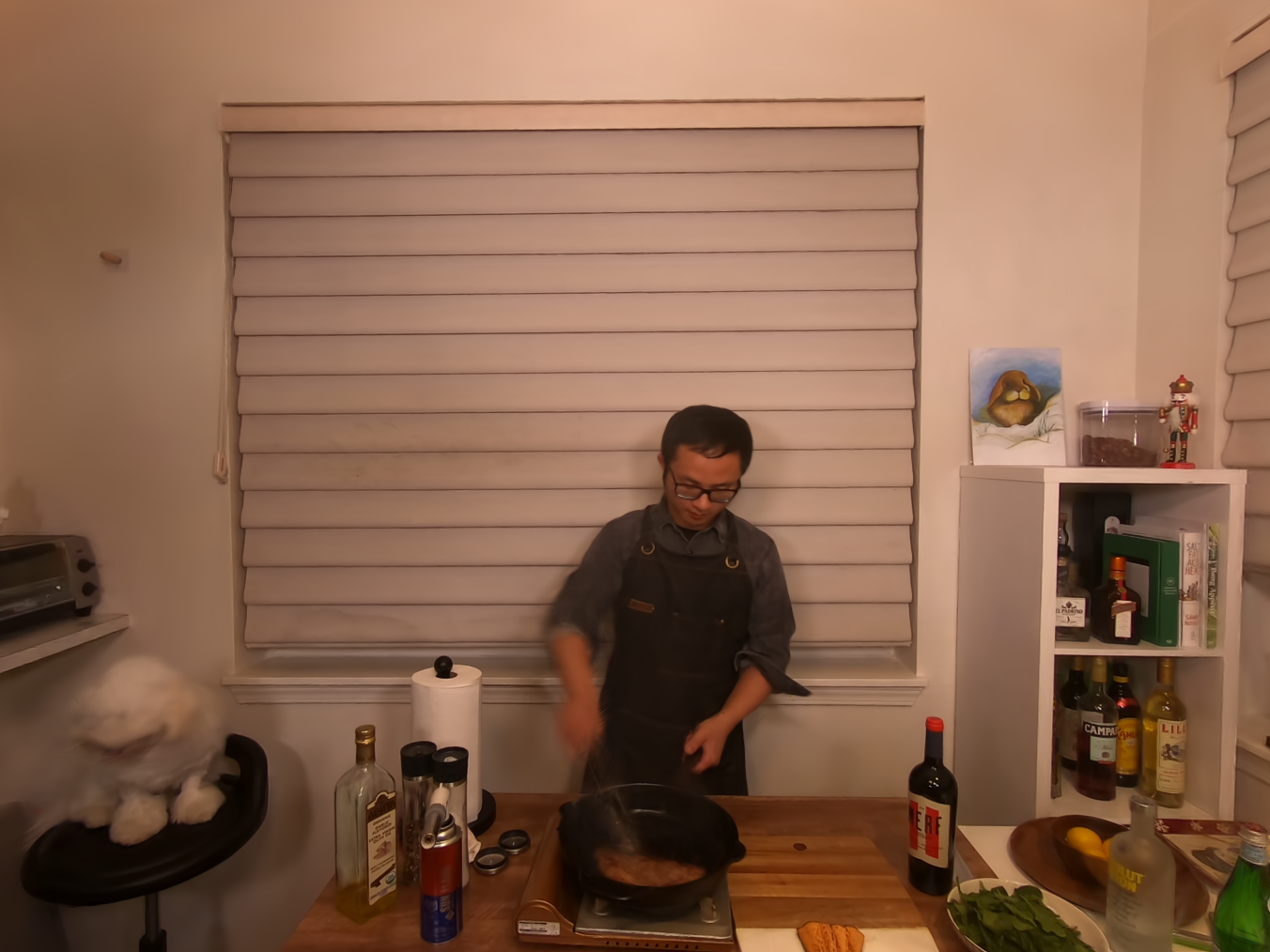} &
			\includegraphics[width=0.17\textwidth]{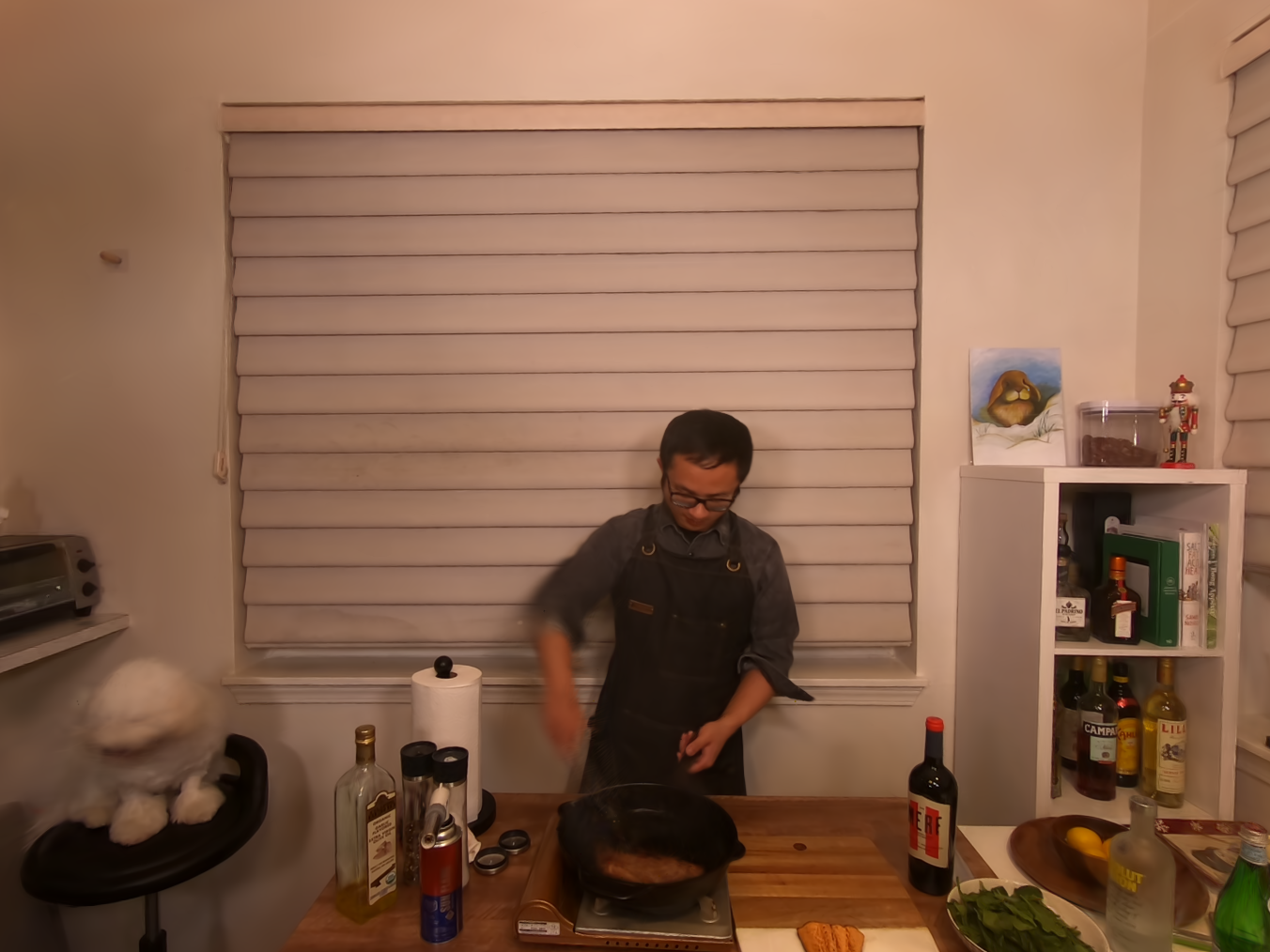} &
			
			\\
			\raisebox{25pt}{\rotatebox[origin=c]{90}{Flame Steak}}&
			\includegraphics[width=0.17\textwidth]{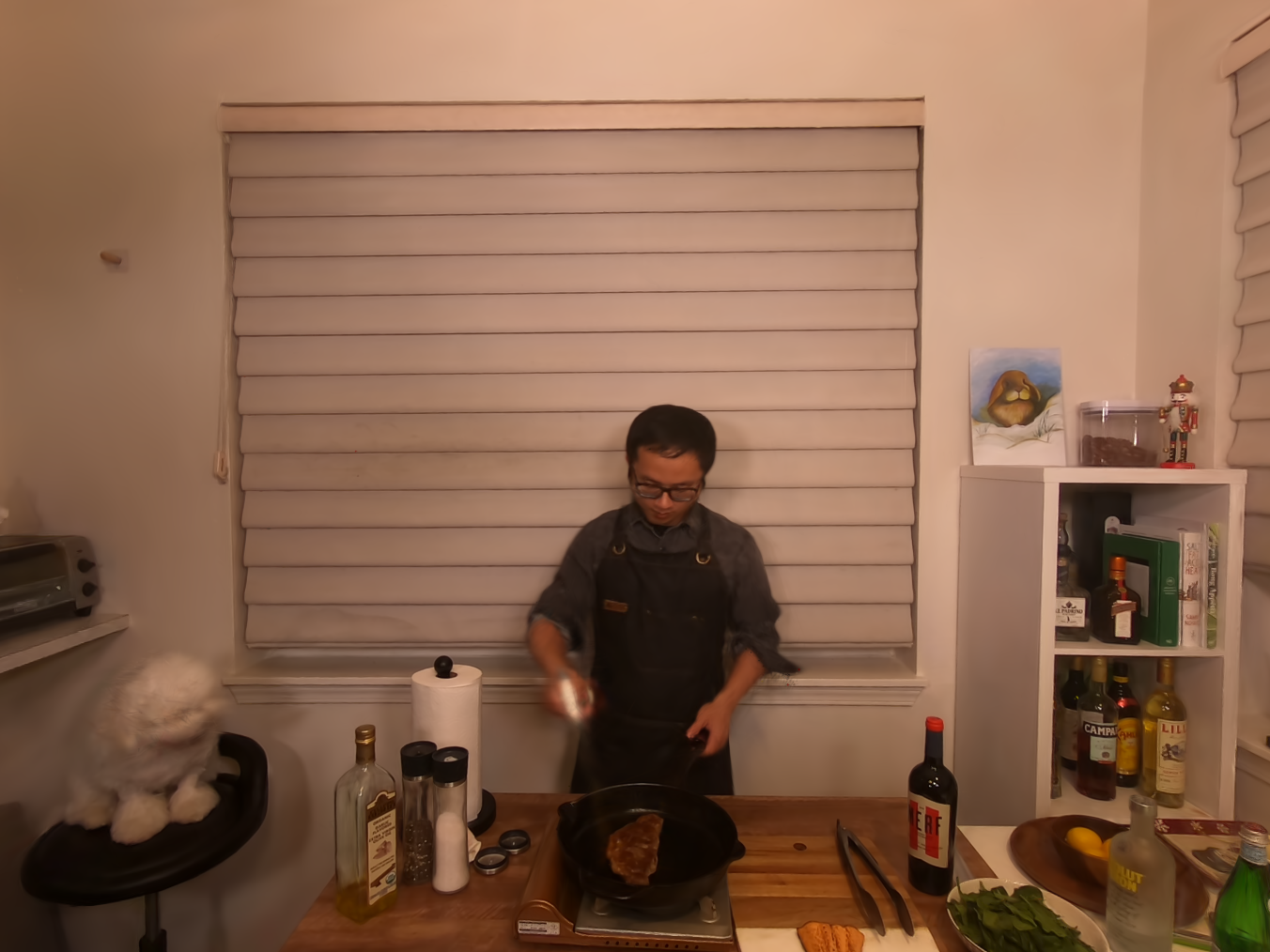} &
			\includegraphics[width=0.17\textwidth]{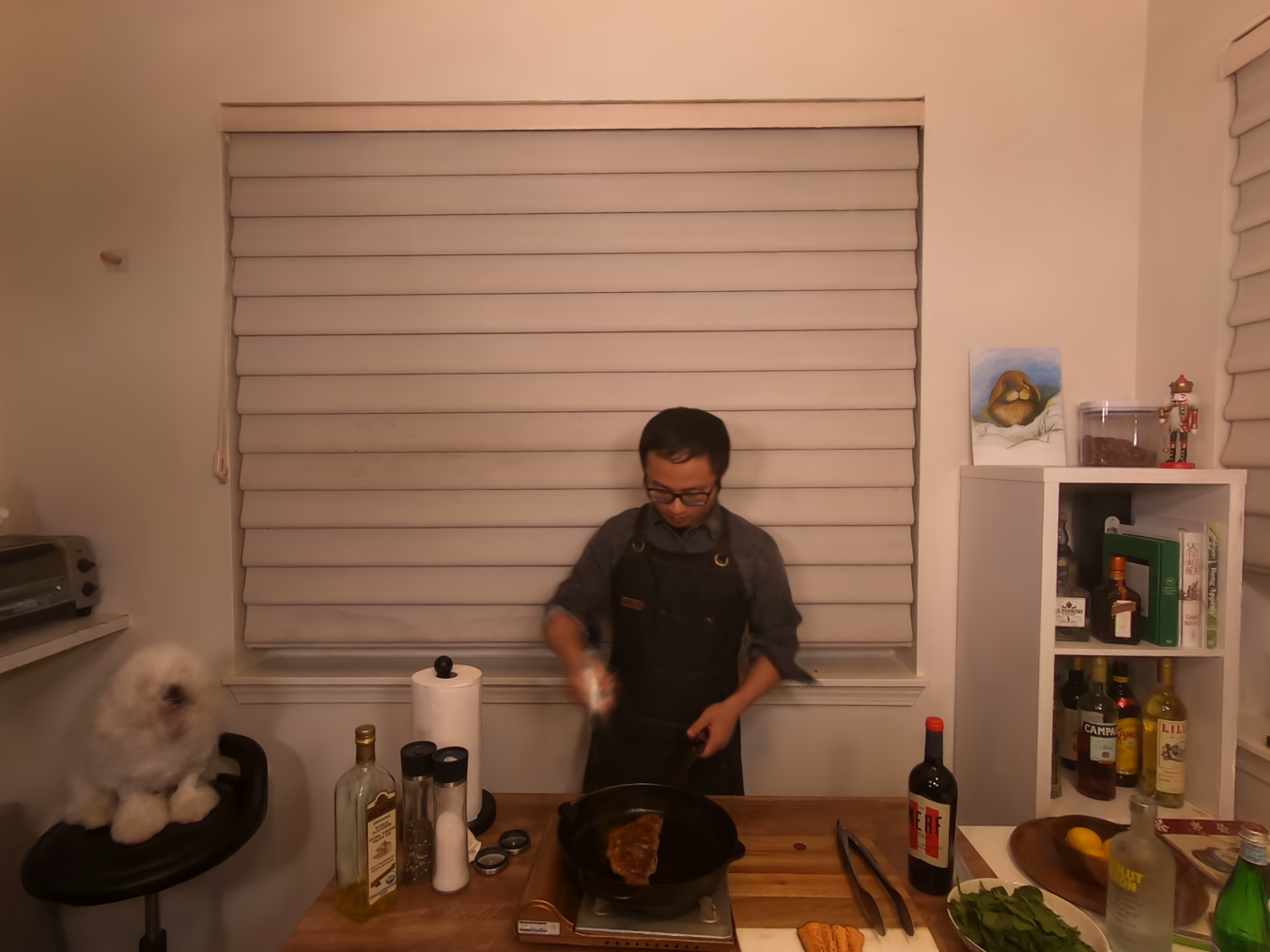} &
			\includegraphics[width=0.17\textwidth]{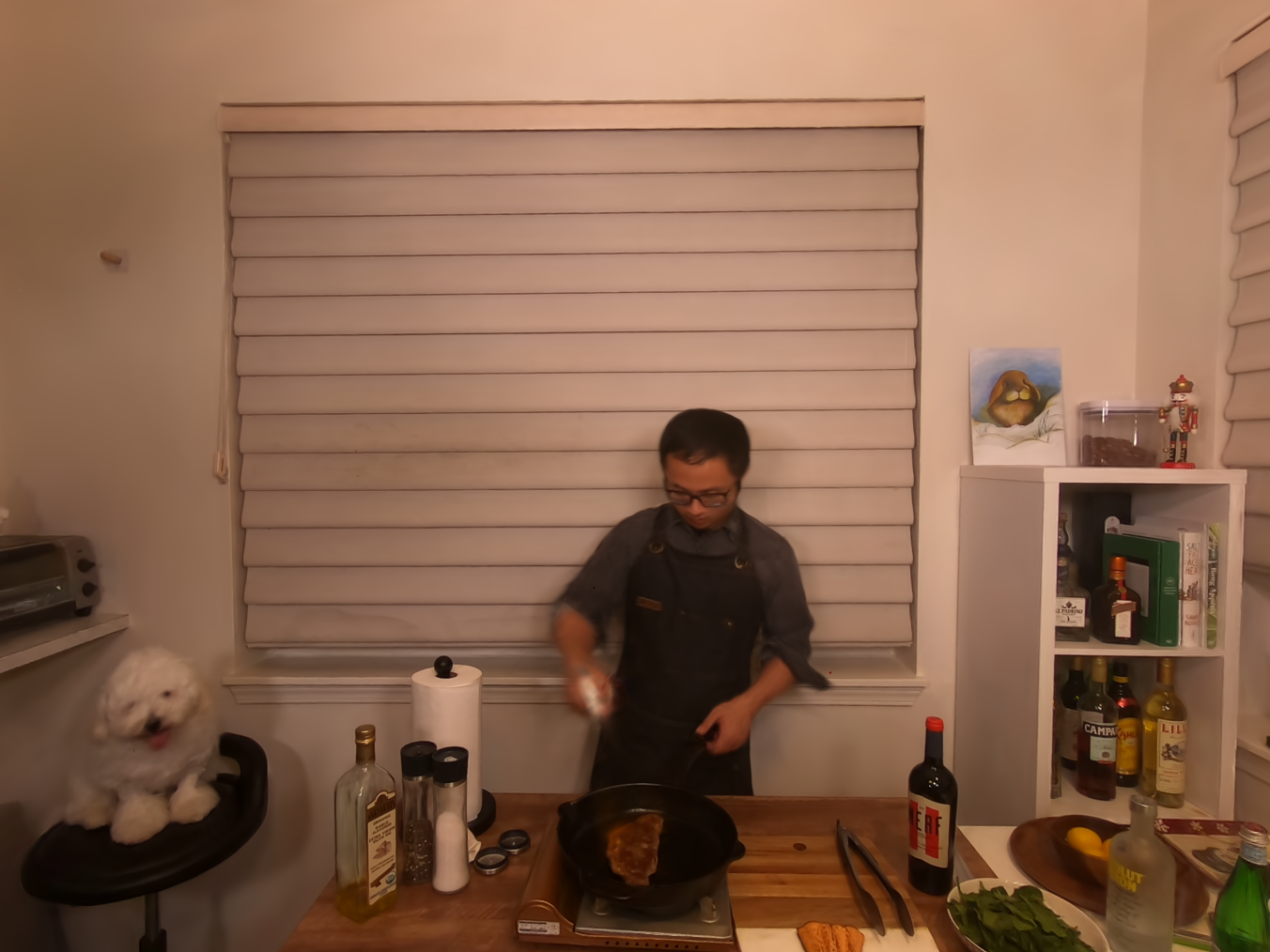} &
			\includegraphics[width=0.17\textwidth]{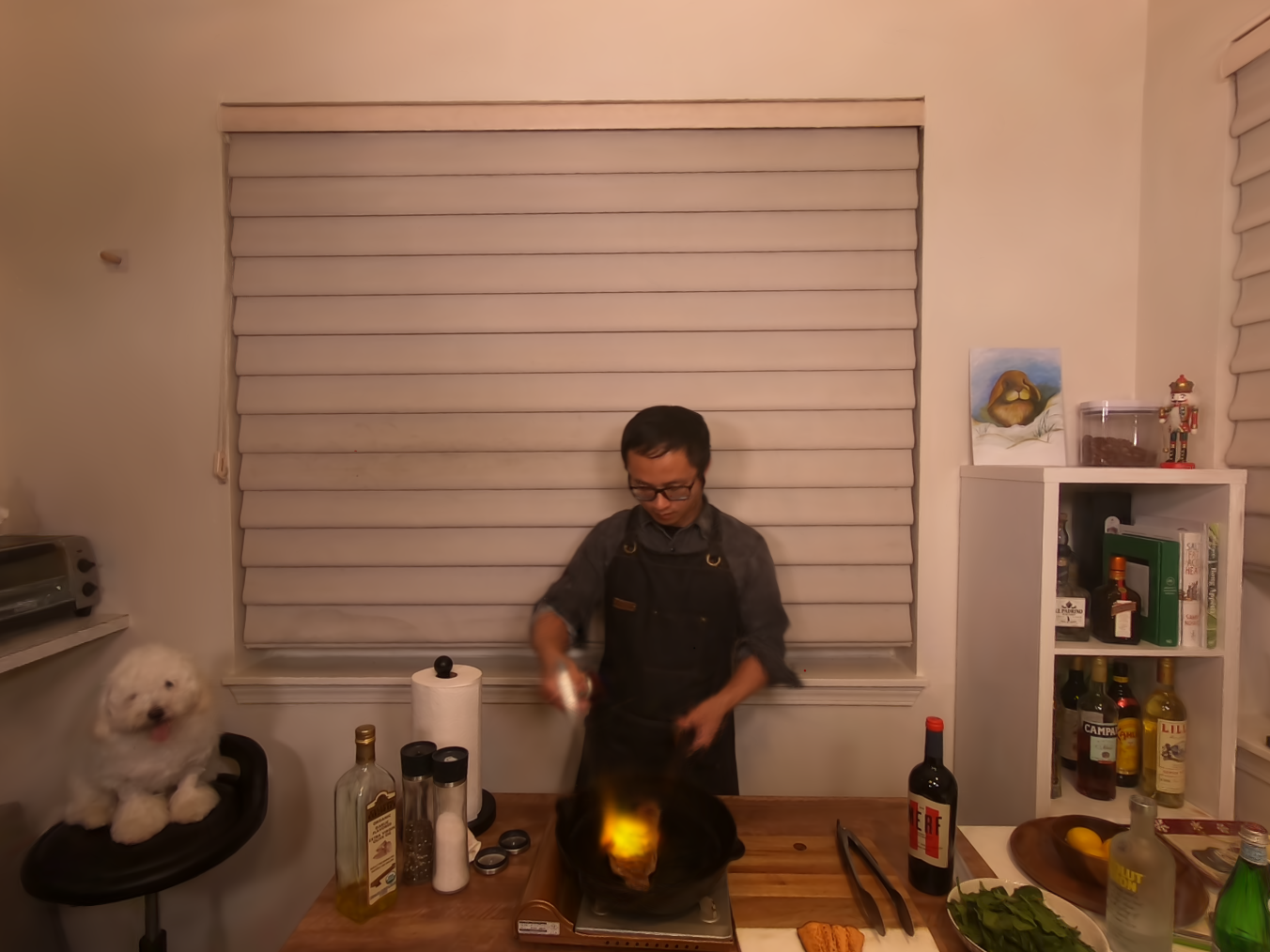} &
			\includegraphics[width=0.17\textwidth]{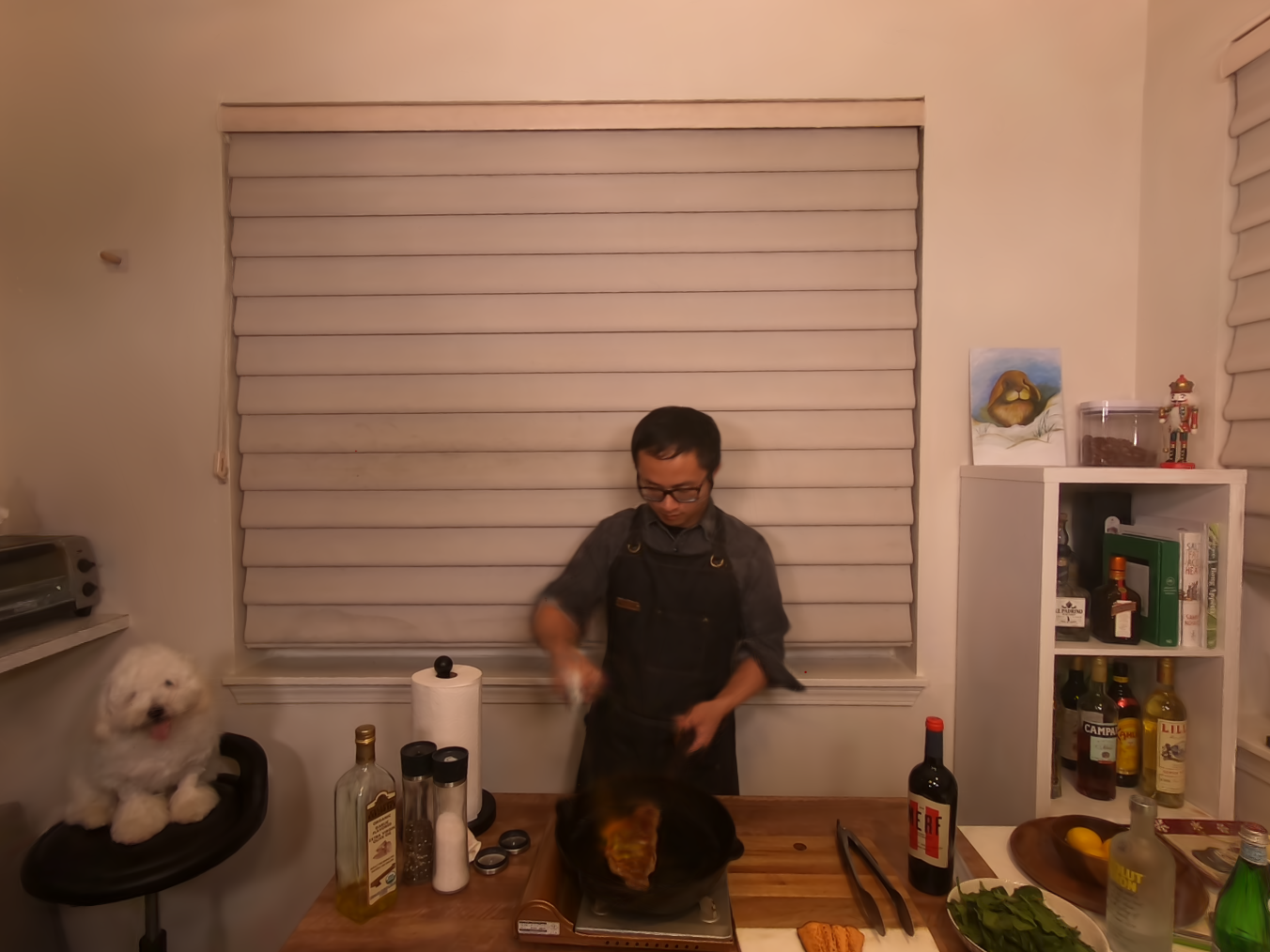} &
			
			\\
			\raisebox{25pt}{\rotatebox[origin=c]{90}{Cook Spinach}}&
			\includegraphics[width=0.17\textwidth]{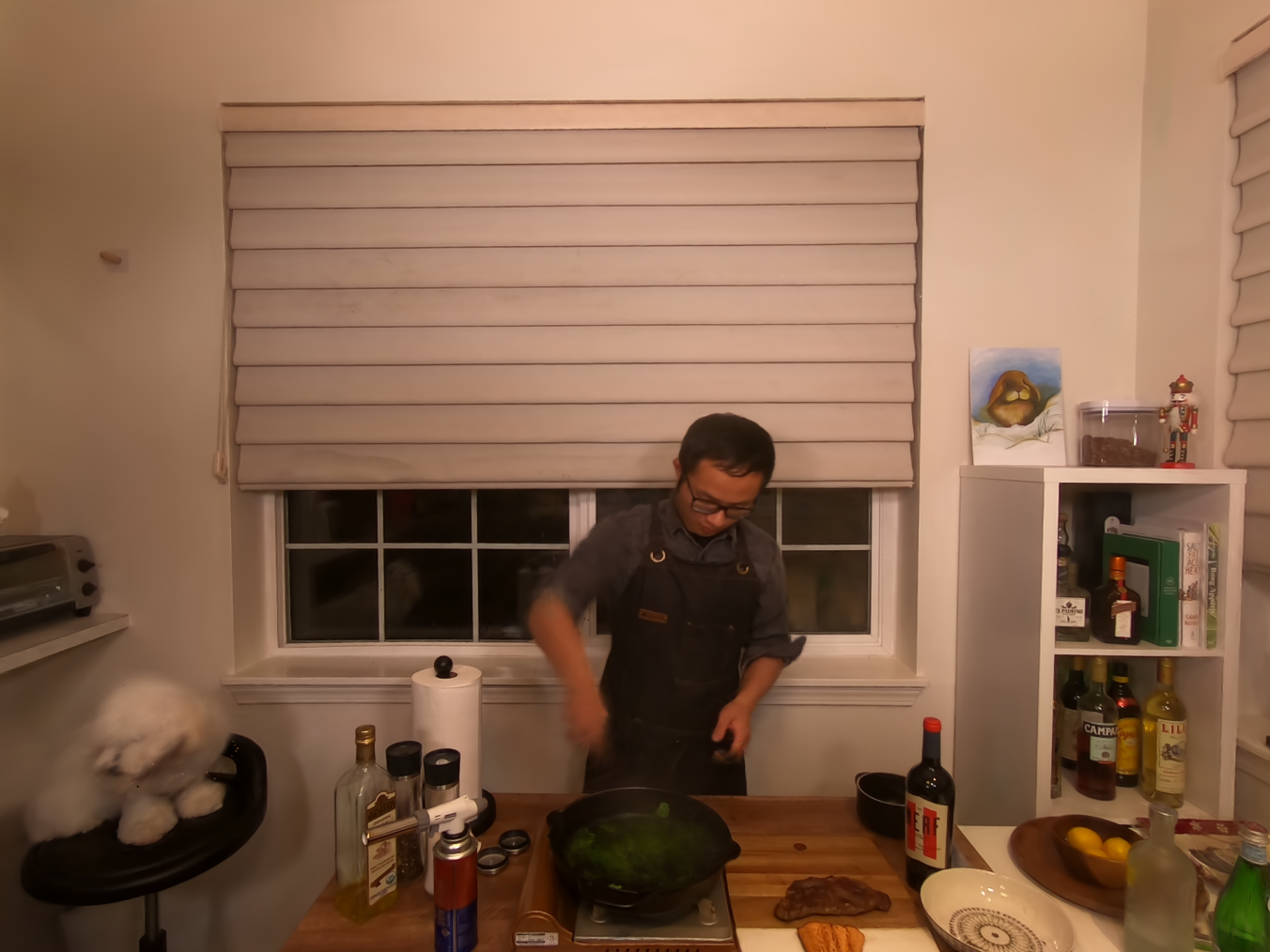} &
			\includegraphics[width=0.17\textwidth]{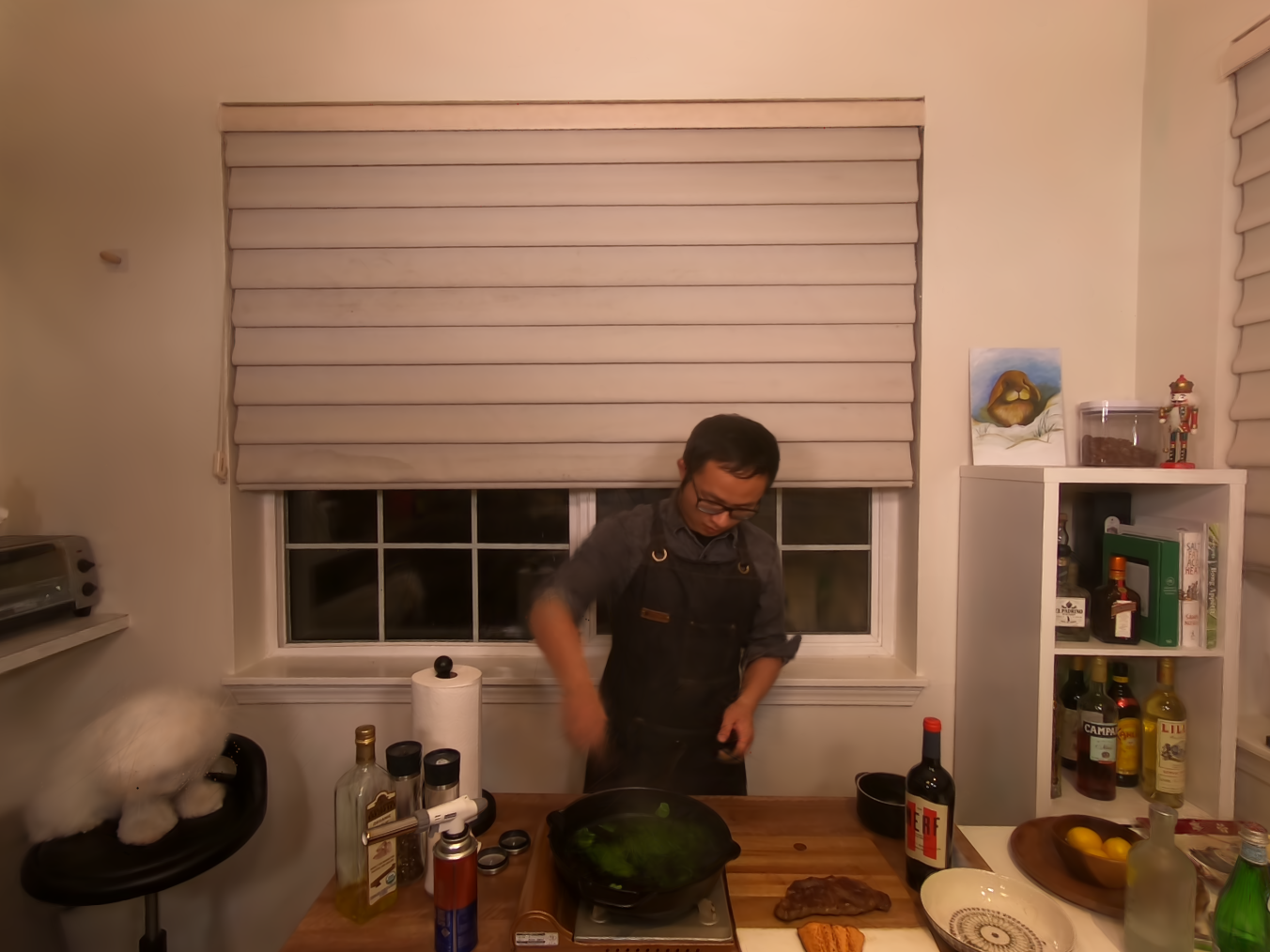} &
			\includegraphics[width=0.17\textwidth]{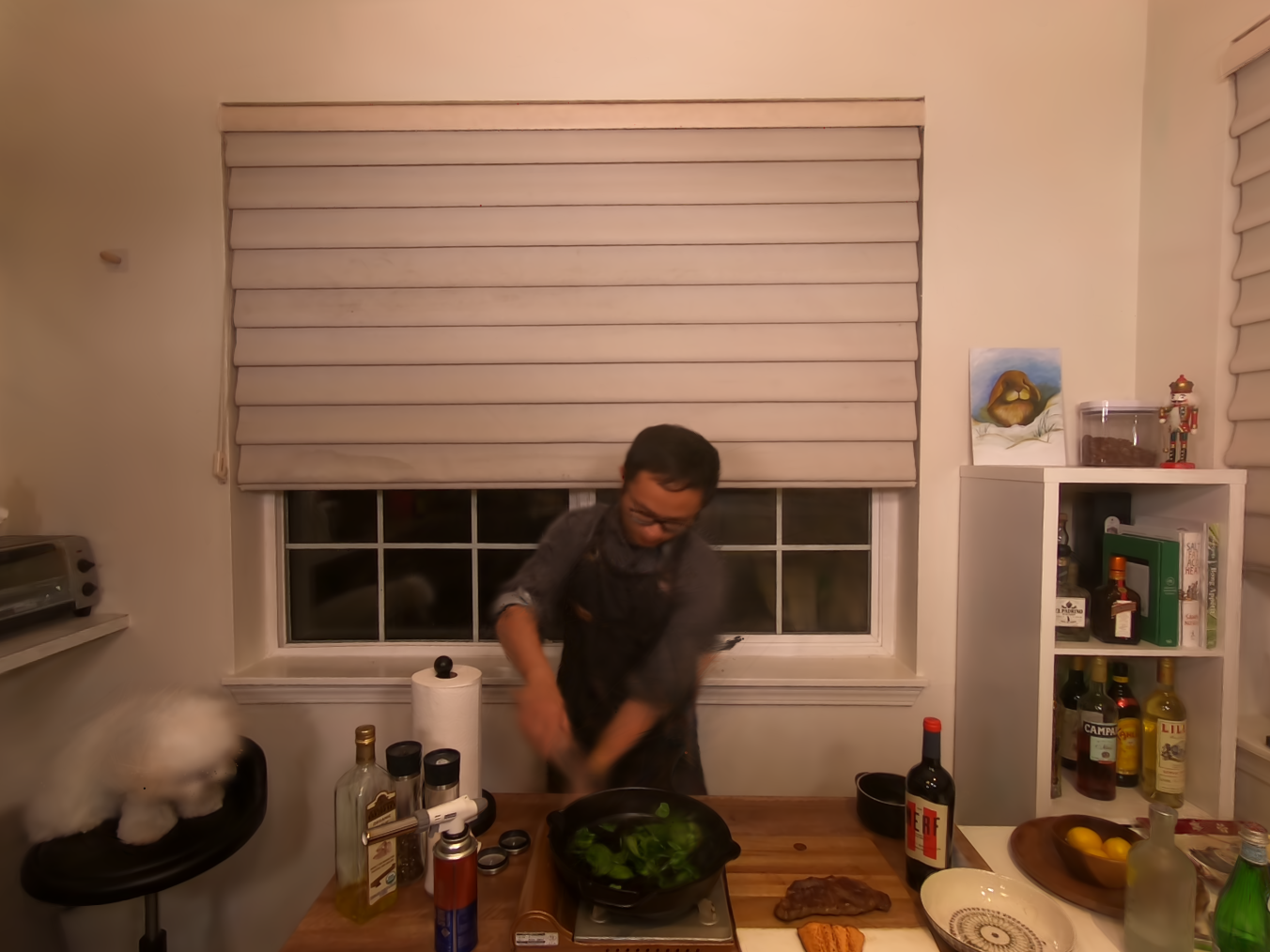} &
			\includegraphics[width=0.17\textwidth]{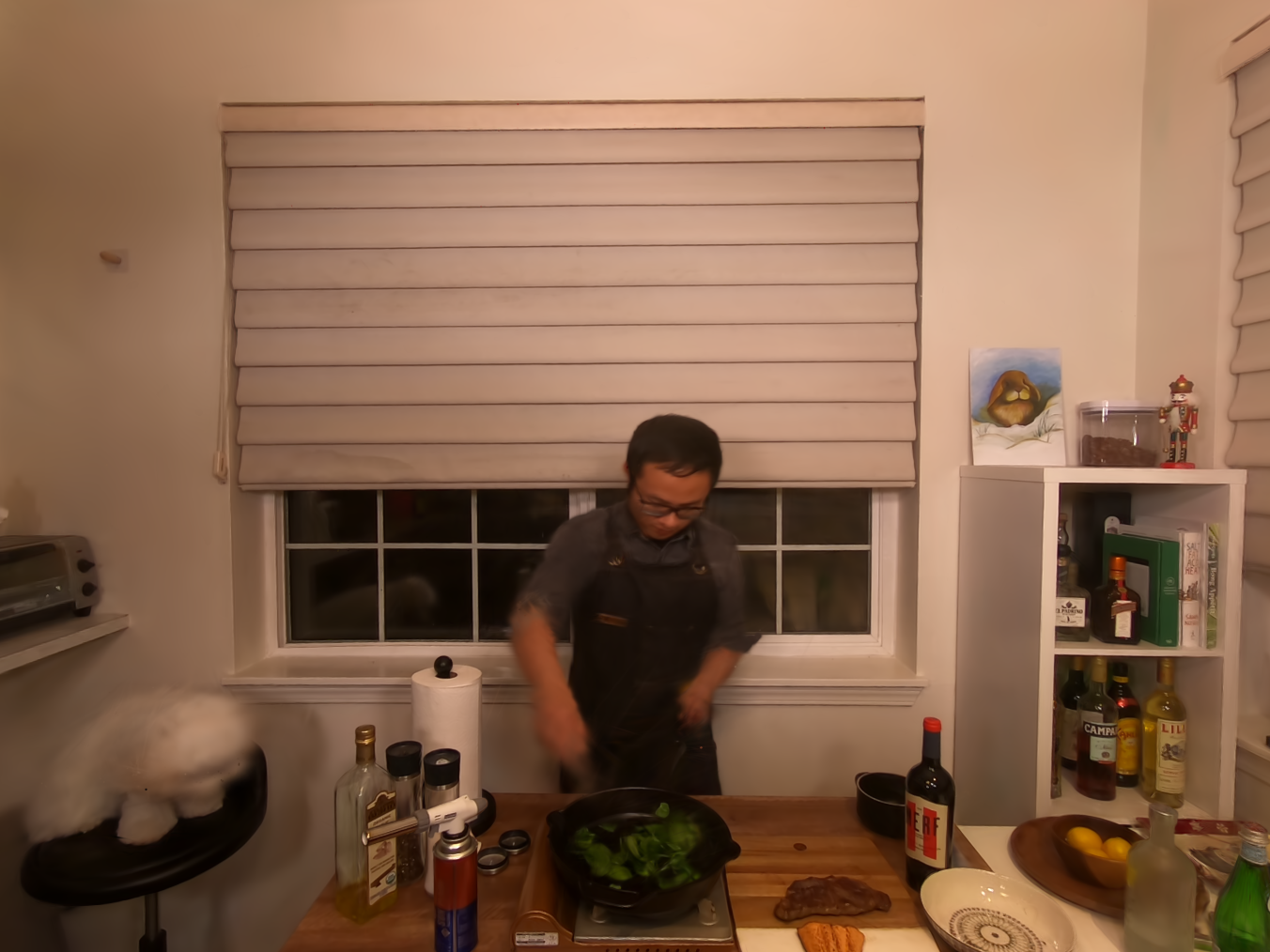} &
			\includegraphics[width=0.17\textwidth]{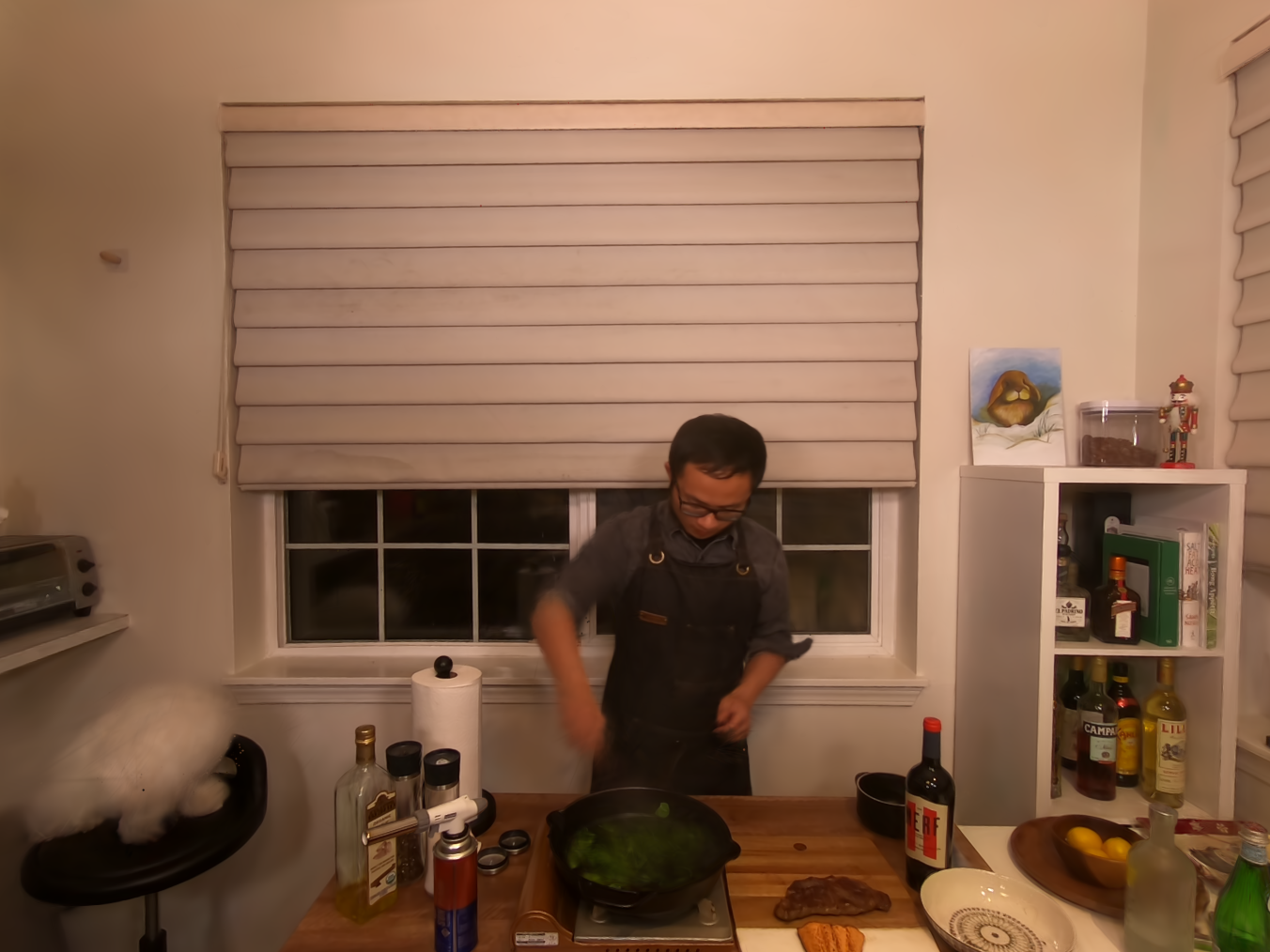} &

		\end{tabular}
	}
	\vspace{-5pt}
	\caption{Results on Neu3DV dataset.
	}\label{fig:dynerf}
\end{figure*}

\begin{figure*}
    \centering
    \addtolength{\tabcolsep}{-6.5pt}
    \footnotesize{
        \setlength{\tabcolsep}{1pt} % Default value: 6pt
        \begin{tabular}{p{8.2pt}ccccccc}
            & GT & Ours & NDVG~\cite{guo2022_NDVG_arxiv} & 4D-GS~\cite{4DGaussian} & FDNeRF~\cite{Guo2023_Forward_ICCV} & TiNeuVox-B~\cite{fang2022_TANV_arxiv}  \\
        \raisebox{35pt}{\rotatebox[origin=c]{90}{Hell Warrior}}&
             % \raisebox{20pt}{\rotatebox[origin=c]{90}{Ours}}&
             \includegraphics[width=0.15\textwidth]{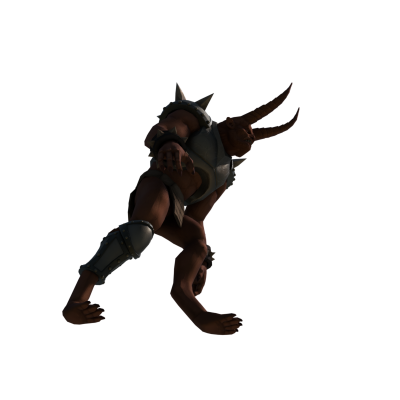} &
            \includegraphics[width=0.15\textwidth]{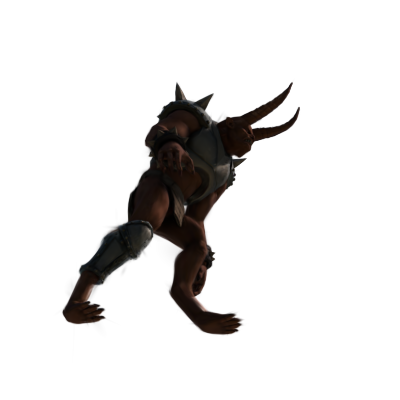} &
            \includegraphics[width=0.15\textwidth]{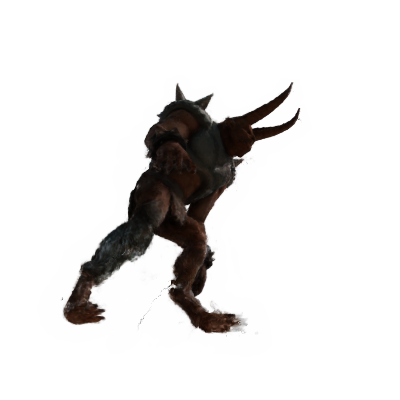} &
            \includegraphics[width=0.15\textwidth]{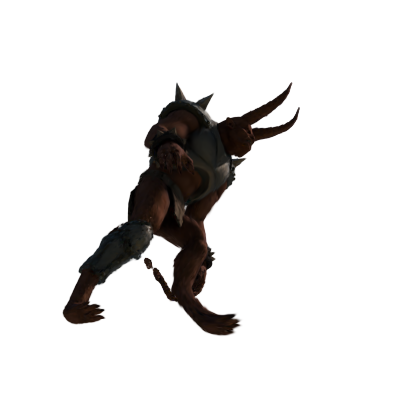} &
            \includegraphics[width=0.15\textwidth]{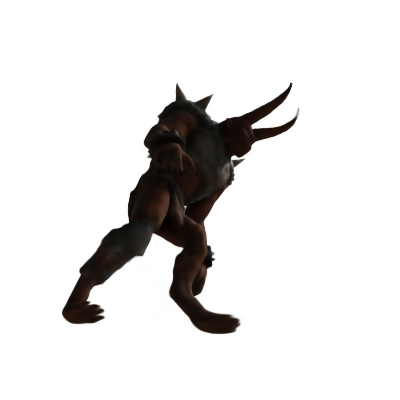} &
            \includegraphics[width=0.15\textwidth]{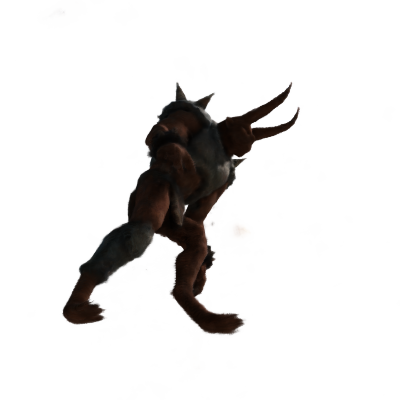} 
             \\
        \raisebox{35pt}{\rotatebox[origin=c]{90}{Hook}}&
             \includegraphics[width=0.15\textwidth]{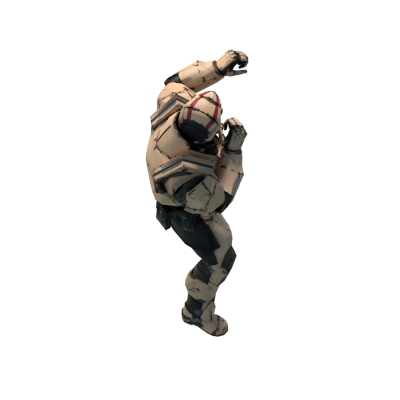} &
            \includegraphics[width=0.15\textwidth]{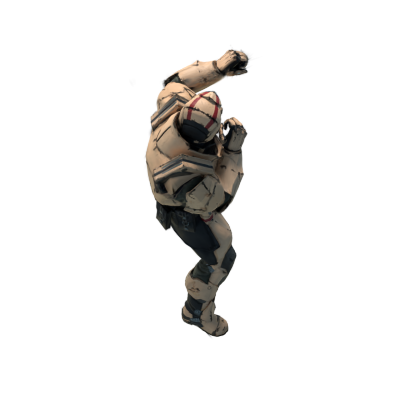} &
            \includegraphics[width=0.15\textwidth]{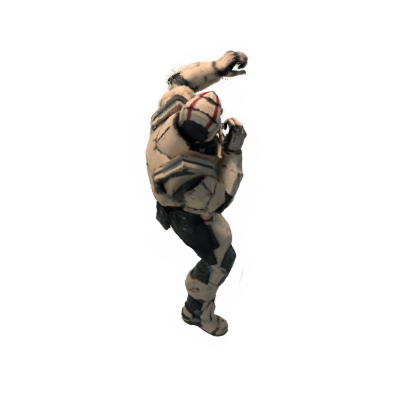} &
            \includegraphics[width=0.15\textwidth]{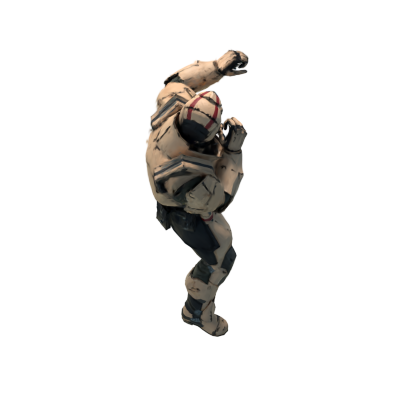} &
            \includegraphics[width=0.15\textwidth]{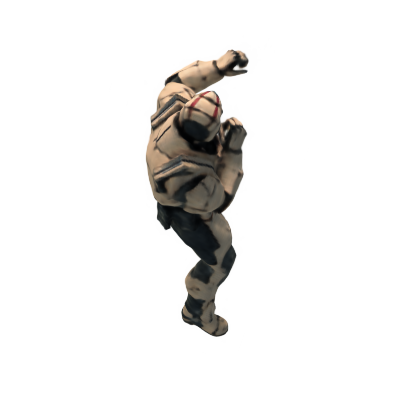} &
            \includegraphics[width=0.15\textwidth]{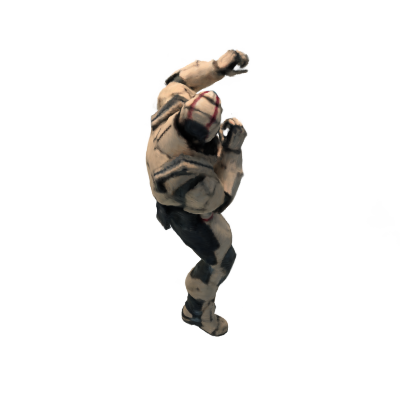} 
            \\
            \raisebox{35pt}{\rotatebox[origin=c]{90}{Jumping Jacks}}&
             \includegraphics[width=0.15\textwidth]{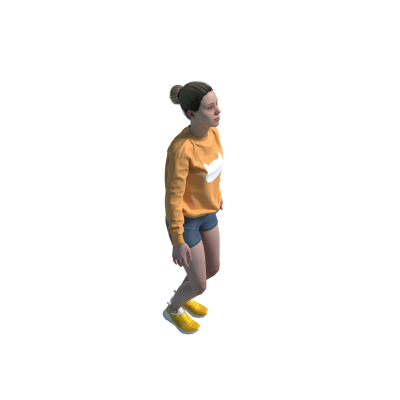} &
            \includegraphics[width=0.15\textwidth]{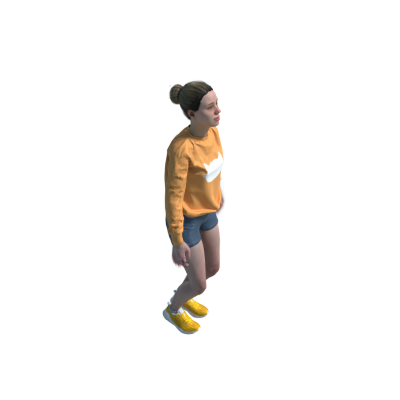} &
            \includegraphics[width=0.15\textwidth]{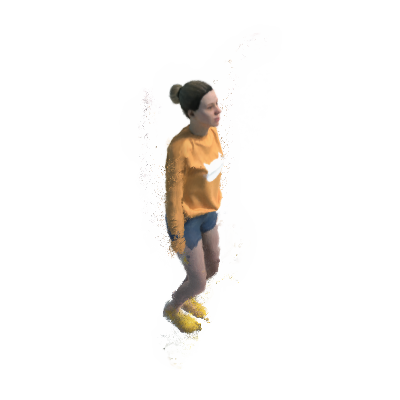} &
            \includegraphics[width=0.15\textwidth]{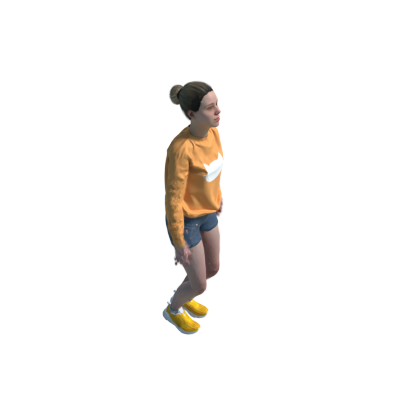} &
            \includegraphics[width=0.15\textwidth]{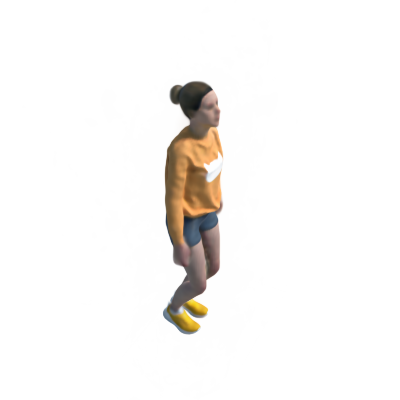} &
            \includegraphics[width=0.15\textwidth]{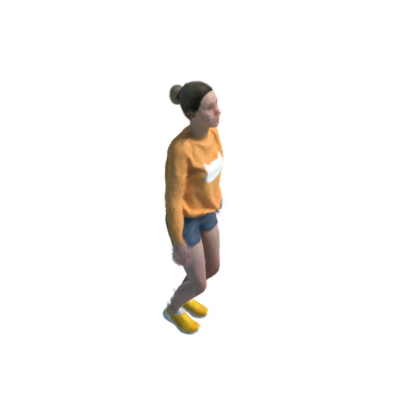} 

            \\
            \raisebox{35pt}{\rotatebox[origin=c]{90}{Lego}}&
             \includegraphics[width=0.15\textwidth]{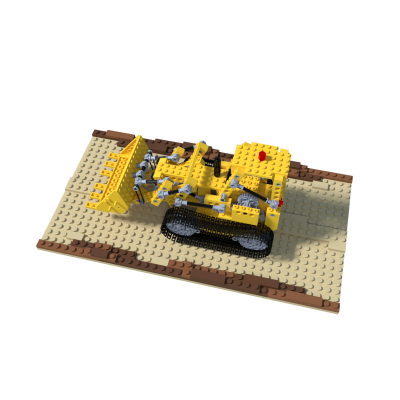} &
            \includegraphics[width=0.15\textwidth]{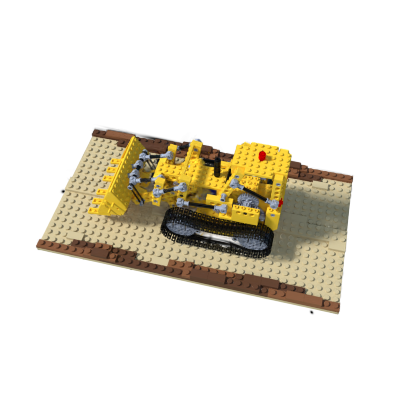} &
            \includegraphics[width=0.15\textwidth]{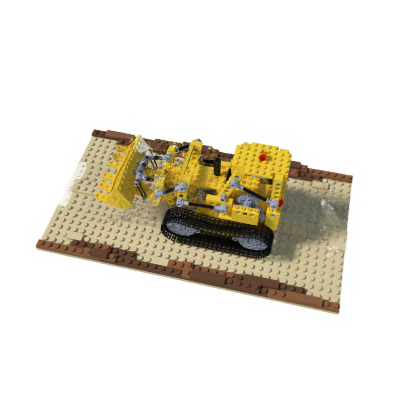} &
            \includegraphics[width=0.15\textwidth]{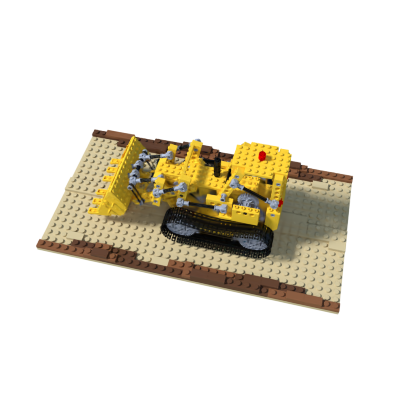} &
            \includegraphics[width=0.15\textwidth]{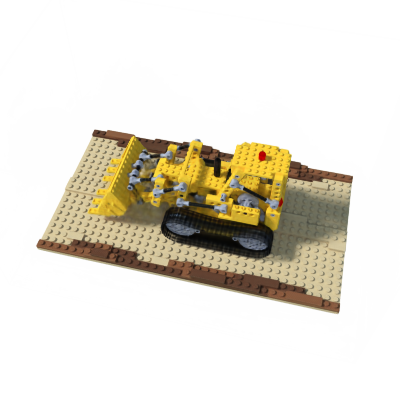} &
            \includegraphics[width=0.15\textwidth]{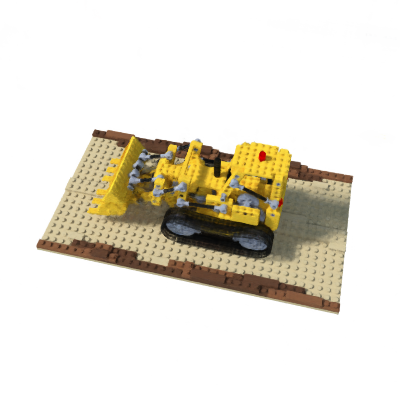} 

            \\
            \raisebox{35pt}{\rotatebox[origin=c]{90}{Mutant}}&
            \includegraphics[width=0.15\textwidth]{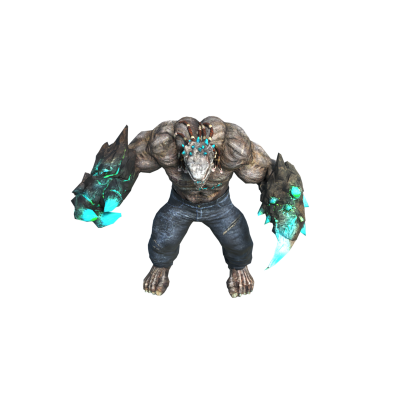} &
            \includegraphics[width=0.15\textwidth]{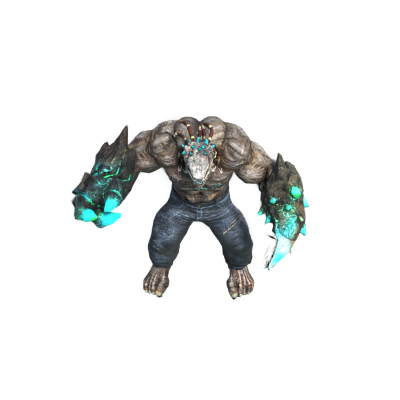} &
            \includegraphics[width=0.15\textwidth]{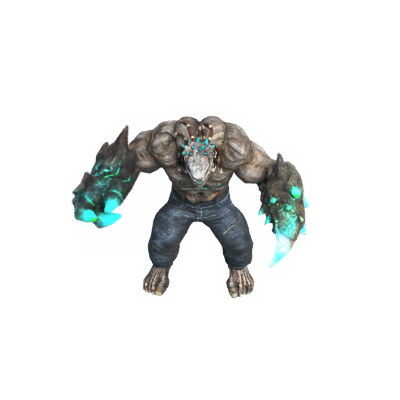} &
            \includegraphics[width=0.15\textwidth]{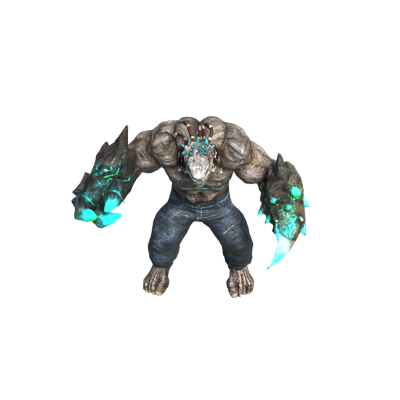} &
            \includegraphics[width=0.15\textwidth]{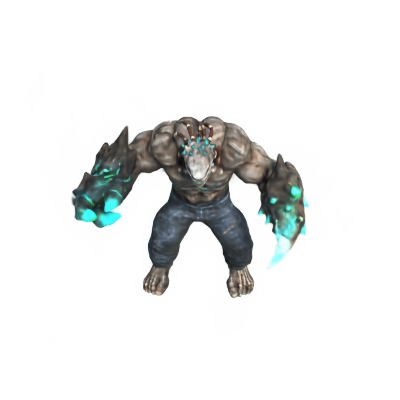} &
            \includegraphics[width=0.15\textwidth]{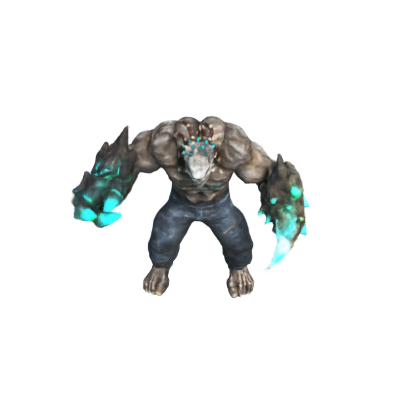} 

            \\
            \raisebox{35pt}{\rotatebox[origin=c]{90}{Stand Up}}&
             \includegraphics[width=0.15\textwidth]{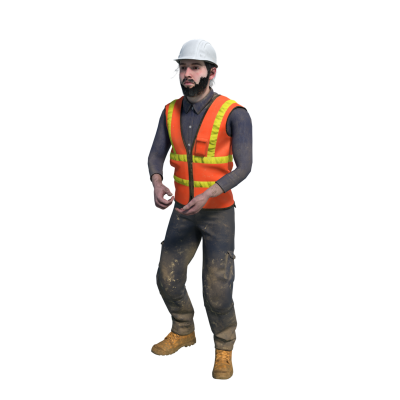} &
            \includegraphics[width=0.15\textwidth]{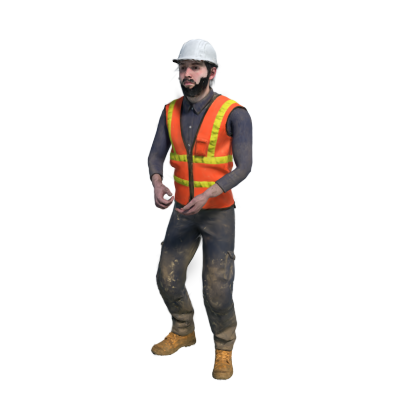} &
            \includegraphics[width=0.15\textwidth]{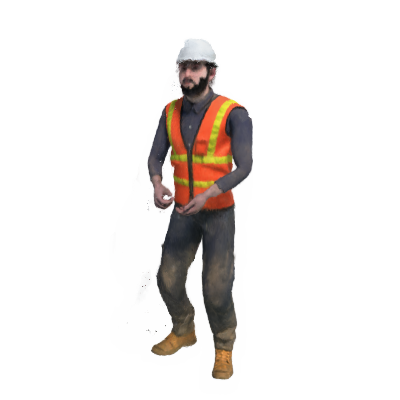} &
            \includegraphics[width=0.15\textwidth]{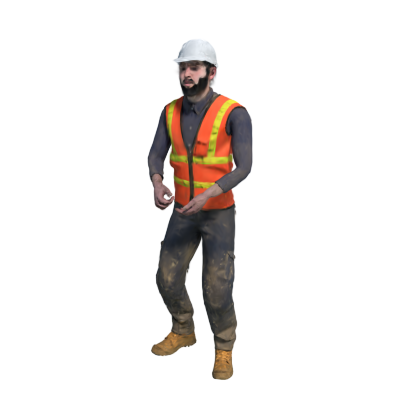} &
            \includegraphics[width=0.15\textwidth]{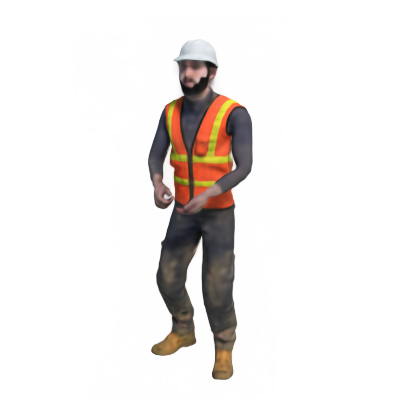} &
            \includegraphics[width=0.15\textwidth]{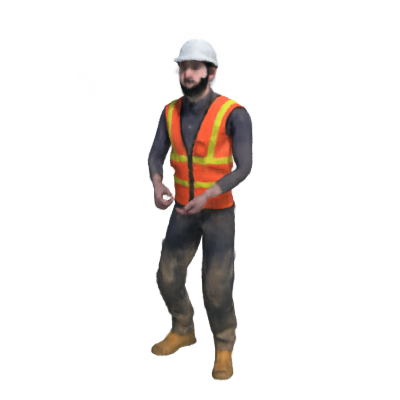} 

            \\
            \raisebox{35pt}{\rotatebox[origin=c]{90}{T-Rex}}&
             \includegraphics[width=0.15\textwidth]{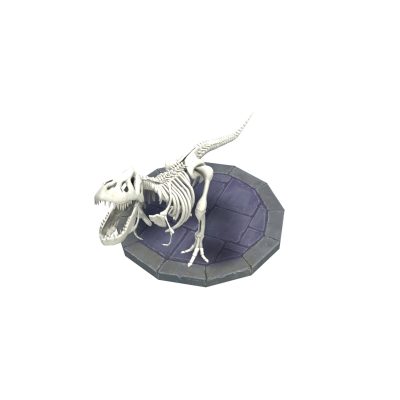} &
            \includegraphics[width=0.15\textwidth]{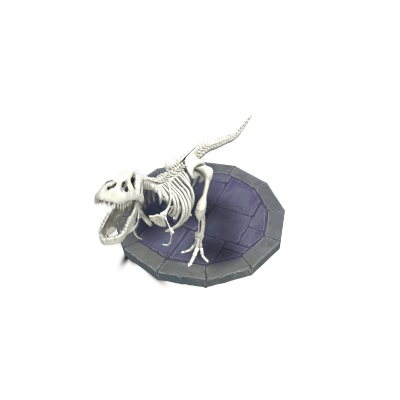} &
            \includegraphics[width=0.15\textwidth]{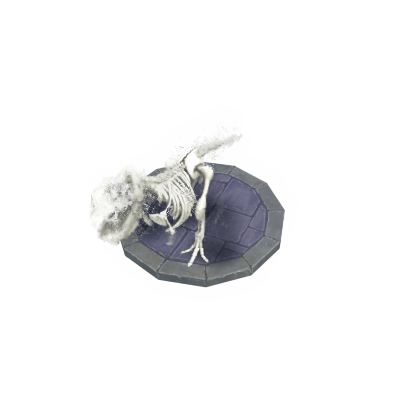} &
            \includegraphics[width=0.15\textwidth]{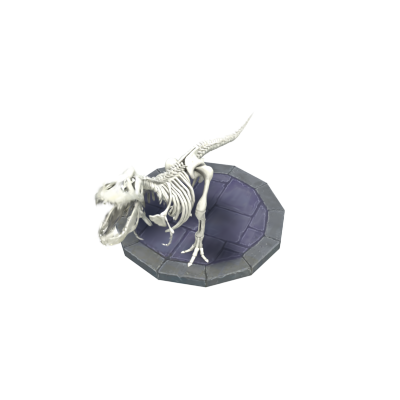} &
            \includegraphics[width=0.15\textwidth]{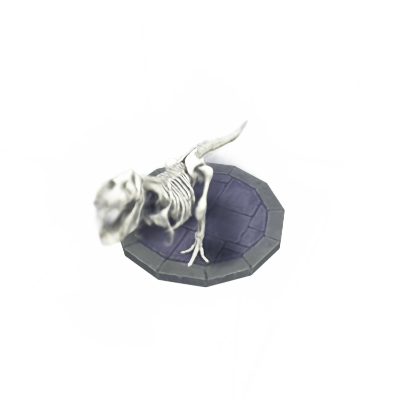} &
            \includegraphics[width=0.15\textwidth]{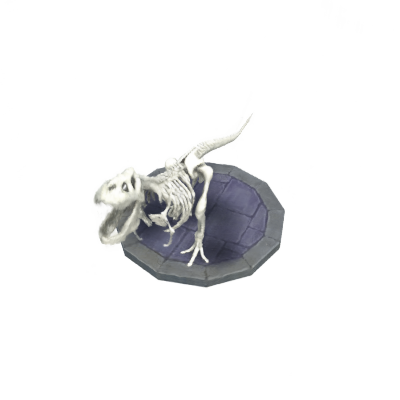} 
            
            \\
            \raisebox{35pt}{\rotatebox[origin=c]{90}{Bouncing Balls}}&
            \includegraphics[width=0.15\textwidth]{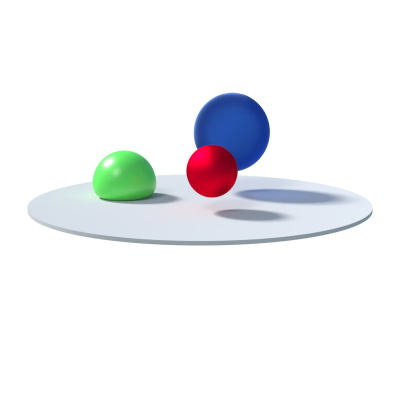} &
            \includegraphics[width=0.15\textwidth]{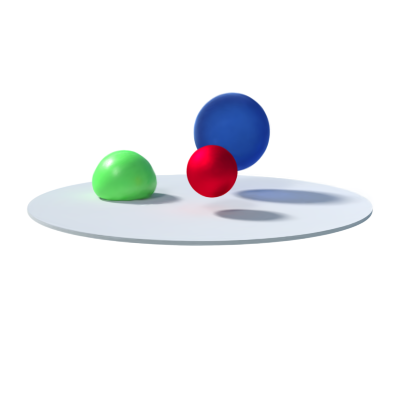} &
            \includegraphics[width=0.15\textwidth]{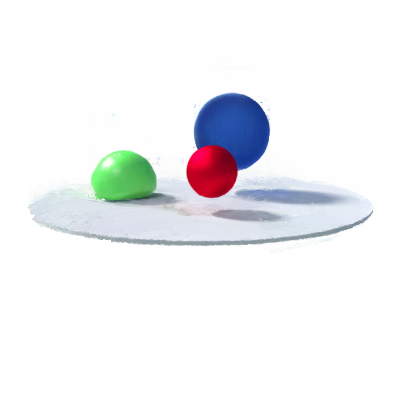} &
            \includegraphics[width=0.15\textwidth]{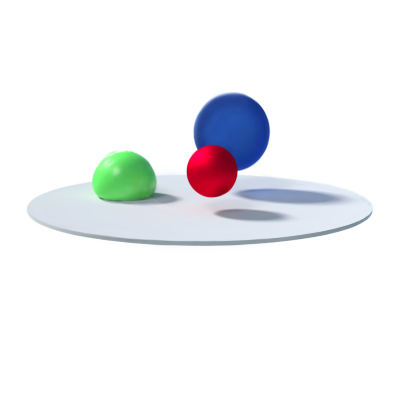} &
            \includegraphics[width=0.15\textwidth]{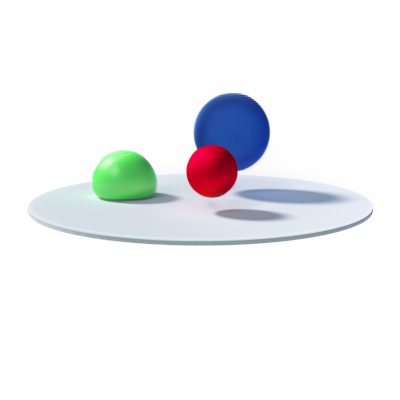} &
            \includegraphics[width=0.15\textwidth]{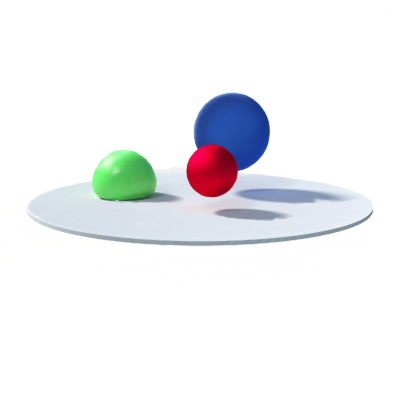}

        \end{tabular}
    }
\vspace{-5pt}
	\caption{\textbf{Qualitative comparison on the D-NeRF synthetic dataset.} We show synthesized images on the D-NeRF synthetic dataset of our method and other competing methods.
}\label{fig:dnerf-quality}
\end{figure*}

\begin{figure*}
    \centering
    \addtolength{\tabcolsep}{-6.5pt}
    \footnotesize{
        \setlength{\tabcolsep}{1pt} % Default value: 6pt
        \begin{tabular}{p{8.2pt}ccccccc}
            & GT & Ours & NDVG~\cite{guo2022_NDVG_arxiv} & 3D-GS~\cite{3Dgaussian} & FDNeRF~\cite{Guo2023_Forward_ICCV} & TiNeuVox-B~\cite{fang2022_TANV_arxiv}  \\
        \raisebox{60pt}{\rotatebox[origin=c]{90}{Broom}}&
             % \raisebox{20pt}{\rotatebox[origin=c]{90}{Ours}}&
             \includegraphics[width=0.15\textwidth]{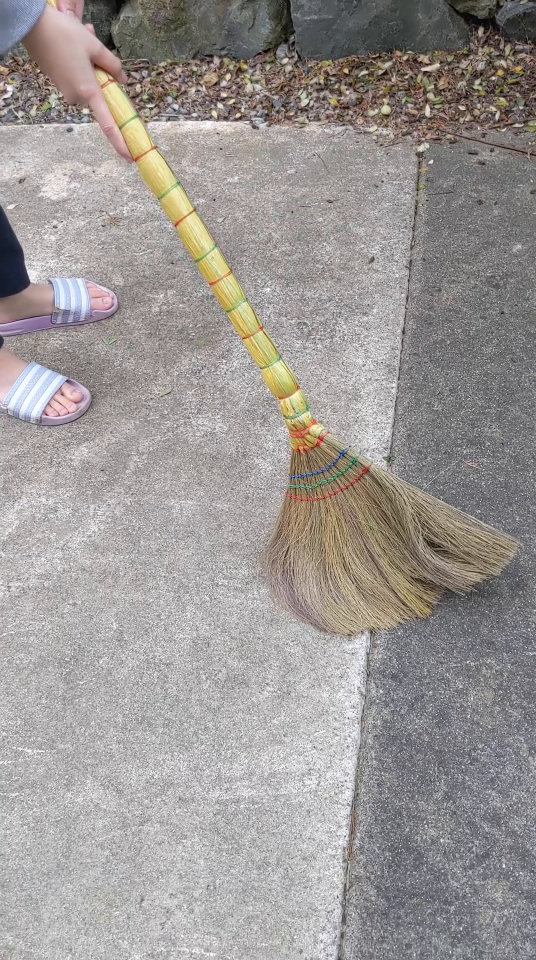} &
            \includegraphics[width=0.15\textwidth]{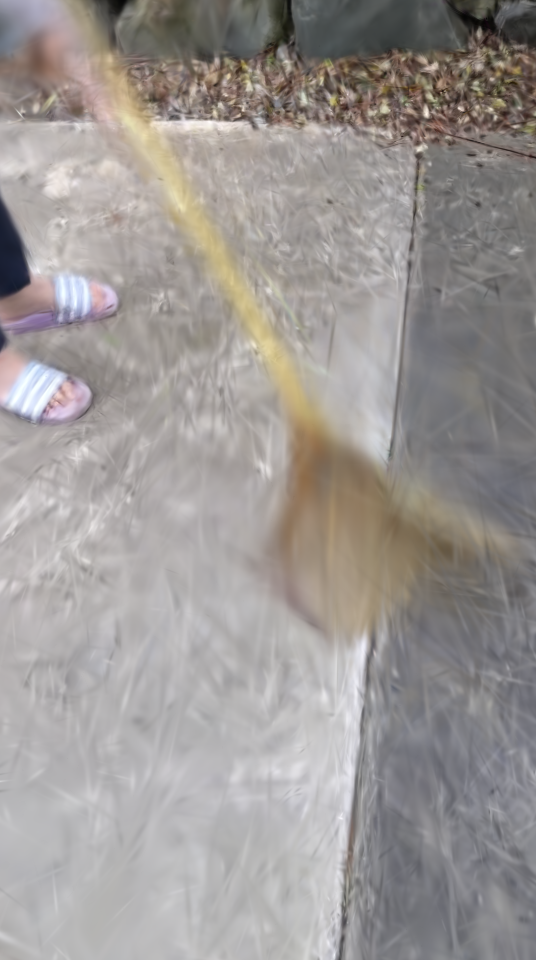} &
            \includegraphics[width=0.15\textwidth]{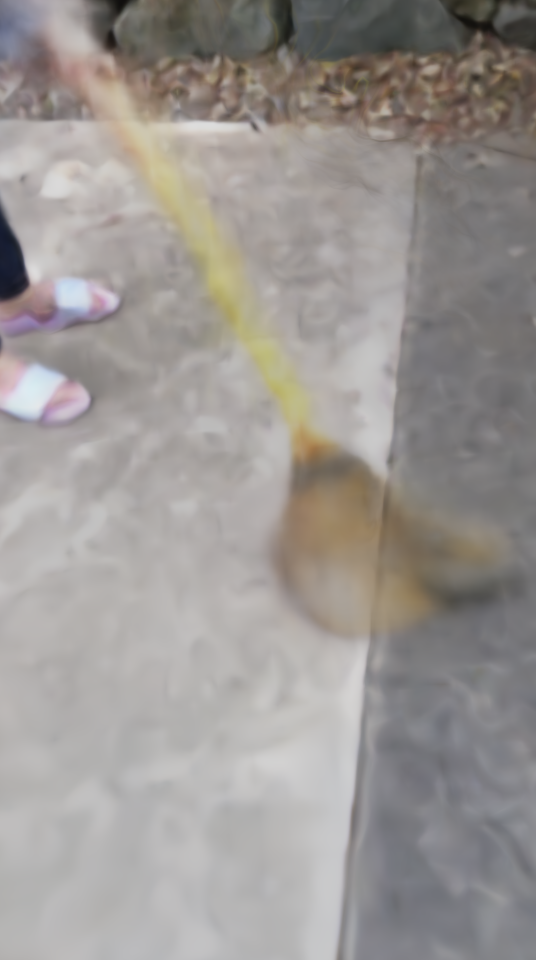} &
            \includegraphics[width=0.15\textwidth]{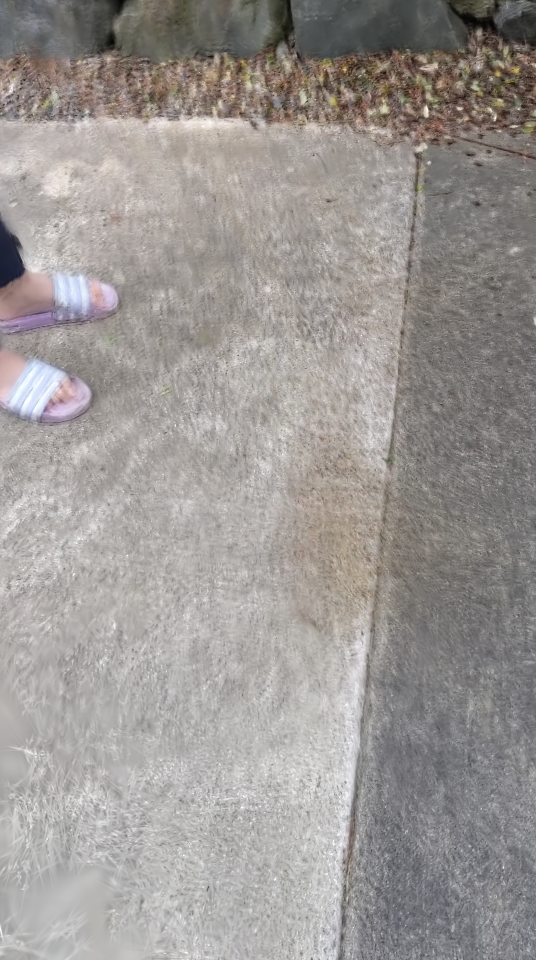} &
            \includegraphics[width=0.15\textwidth]{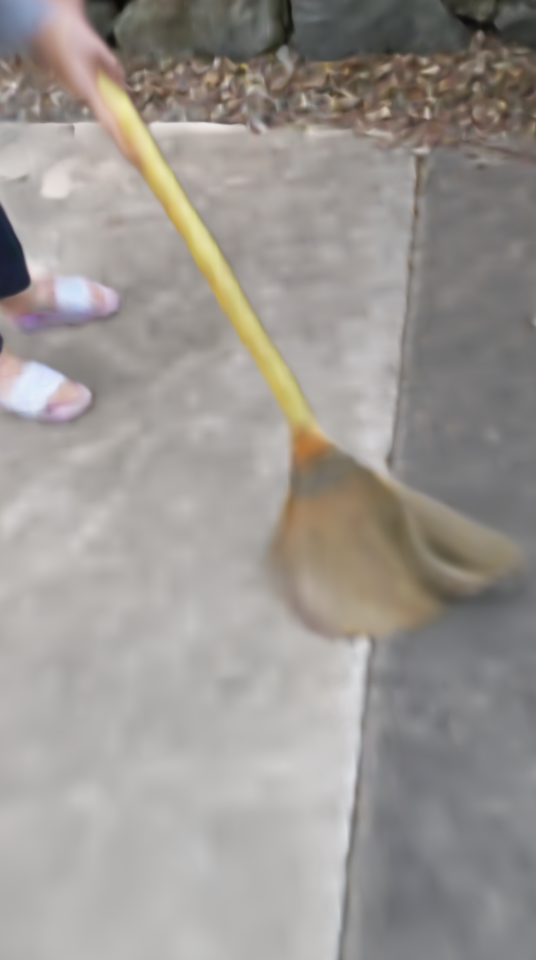} &
            \includegraphics[width=0.15\textwidth]{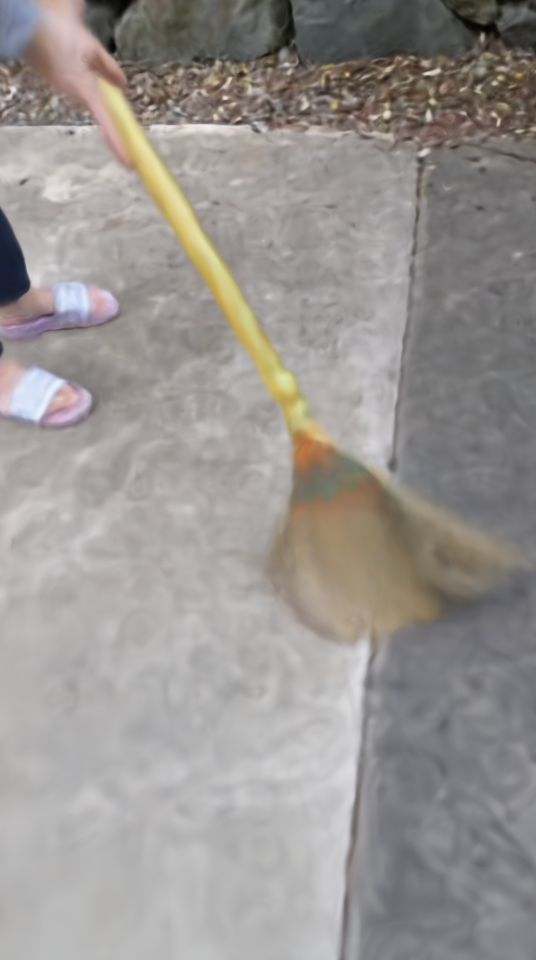}
             \\
        \raisebox{60pt}{\rotatebox[origin=c]{90}{Peel Banana}}&
              \includegraphics[width=0.15\textwidth]{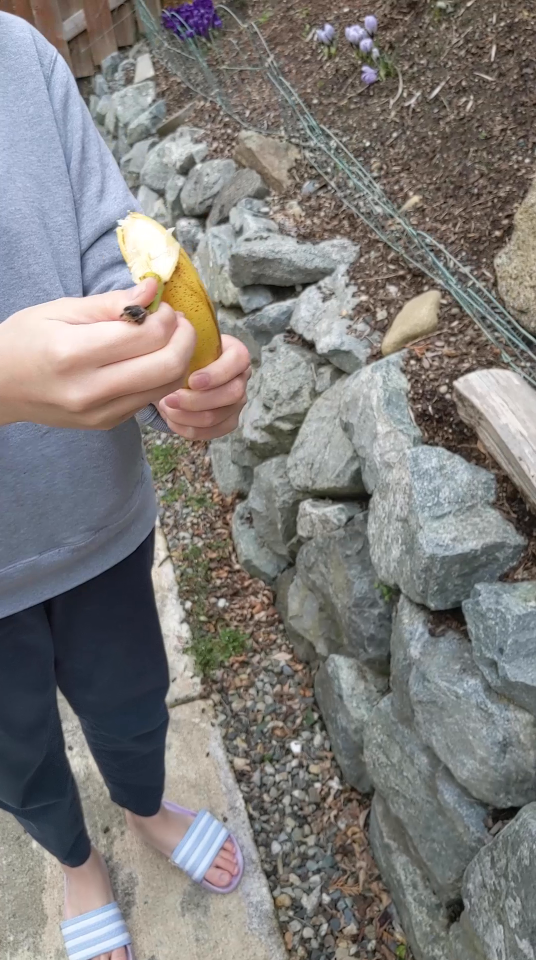} &
            \includegraphics[width=0.15\textwidth]{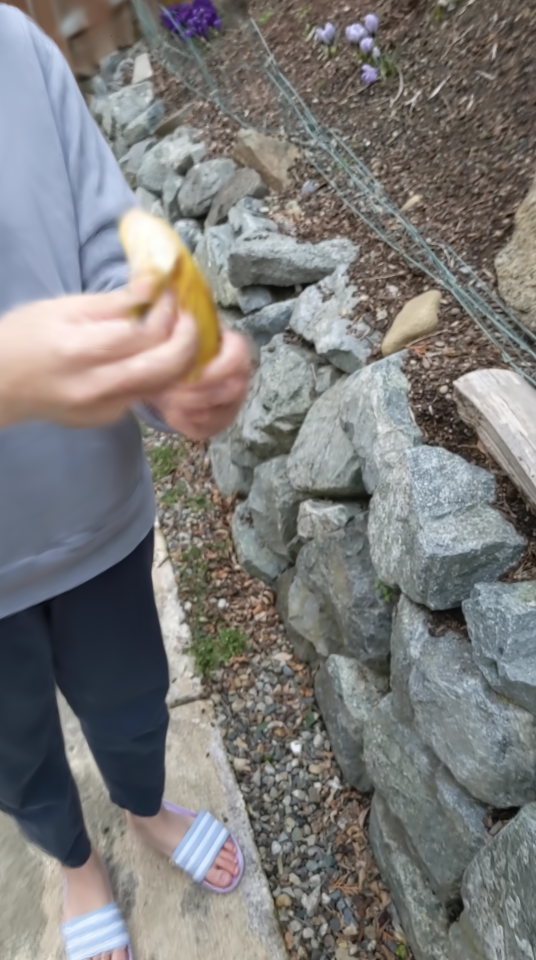} &
            \includegraphics[width=0.15\textwidth]{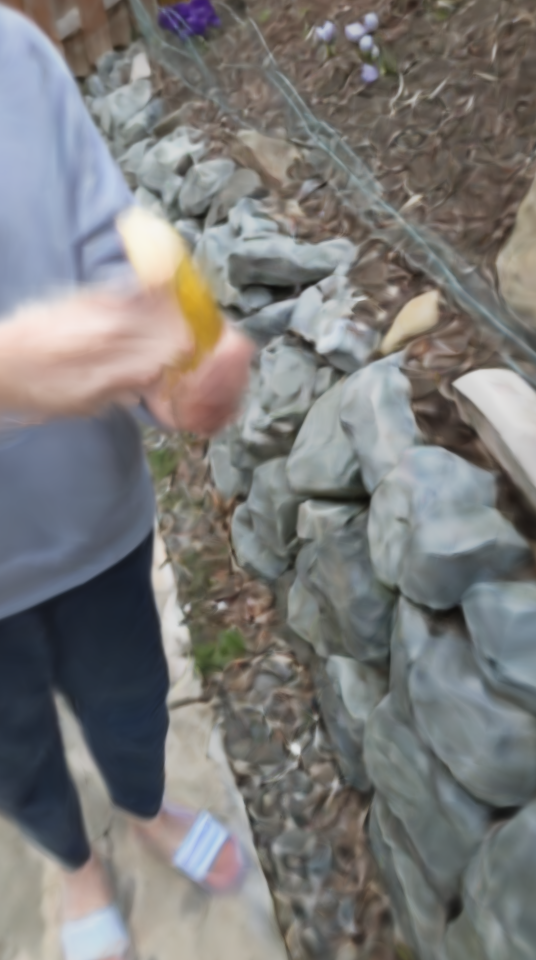} &
            \includegraphics[width=0.15\textwidth]{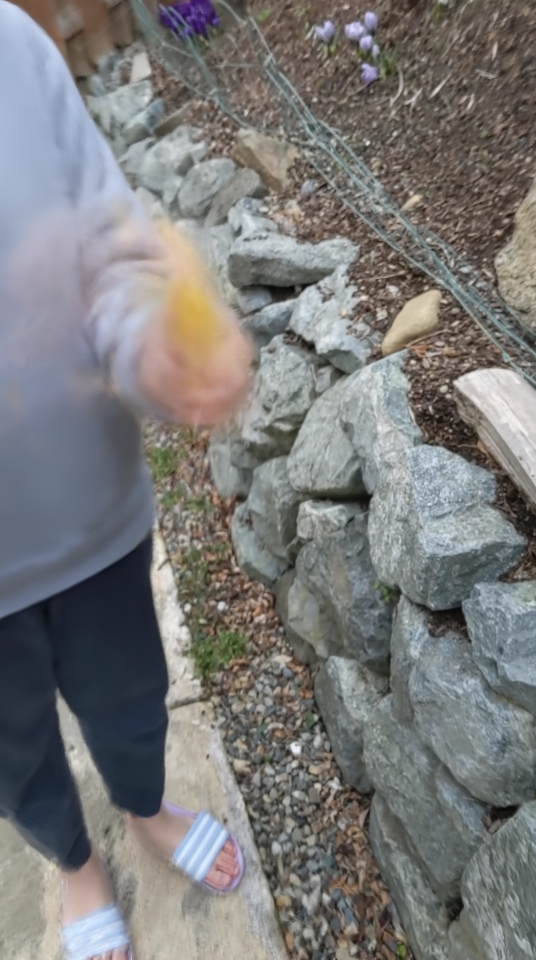} &
            \includegraphics[width=0.15\textwidth]{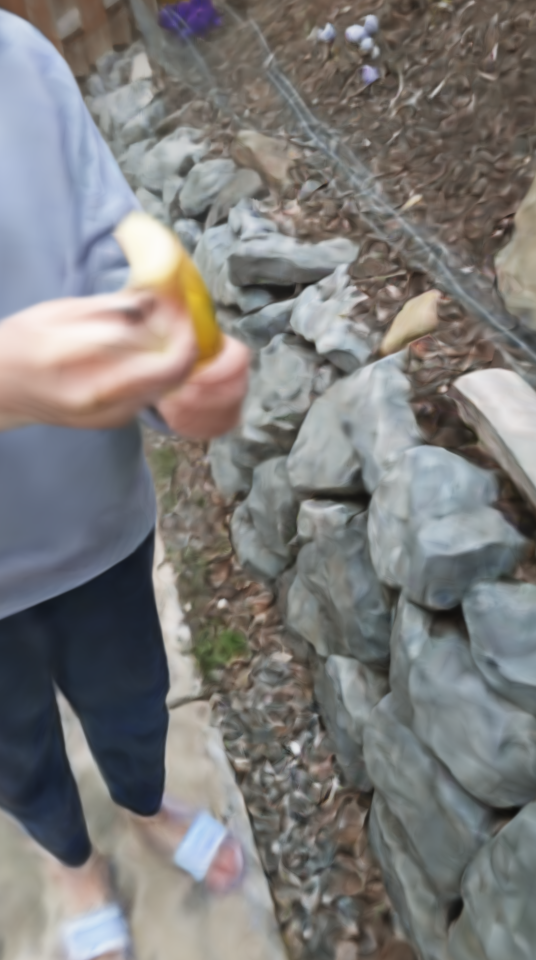} &
            \includegraphics[width=0.15\textwidth]{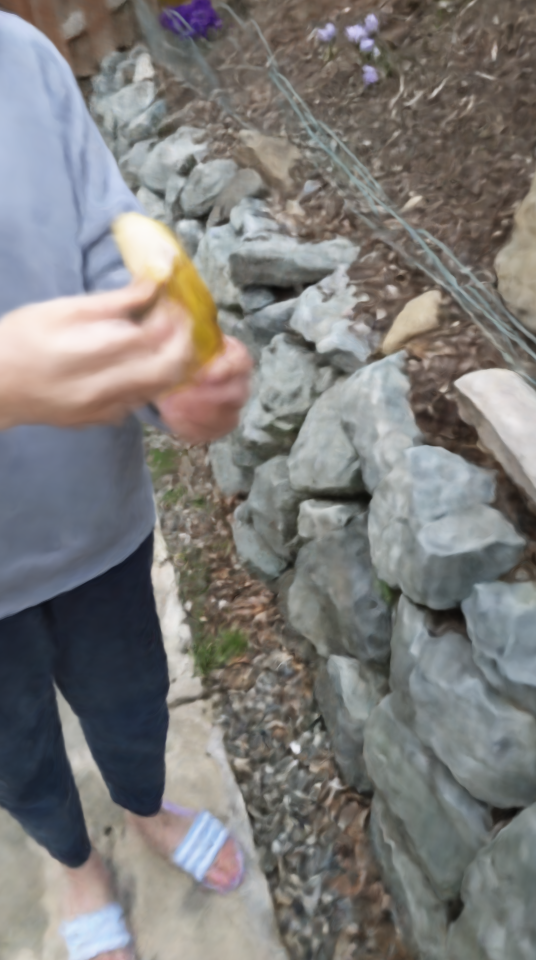}
            \\
            \raisebox{60pt}{\rotatebox[origin=c]{90}{Chicken}}&
              \includegraphics[width=0.15\textwidth]{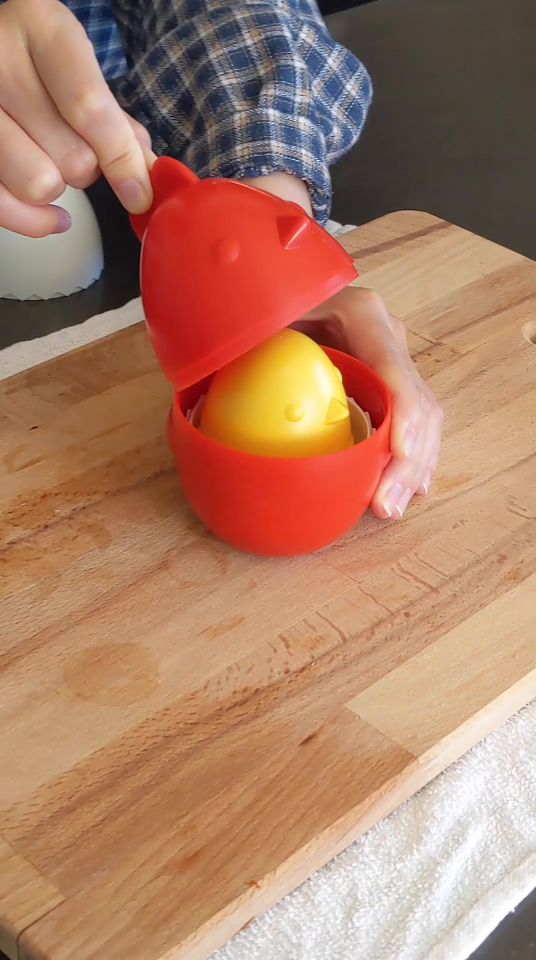} &
            \includegraphics[width=0.15\textwidth]{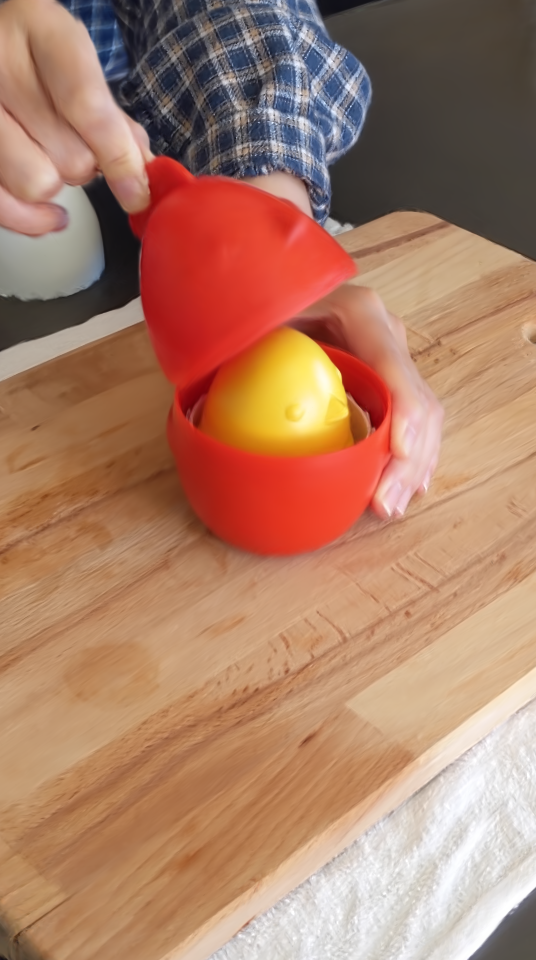} &
            \includegraphics[width=0.15\textwidth]{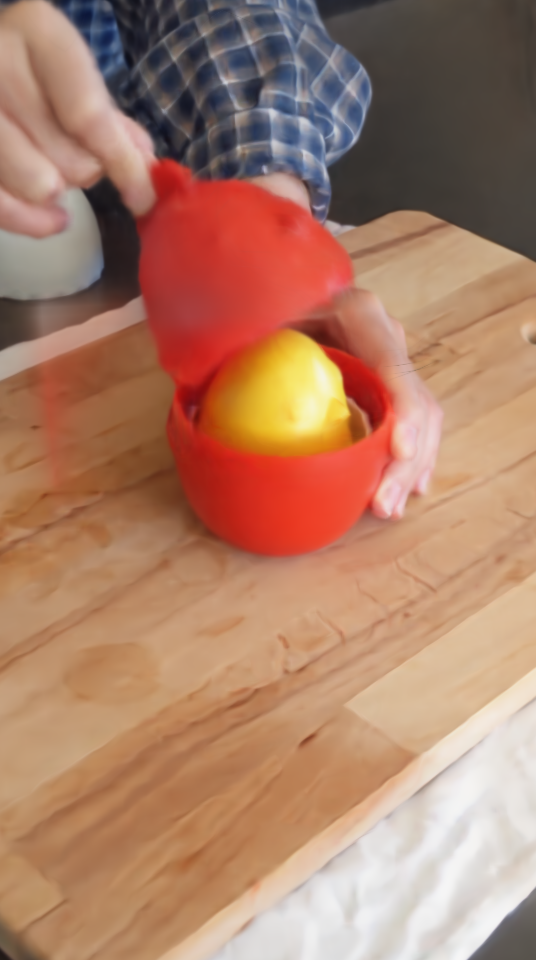} &
            \includegraphics[width=0.15\textwidth]{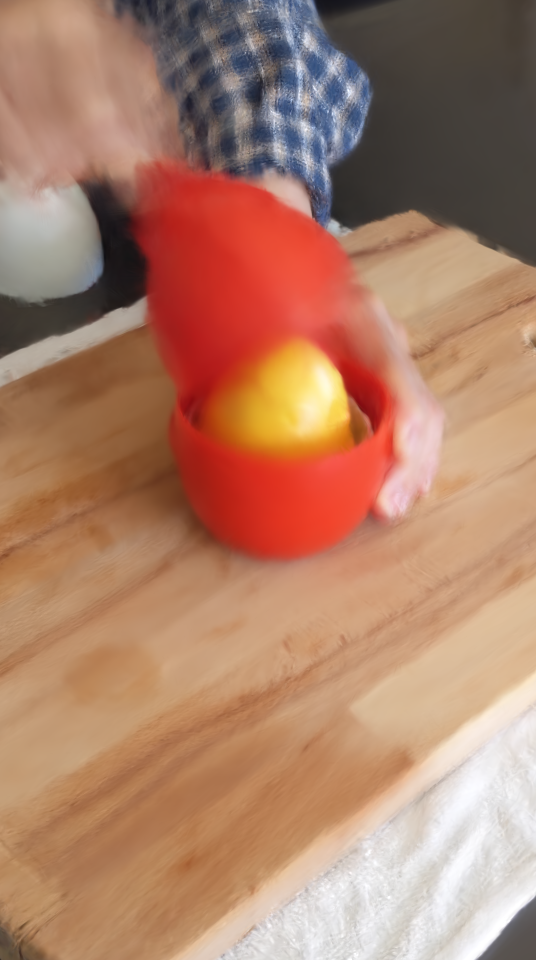} &
            \includegraphics[width=0.15\textwidth]{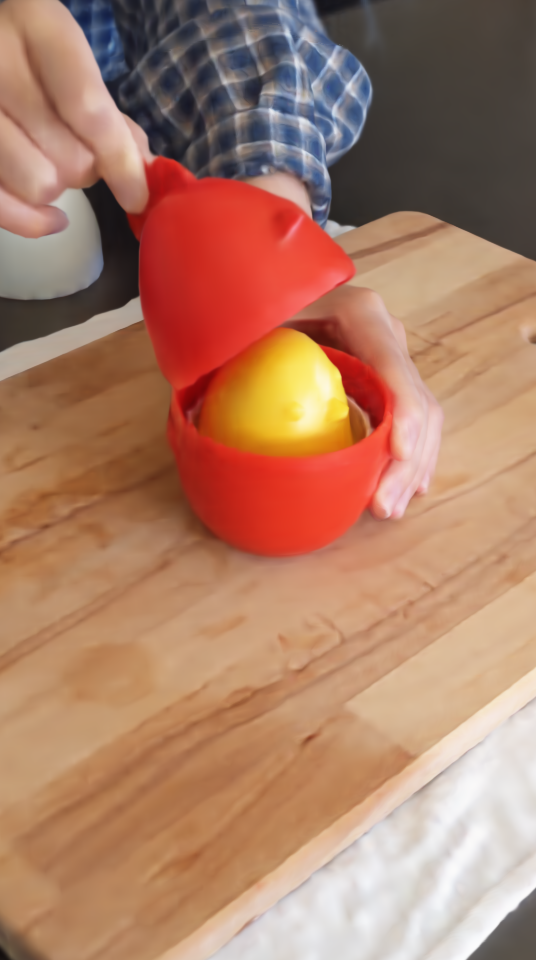} &
            \includegraphics[width=0.15\textwidth]{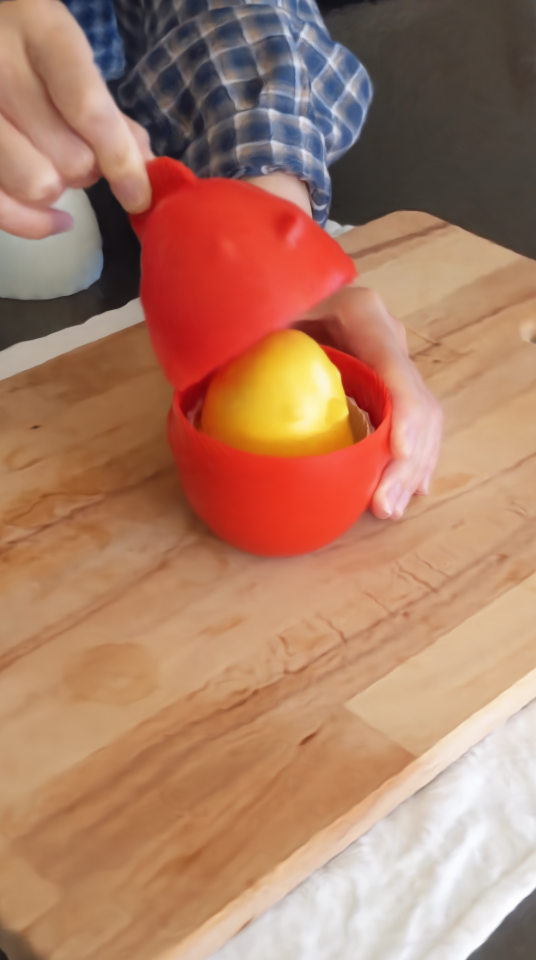}

            \\
            \raisebox{60pt}{\rotatebox[origin=c]{90}{3D Printer}}&
              \includegraphics[width=0.15\textwidth]{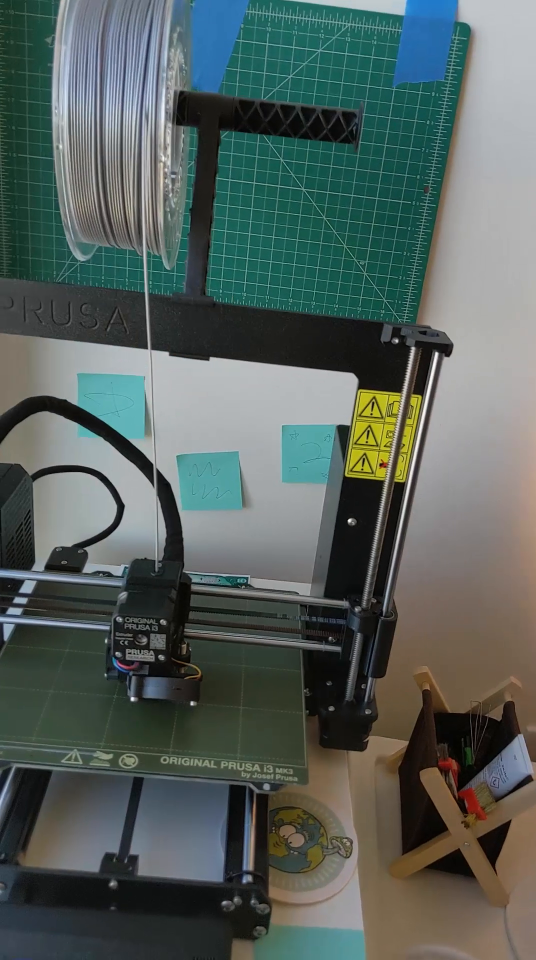} &
            \includegraphics[width=0.15\textwidth]{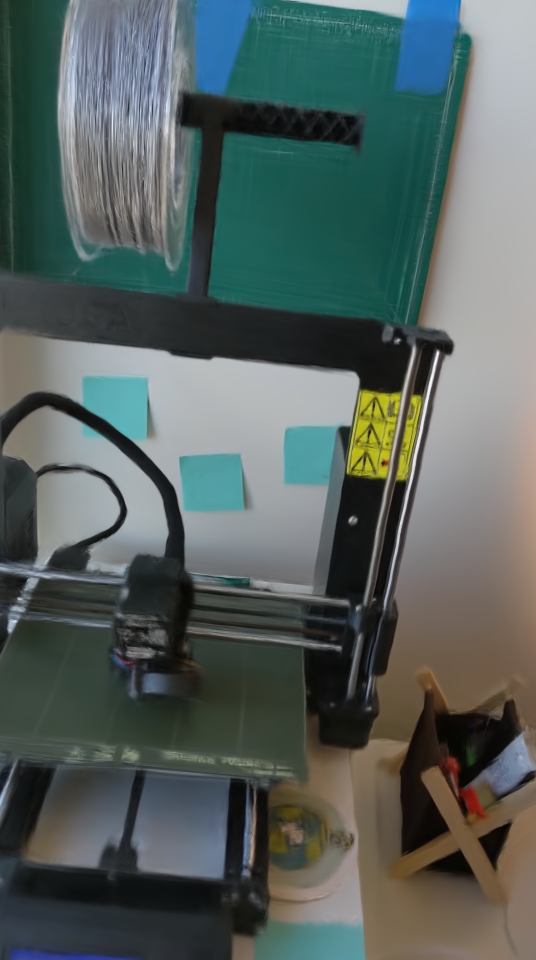} &
            \includegraphics[width=0.15\textwidth]{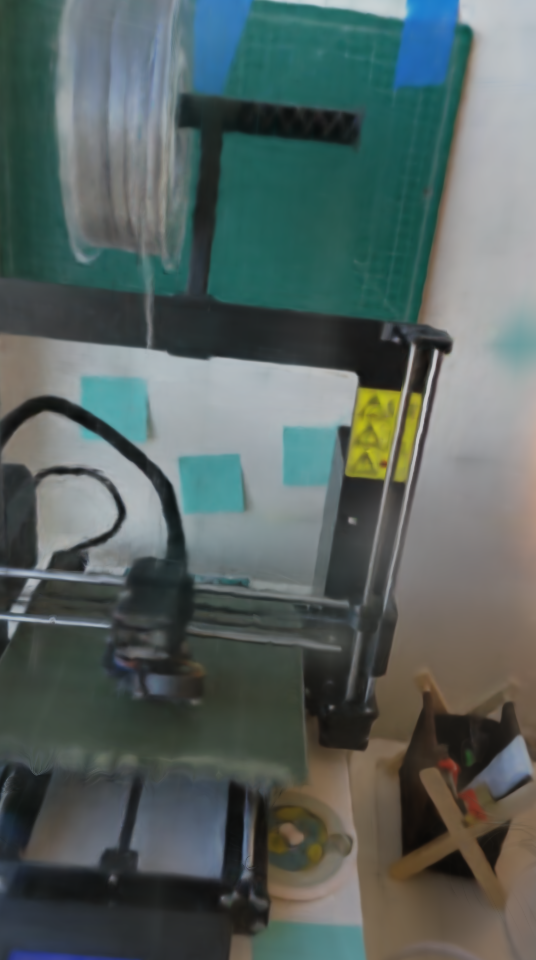} &
            \includegraphics[width=0.15\textwidth]{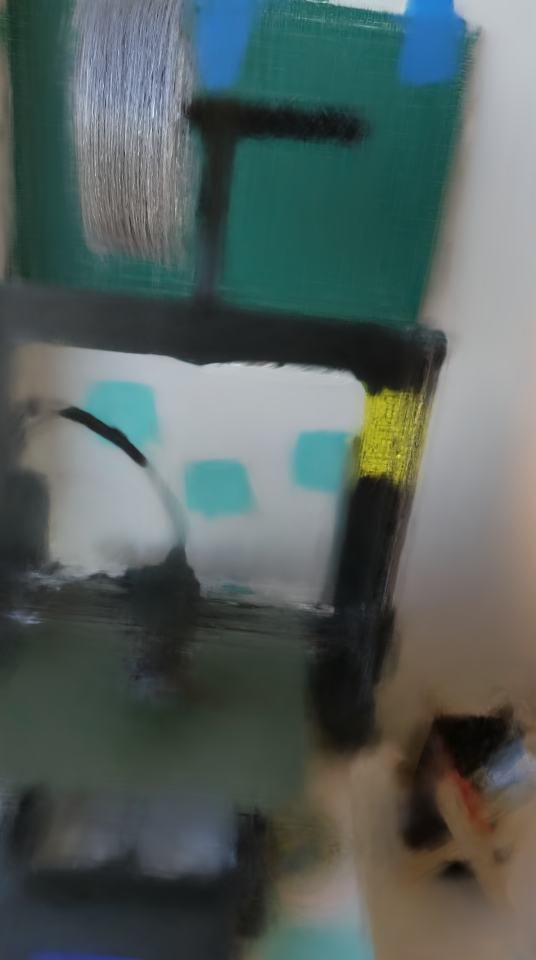} &
            \includegraphics[width=0.15\textwidth]{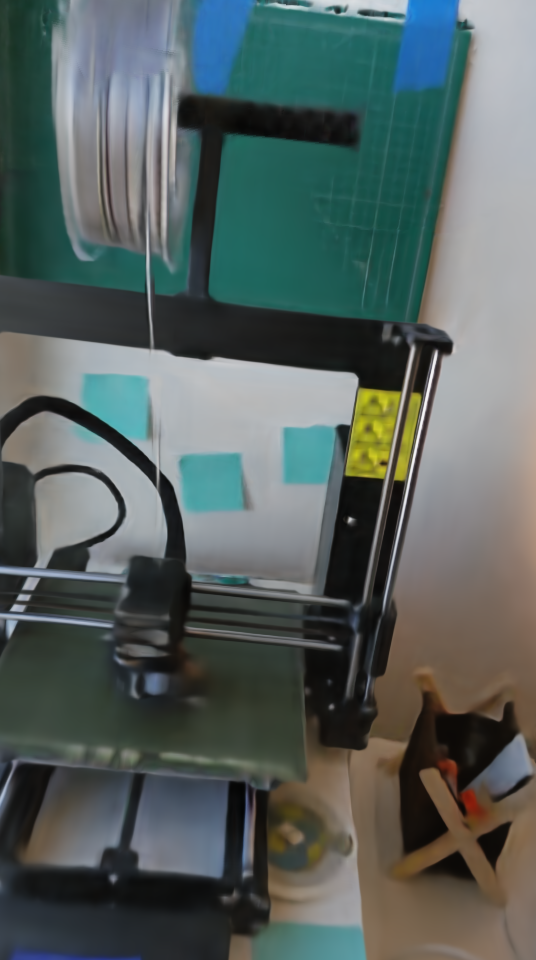} &
            \includegraphics[width=0.15\textwidth]{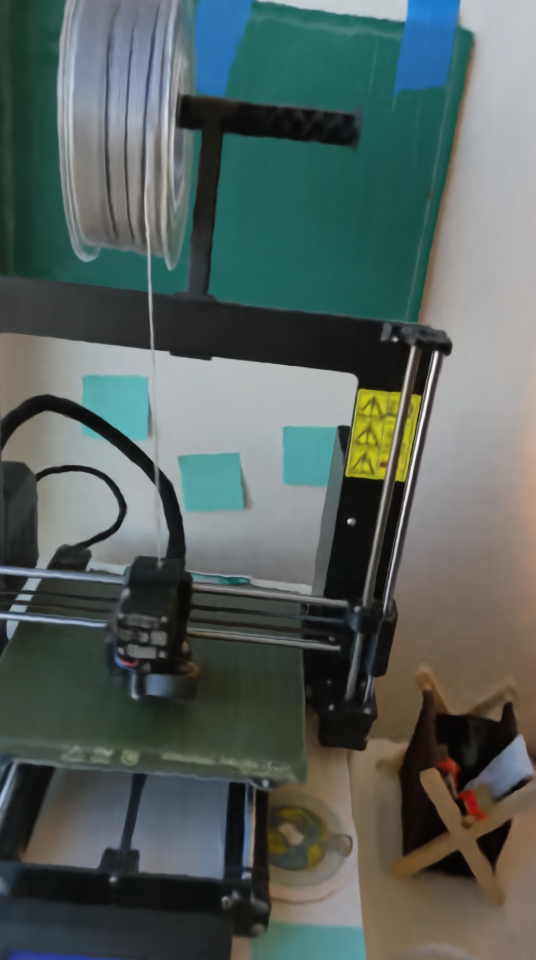}

        \end{tabular}
    }
\vspace{-5pt}
	\caption{\textbf{Qualitative comparison on the HyperNeRF dataset.} We show synthesized images on the HyperNeRF dataset of our method and other competing methods.
}\label{fig:hyper-quality}
\end{figure*}

\noindent\textbf{Quantitative Results}
We show more qualitative comparisons in~\fref{fig:dnerf-quality} and ~\fref{fig:hyper-quality} for D-NeRF synthetic dataset~\cite{pumarola2021_dnerf_cvpr21} and HyperNeRF dataset~\cite{park2021hypernerf}. In our supplementary video, we also showcase the temporal interpolation capability of our method when maintaining a fixed camera viewpoint while time evolves. Additionally, we demonstrate the ability to synthesize novel viewpoints while keeping the time fixed and observing the scene from arbitrary viewpoints.

\begin{figure*}
    \centering
    \addtolength{\tabcolsep}{-6.5pt}
    \footnotesize{
        \setlength{\tabcolsep}{1pt} % Default value: 6pt
        \begin{tabular}{p{8.2pt}ccccccc}
            & $t_0$ & $t_1$ & $t_2$ & $t_3$ & $t_4$ & $t_5$  \\
        \raisebox{35pt}{\rotatebox[origin=c]{90}{Bouncing Balls}}&
             % \raisebox{20pt}{\rotatebox[origin=c]{90}{Ours}}&
             \includegraphics[width=0.15\textwidth]{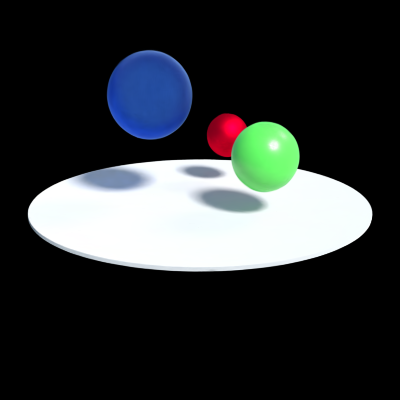} &
            \includegraphics[width=0.15\textwidth]{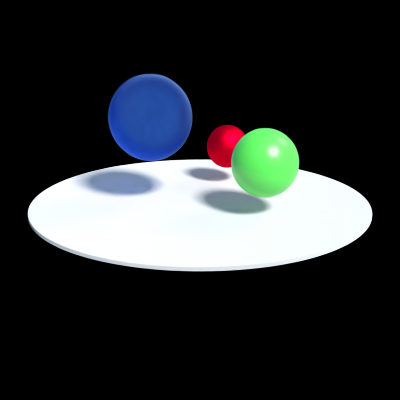} &
            \includegraphics[width=0.15\textwidth]{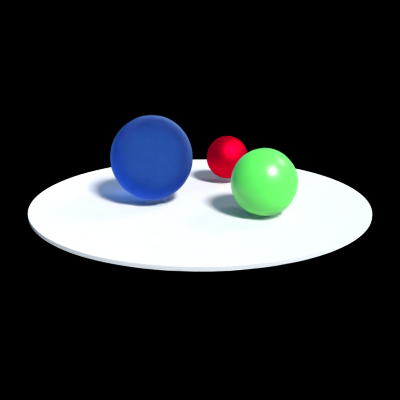} &
            \includegraphics[width=0.15\textwidth]{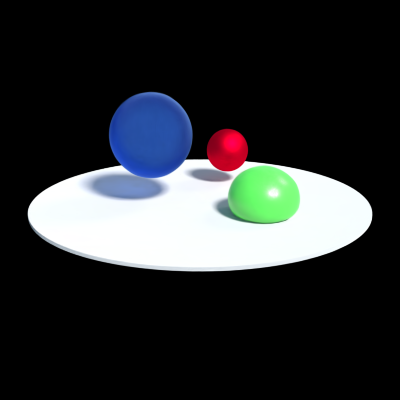} &
            \includegraphics[width=0.15\textwidth]{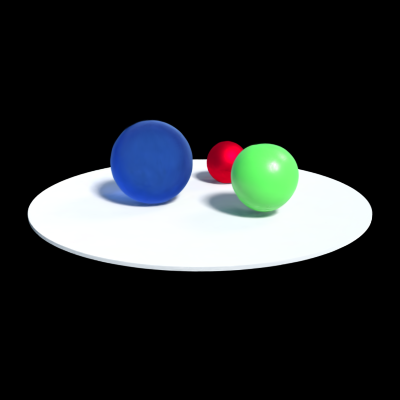} &
            \includegraphics[width=0.15\textwidth]{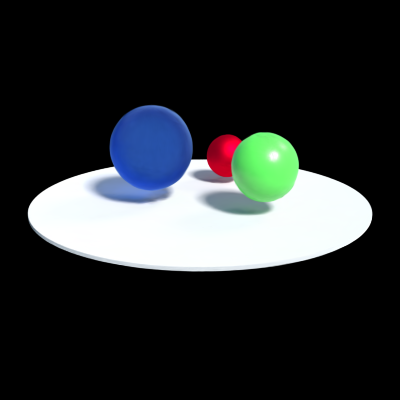} 
             \\
        \raisebox{35pt}{\rotatebox[origin=c]{90}{Hell Warrior}}&
             \includegraphics[width=0.15\textwidth]{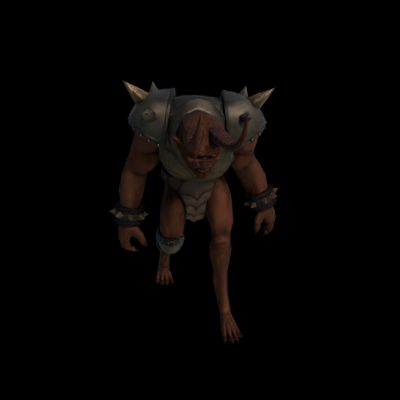} &
            \includegraphics[width=0.15\textwidth]{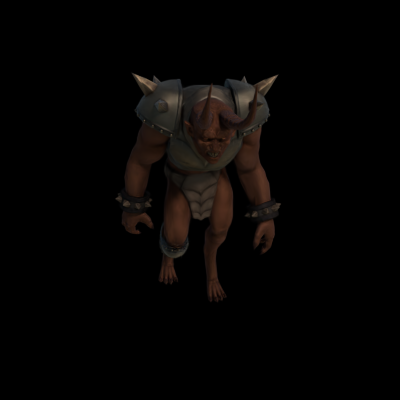} &
            \includegraphics[width=0.15\textwidth]{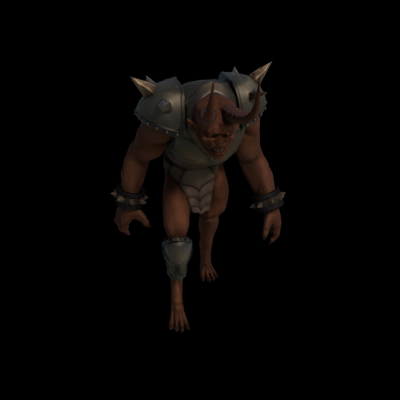} &
            \includegraphics[width=0.15\textwidth]{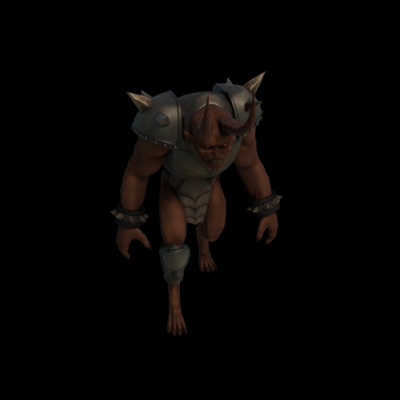} &
            \includegraphics[width=0.15\textwidth]{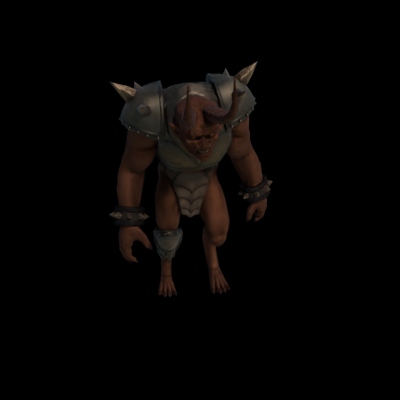} &
            \includegraphics[width=0.15\textwidth]{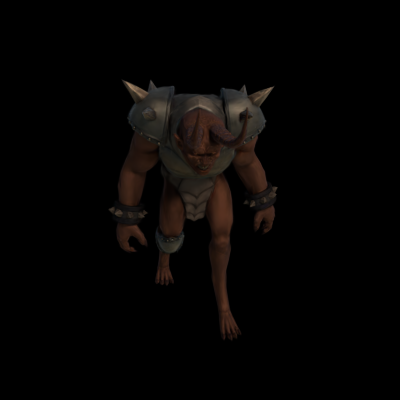} 
            \\
            \raisebox{35pt}{\rotatebox[origin=c]{90}{Hook}}&
             \includegraphics[width=0.15\textwidth]{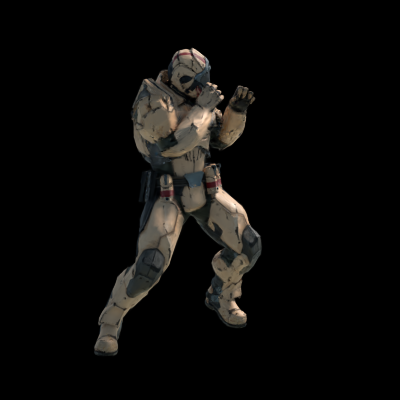} &
            \includegraphics[width=0.15\textwidth]{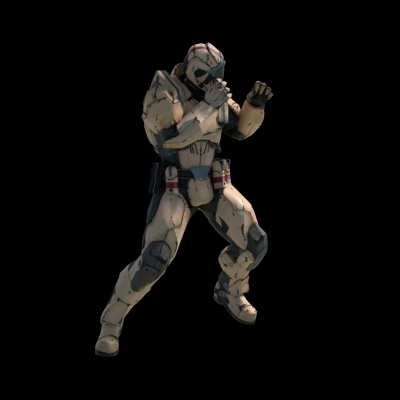} &
            \includegraphics[width=0.15\textwidth]{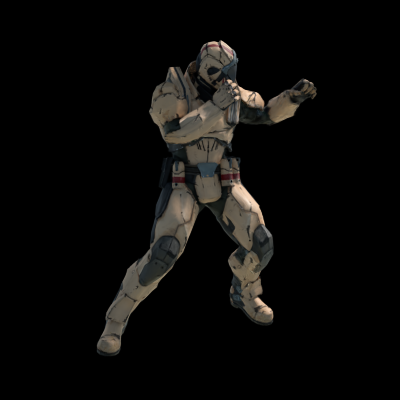} &
            \includegraphics[width=0.15\textwidth]{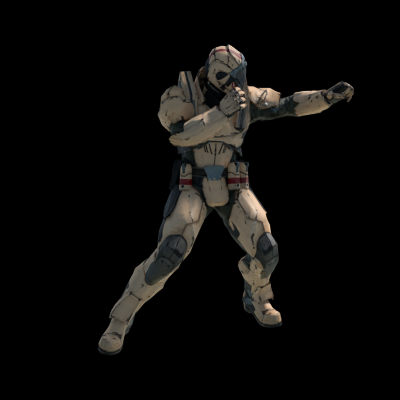} &
            \includegraphics[width=0.15\textwidth]{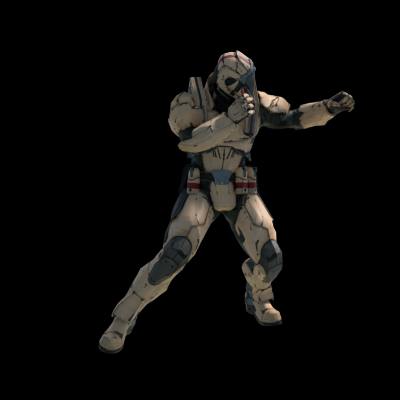} &
            \includegraphics[width=0.15\textwidth]{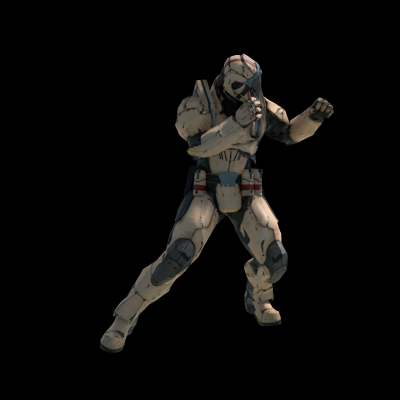} 

            \\
            \raisebox{35pt}{\rotatebox[origin=c]{90}{Jumping Jacks}}&
            \includegraphics[width=0.15\textwidth]{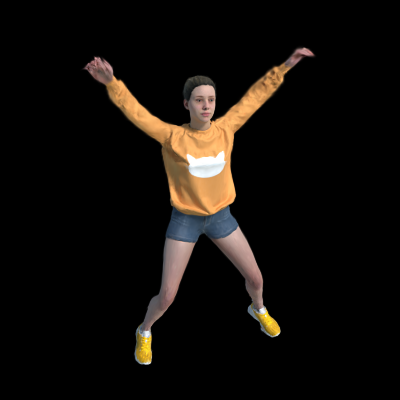} &
            \includegraphics[width=0.15\textwidth]{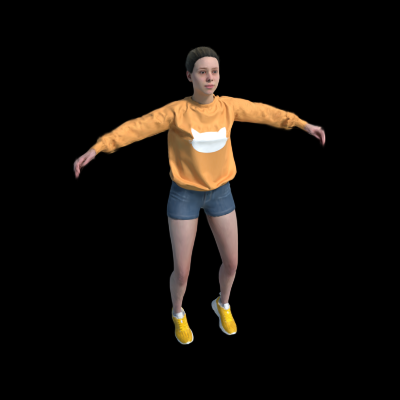} &
            \includegraphics[width=0.15\textwidth]{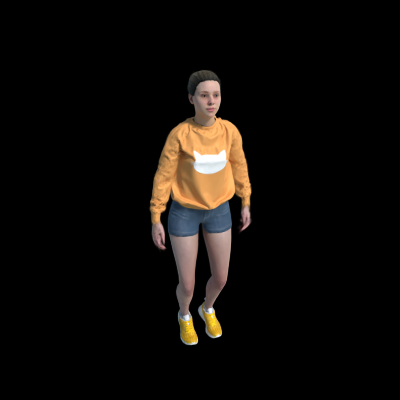} &
            \includegraphics[width=0.15\textwidth]{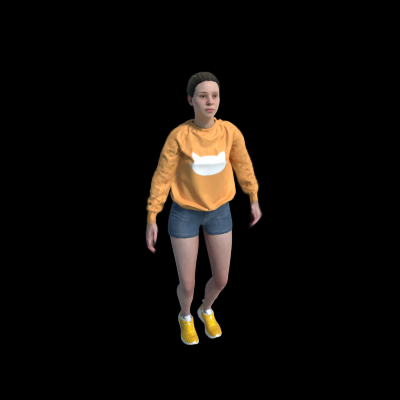} &
            \includegraphics[width=0.15\textwidth]{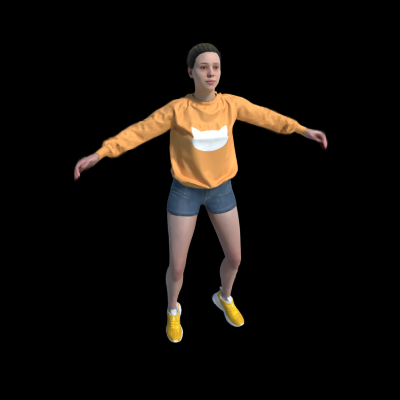} &
            \includegraphics[width=0.15\textwidth]{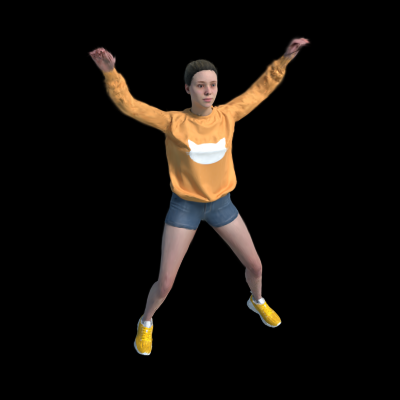} 

            \\
            \raisebox{35pt}{\rotatebox[origin=c]{90}{Lego}}&
            \includegraphics[width=0.15\textwidth]{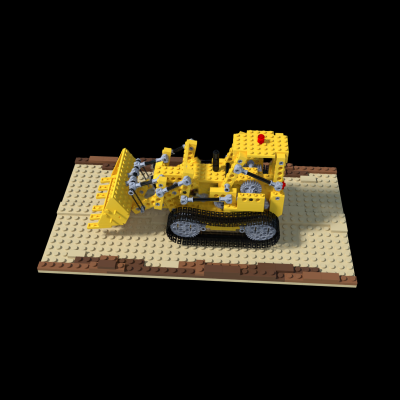} &
            \includegraphics[width=0.15\textwidth]{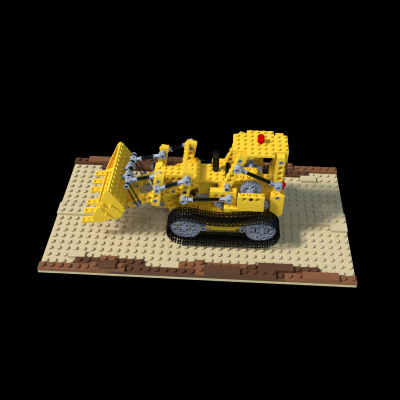} &
            \includegraphics[width=0.15\textwidth]{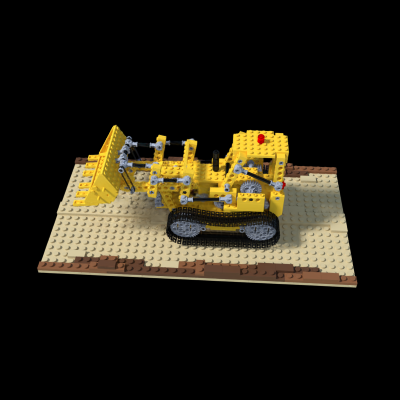} &
            \includegraphics[width=0.15\textwidth]{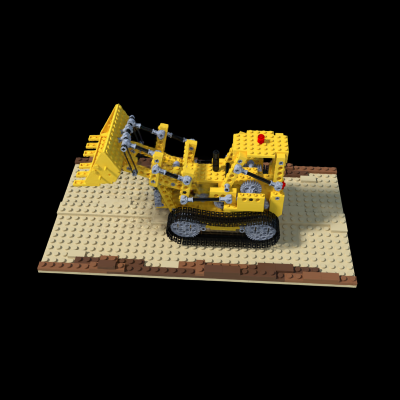} &
            \includegraphics[width=0.15\textwidth]{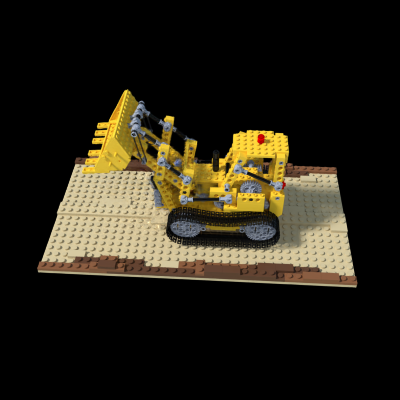} &
            \includegraphics[width=0.15\textwidth]{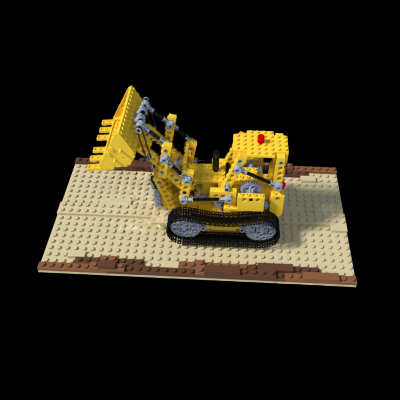} 

            \\
            \raisebox{35pt}{\rotatebox[origin=c]{90}{Mutant}}&
            \includegraphics[width=0.15\textwidth]{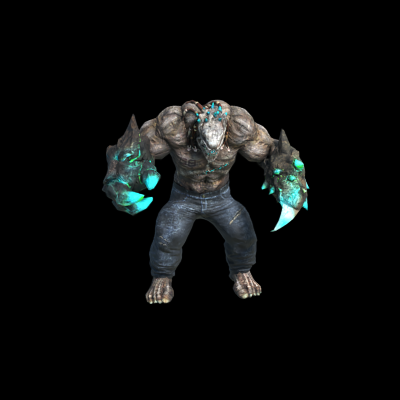} &
            \includegraphics[width=0.15\textwidth]{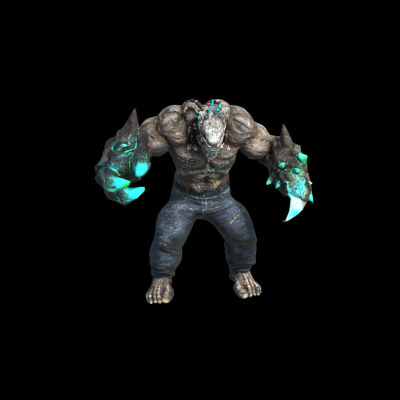} &
            \includegraphics[width=0.15\textwidth]{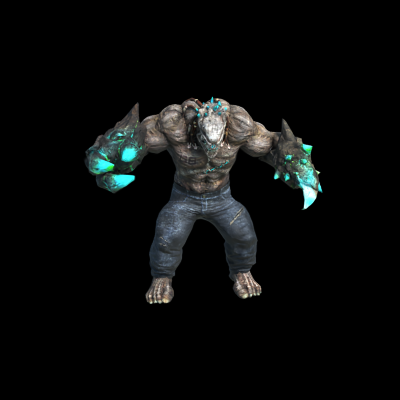} &
            \includegraphics[width=0.15\textwidth]{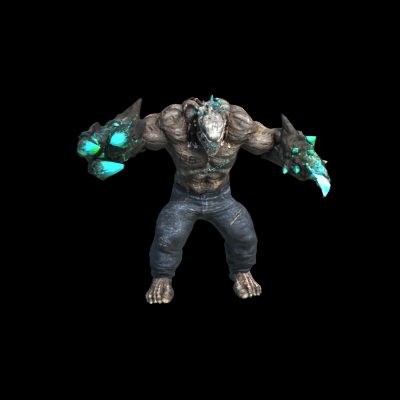} &
            \includegraphics[width=0.15\textwidth]{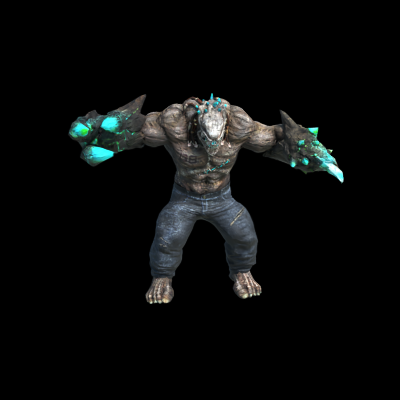} &
            \includegraphics[width=0.15\textwidth]{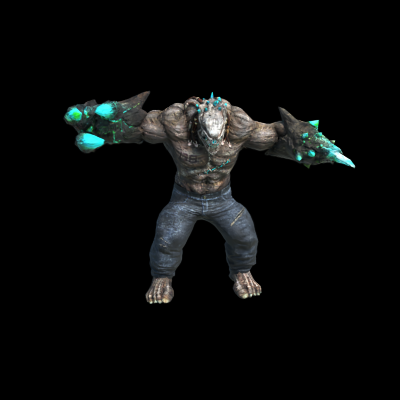} 

            \\
            \raisebox{35pt}{\rotatebox[origin=c]{90}{Stand Up}}&
             \includegraphics[width=0.15\textwidth]{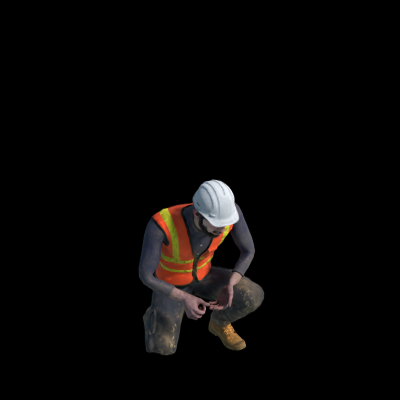} &
            \includegraphics[width=0.15\textwidth]{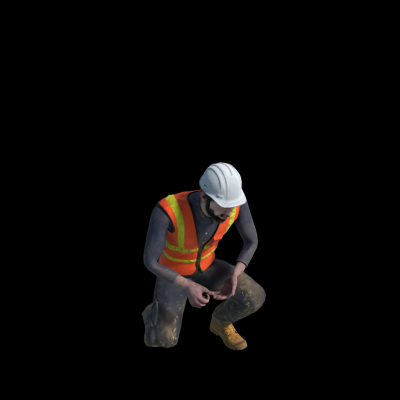} &
            \includegraphics[width=0.15\textwidth]{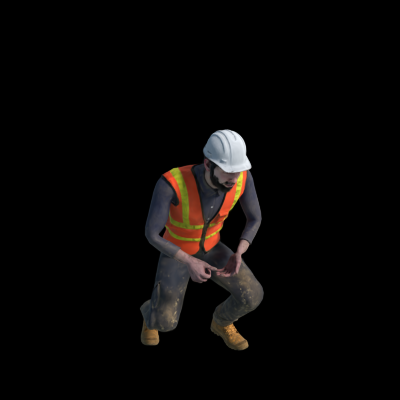} &
            \includegraphics[width=0.15\textwidth]{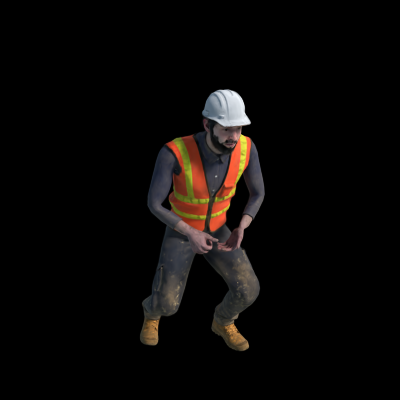} &
            \includegraphics[width=0.15\textwidth]{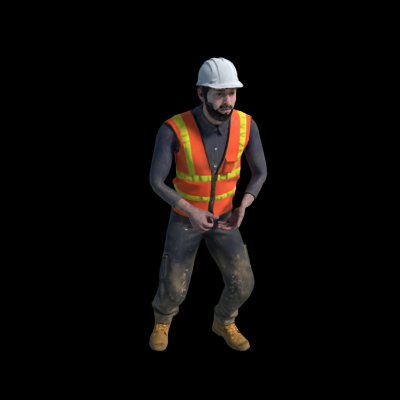} &
            \includegraphics[width=0.15\textwidth]{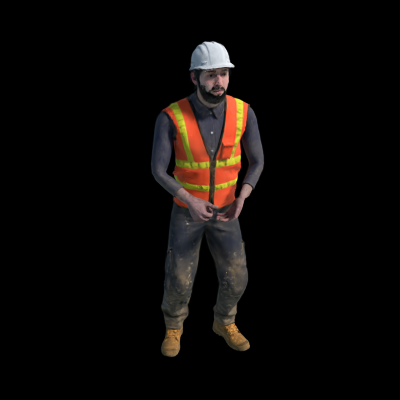} 

            \\
            \raisebox{35pt}{\rotatebox[origin=c]{90}{T-Rex}}&
             \includegraphics[width=0.15\textwidth]{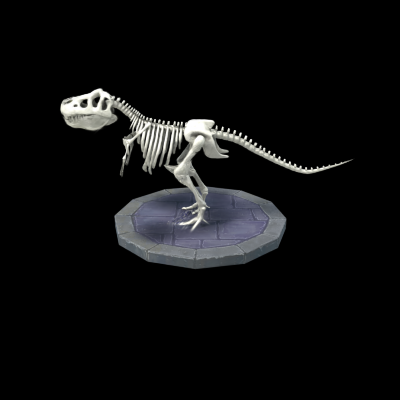} &
            \includegraphics[width=0.15\textwidth]{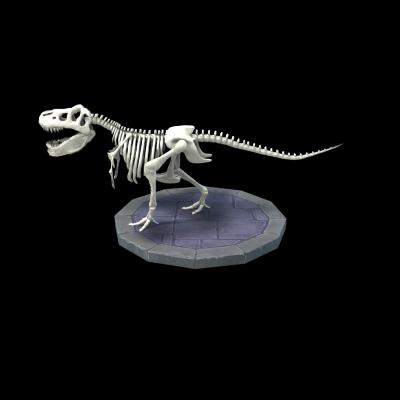} &
            \includegraphics[width=0.15\textwidth]{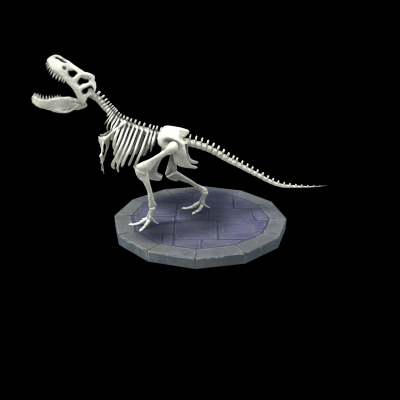} &
            \includegraphics[width=0.15\textwidth]{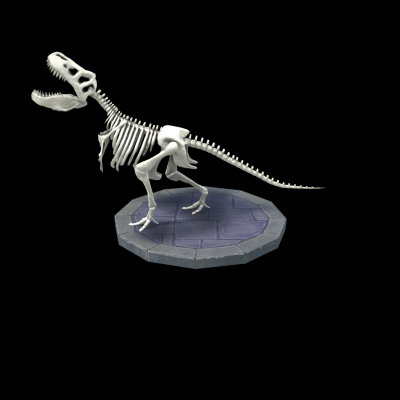} &
            \includegraphics[width=0.15\textwidth]{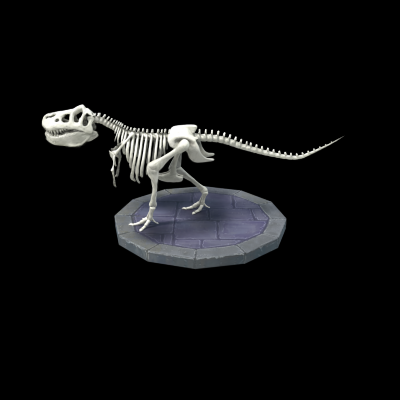} &
            \includegraphics[width=0.15\textwidth]{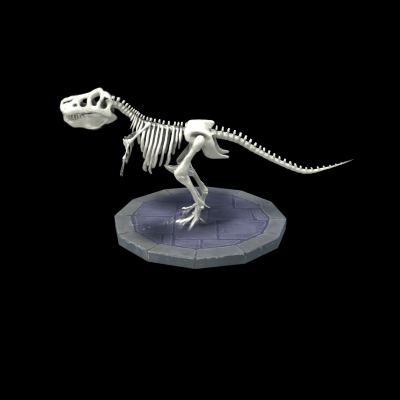}

        \end{tabular}
    }
\vspace{-5pt}
	\caption{\textbf{Temporal Interpolation Capability on the D-NeRF synthetic dataset.} We show the temporal interpolation capabilities of our method. Specifically, we showcase our ability to perform time interpolation by maintaining a fixed camera viewpoint while observing the temporal changes in scene content.
}\label{fig:dnerf-quality-time}
\end{figure*}

\begin{figure*}
    \centering
    \addtolength{\tabcolsep}{-6.5pt}
    \footnotesize{
        \setlength{\tabcolsep}{1pt} % Default value: 6pt
        \begin{tabular}{p{8.2pt}cccccc}
            & $t_0$ & $t_1$ & $t_2$ & $t_3$   \\
        \raisebox{65pt}{\rotatebox[origin=c]{90}{Peel Banana}}&
             % \raisebox{20pt}{\rotatebox[origin=c]{90}{Ours}}&
             \includegraphics[width=0.17\textwidth]{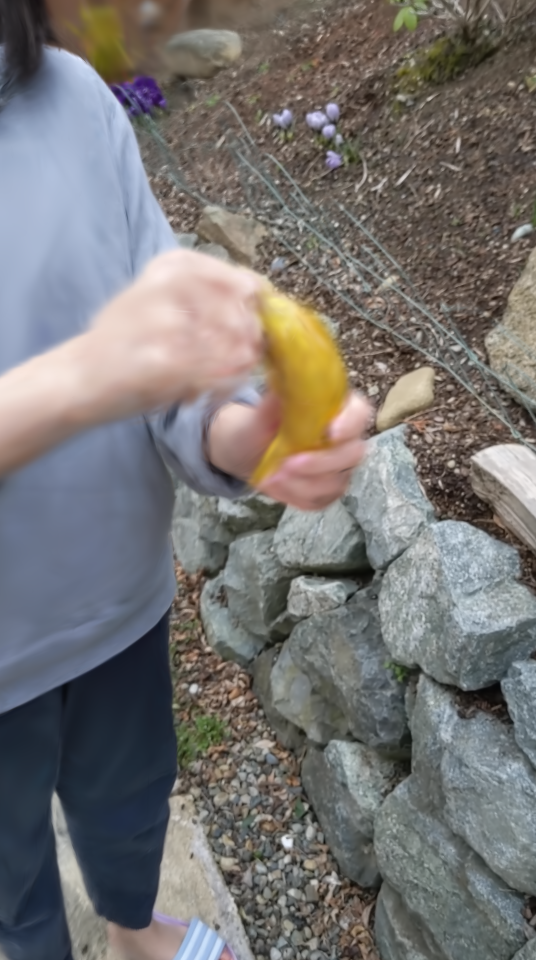} &
            \includegraphics[width=0.17\textwidth]{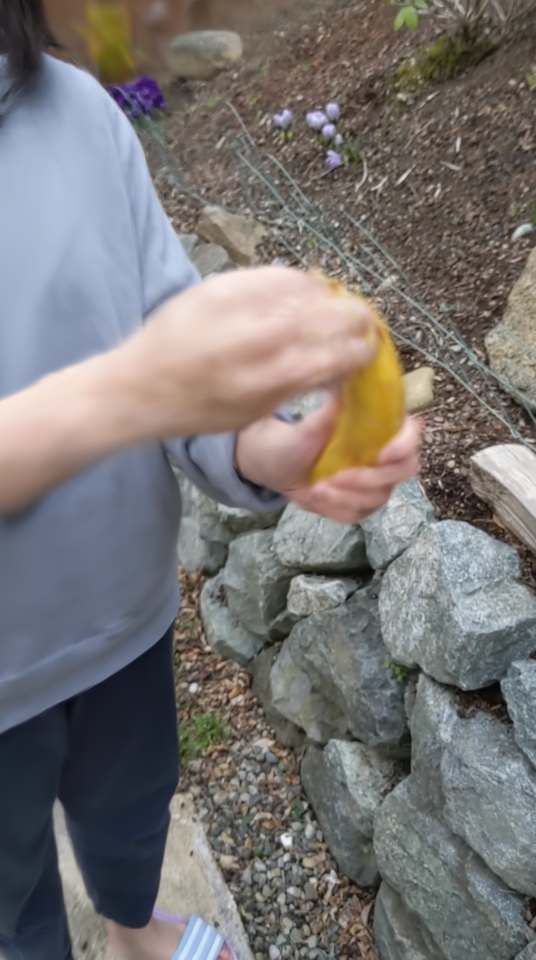} &
            \includegraphics[width=0.17\textwidth]{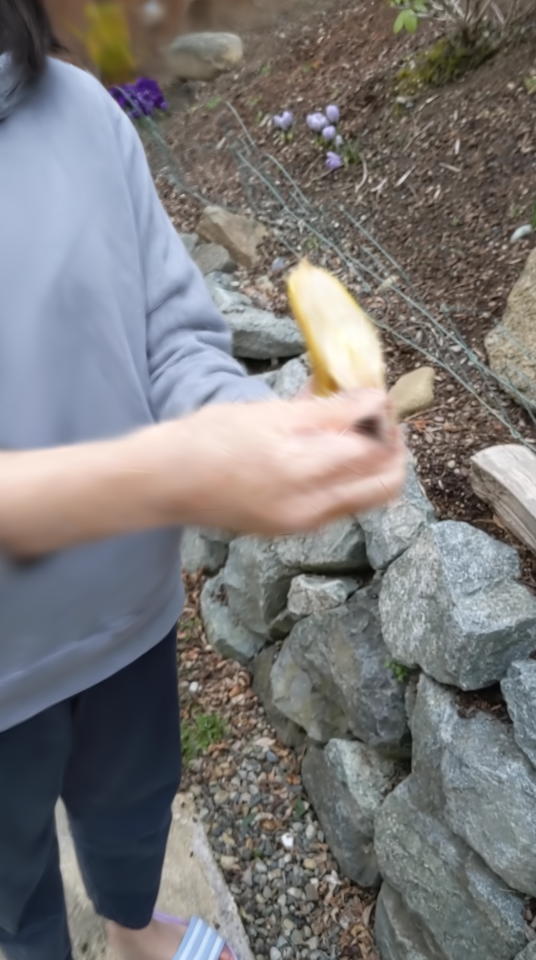} &
            \includegraphics[width=0.17\textwidth]{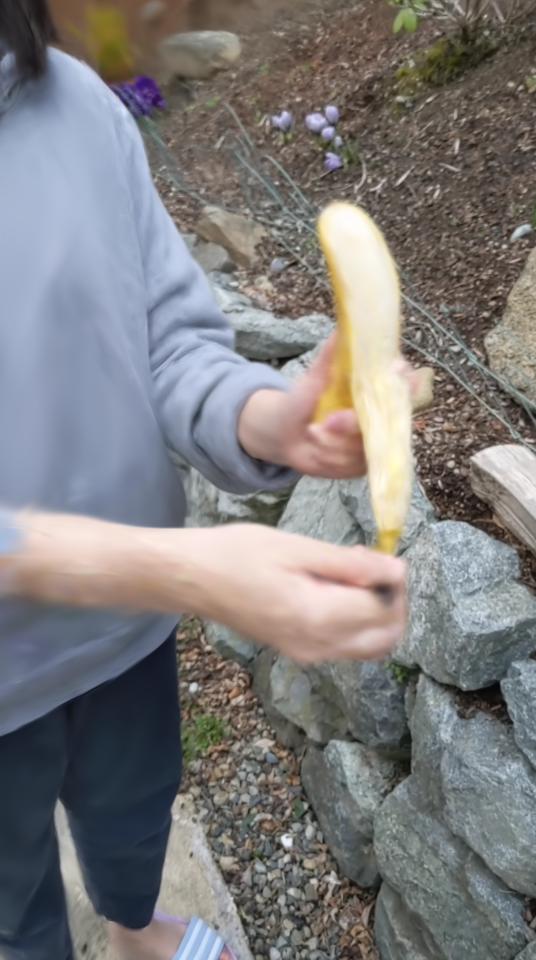} &

             \\
        \raisebox{65pt}{\rotatebox[origin=c]{90}{Chicken}}&
             \includegraphics[width=0.17\textwidth]{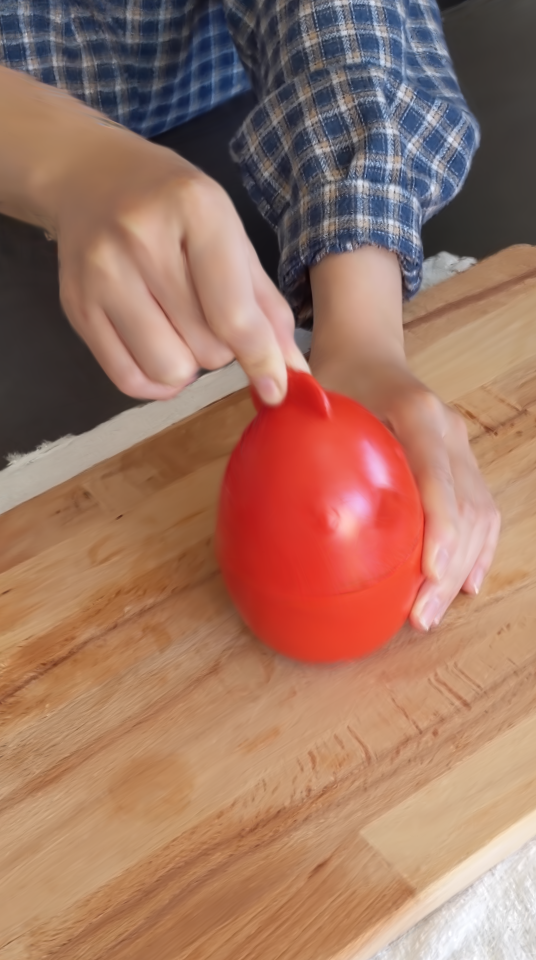} &
            \includegraphics[width=0.17\textwidth]{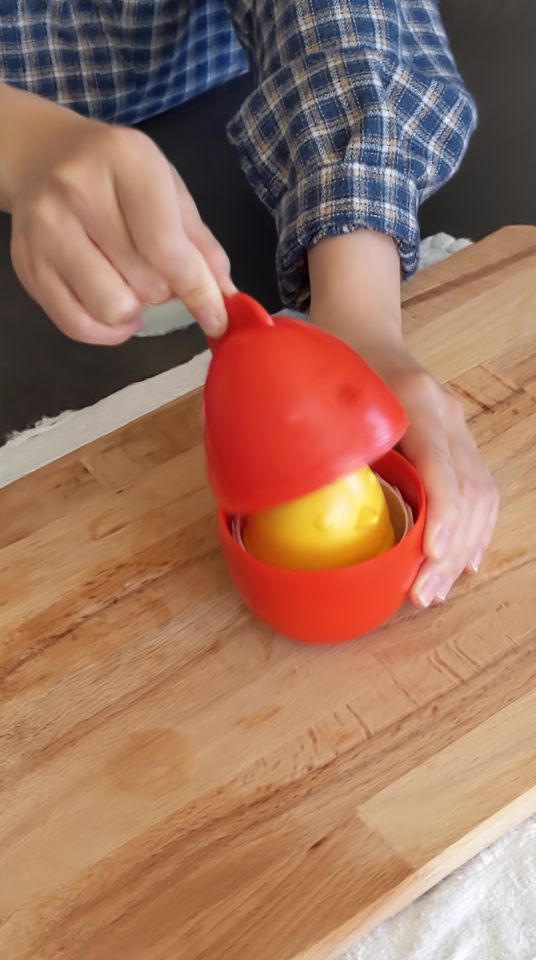} &
            \includegraphics[width=0.17\textwidth]{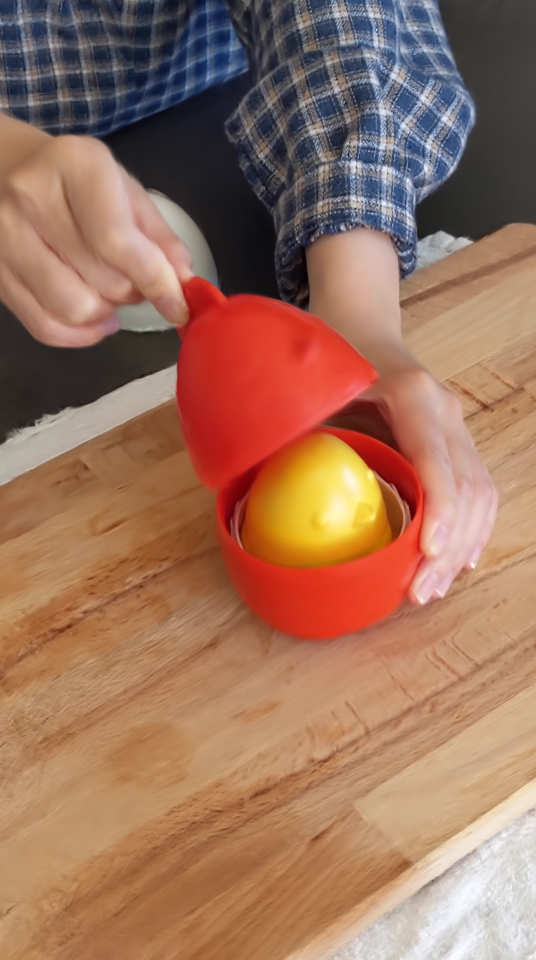} &
            \includegraphics[width=0.17\textwidth]{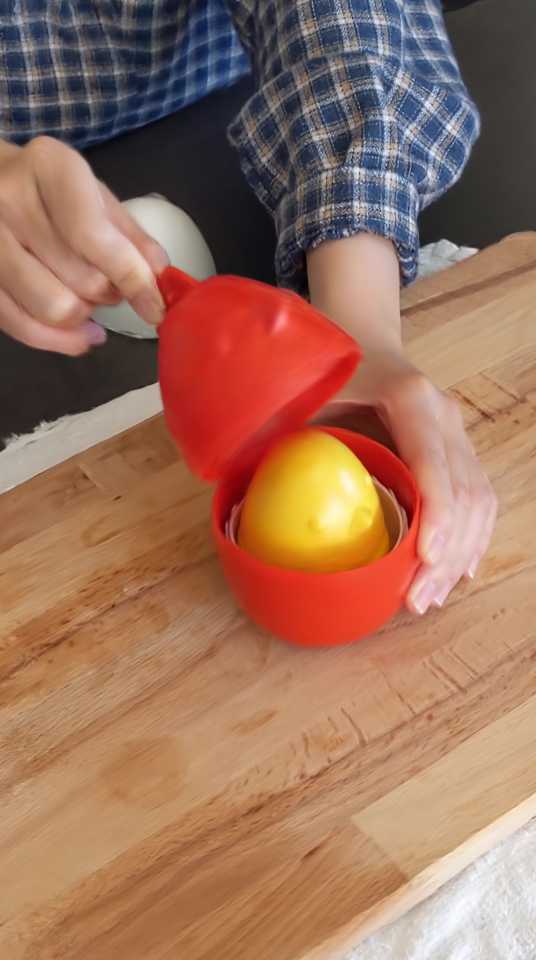} &

            \\
            \raisebox{65pt}{\rotatebox[origin=c]{90}{Split Cookie}}&
             \includegraphics[width=0.17\textwidth]{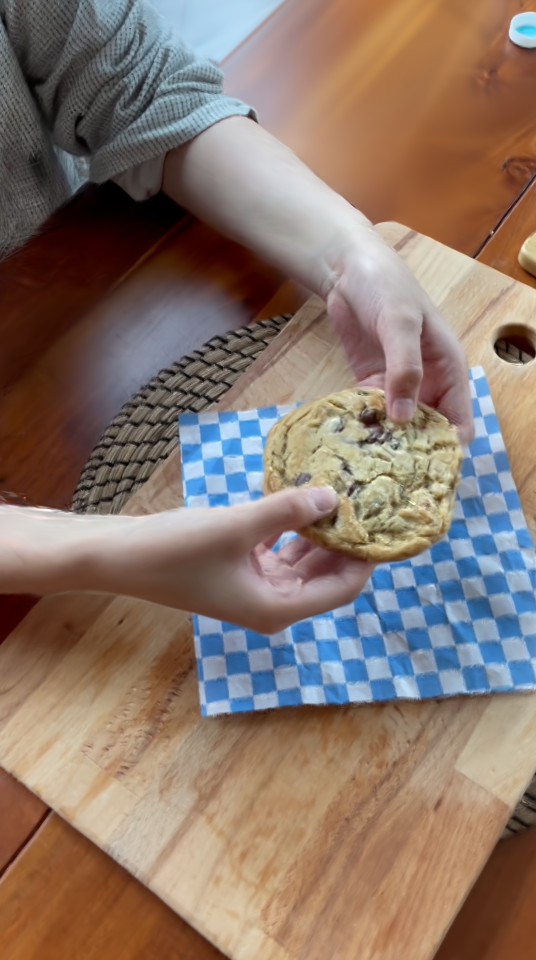} &
            \includegraphics[width=0.17\textwidth]{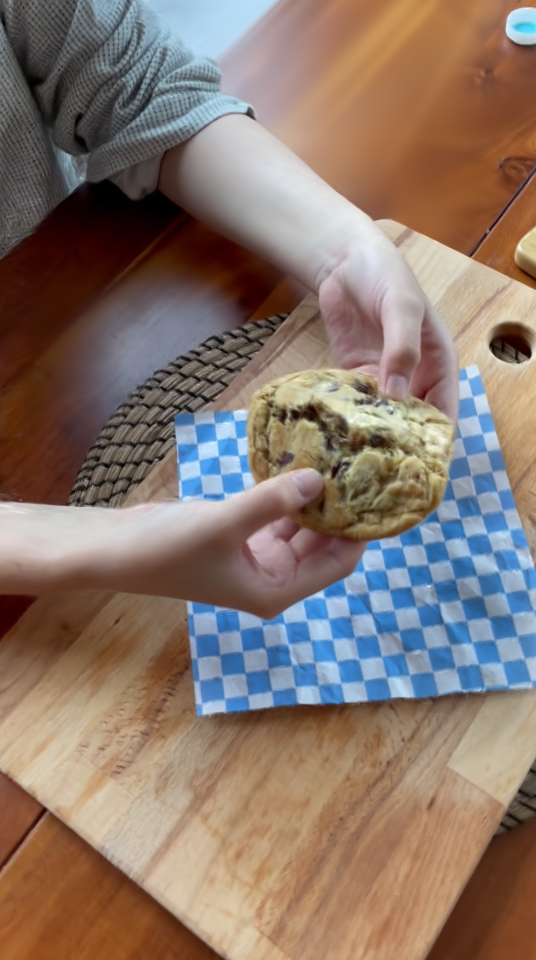} &
            \includegraphics[width=0.17\textwidth]{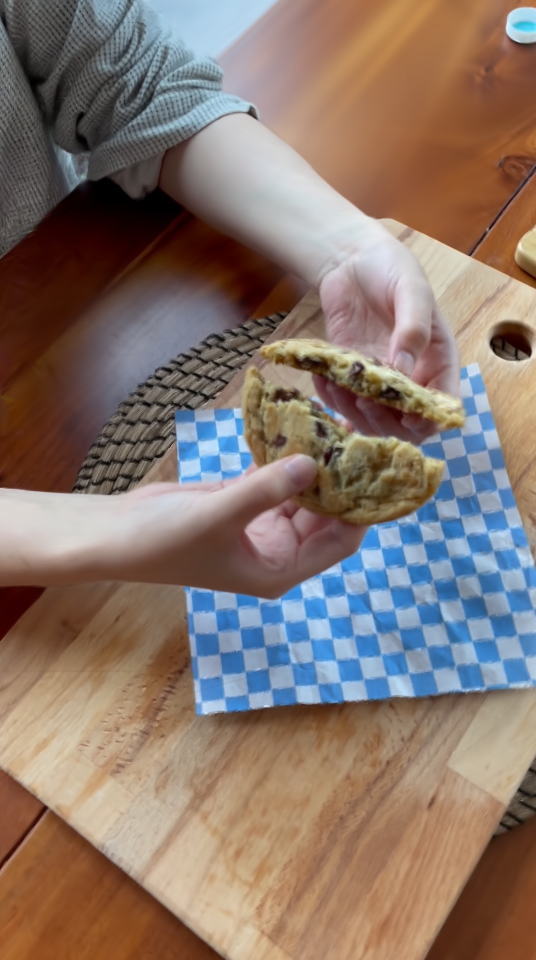} &
            \includegraphics[width=0.17\textwidth]{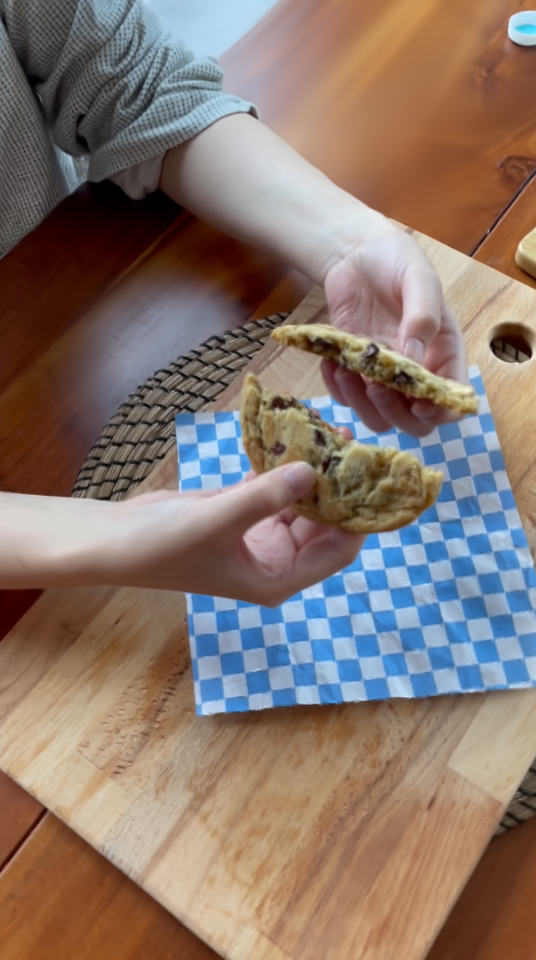} &

            \\
            \raisebox{20pt}{\rotatebox[origin=c]{90}{Cut Lemon}}&
            \includegraphics[width=0.17\textwidth]{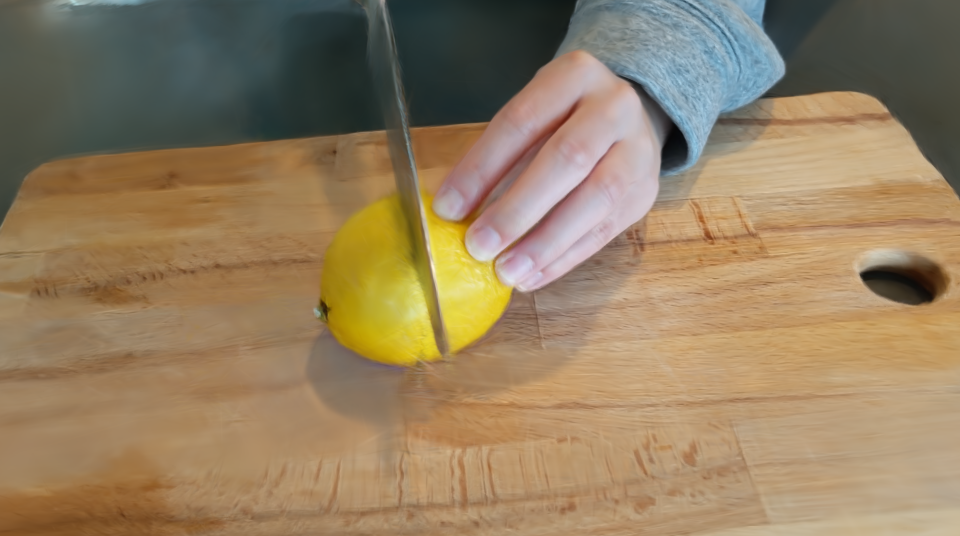} &
            \includegraphics[width=0.17\textwidth]{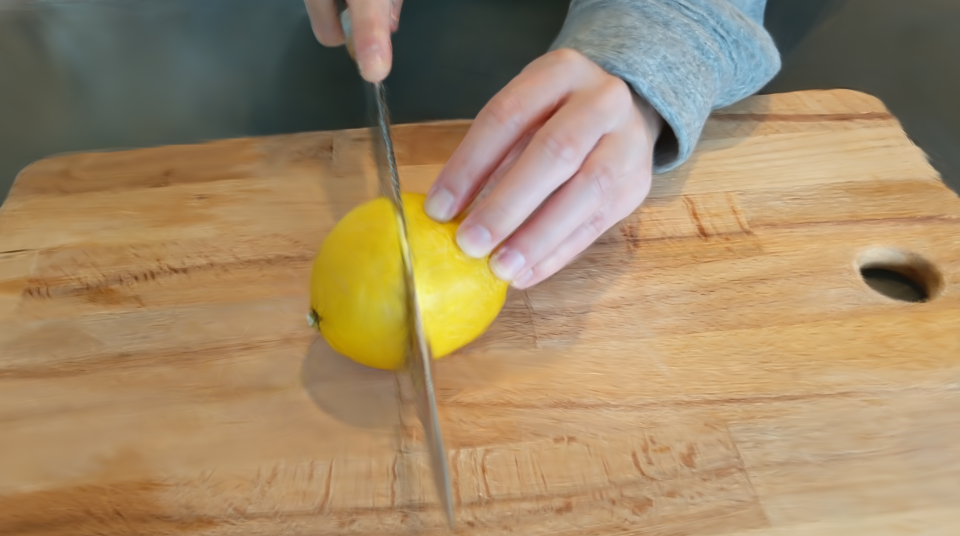} &
            \includegraphics[width=0.17\textwidth]{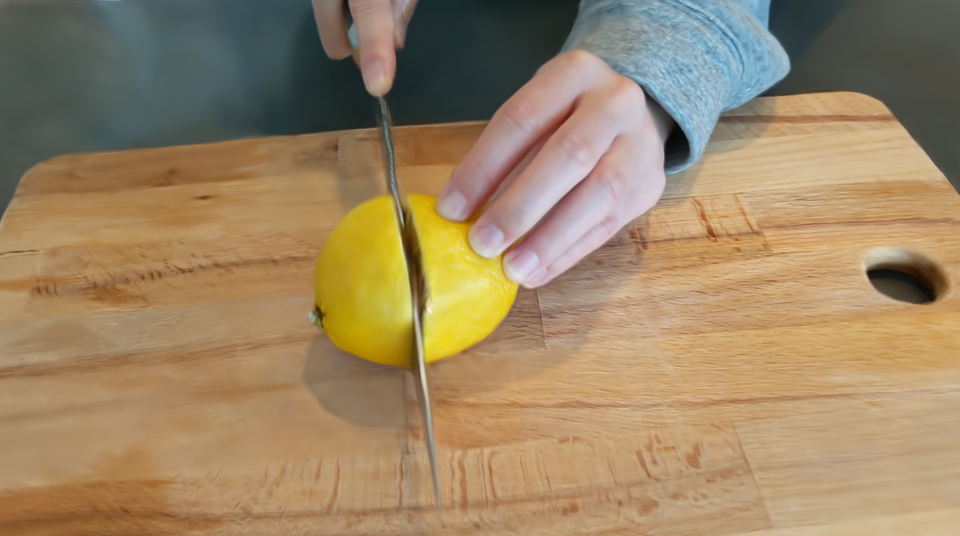} &
            \includegraphics[width=0.17\textwidth]{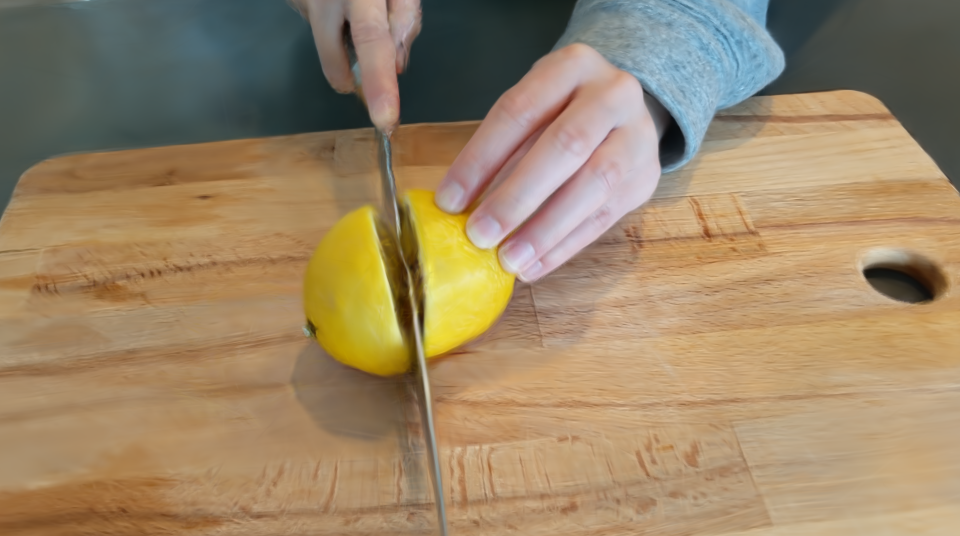} &

        \end{tabular}
    }
\vspace{-5pt}
	\caption{\textbf{Temporal Interpolation Capability on HyperNeRF dataset.} We show the temporal interpolation capabilities of our method. Specifically, we showcase our ability to perform time interpolation by maintaining a fixed camera viewpoint while observing the temporal changes in scene content.
}\label{fig:hyper-quality-time}
\end{figure*}

\noindent\textbf{Temporal Interpolation}
We show the temporal interpolation ability of our method. In Fig.~\ref{fig:dnerf-quality-time} and Fig.~\ref{fig:hyper-quality-time}, we fix the camera viewpoint and show the results for temporal changes of the D-NeRF synthetic dataset~\cite{pumarola2021_dnerf_cvpr21} and HyperNeRF dataset~\cite{park2021hypernerf}. Our method shows great temporal interpolation abilities for both synthetic and real datasets. More results are presented in our homepage.

\subsection{Limitations and Impacts}
\noindent\textbf{Limitations}
First, our proposed method represents the deformation of Gaussians from the canonical space to time space. However, it can only chronicle a point within the scene from start to finish, lacking the capability to depict a point that abruptly emerges or disappears in the scene at a specific moment. 
Second, our proposed method essentially describes the motion and deformation of points in the canonical space. It necessitates acquiring precise camera poses in advance. However, in the context of dynamic scene modeling, obtaining accurate camera poses is inherently very challenging. Our approach is also constrained by this limitation.
Last, our method struggles to describe excessively complex motions and long time videos, such as rapid movements of objects within the scene. This challenge results in the network facing difficulties in estimating point motions, ultimately leading to failures, as shown in ~\fref{fig:dynerf}, we provide some cases in the test camera on Neu3DV dataset~\cite{li2022neural}. Due to the lack of explicit modeling of motion, our method exhibits insufficient capability in capturing fine-grained movements over long temporal sequences. However, it still maintains the ability to describe general motions, such as the swinging of curtains and human body movements.

\noindent\textbf{Broader Impacts}
Our proposed method can be applied to various industries, including visual effects synthesis in the film industry, game modeling, autonomous driving simulation, and more. For the film industry and game modeling, dynamic scenes can be synthesized by our method. In autonomous driving simulation, our proposed method can provide more data from different viewpoints, which will contribute to the advancement of autonomous driving.

% \begin{figure}[th]
% \centering
% \includegraphics[width=\columnwidth]{images/supp/dnerfpcd.png}
% %\vspace{-6pt}
% \caption{Visualization of the point cloud of Lego with different iterations.}
% \label{fig:dnerfpcd}
% \vspace{-6pt}
% \end{figure}

% \begin{figure}[th]
% \centering
% \includegraphics[width=\columnwidth]{images/supp/bananapcd.png}
% %\vspace{-6pt}
% \caption{Visualization of the point cloud of Peel banana with different iterations.}
% \label{fig:bananapcd}
% \vspace{-6pt}
% \end{figure}

\subsection{Future Work}
In the future, we plan to exploit the motion mask to distinguish the dynamic points and static points of the scene, which will decrease the computing resource by only estimating the deformation of dynamic points. Also, we will investigate explicit motion modeling by exploiting the foreground and background motion segmentation cues.

%\clearpage

{
    \small
    \bibliographystyle{ieeenat_fullname}
    
    \bibliography{sec/9_references}
}


\end{document}